\documentclass[journal]{IEEEtran}
\usepackage[utf8]{inputenc}
\usepackage{mathtools,lipsum,cuted}
\usepackage{amsmath,amsfonts}
\usepackage{algorithmic}
\usepackage{array}
\usepackage{subfigure}
\usepackage{stfloats}
\usepackage{url}
\usepackage{bm}
\usepackage{verbatim}
\usepackage{graphicx}
\usepackage{booktabs}
\usepackage{booktabs}
\usepackage{enumerate}
\renewcommand\eqref[1]{(\autoref{#1})}
\usepackage{mathtools,lipsum,cuted}
\usepackage{amsmath}     
\usepackage{amssymb}                       
\usepackage{mathrsfs}    
\usepackage{upgreek}
\usepackage{bm}     
\usepackage{authblk}
\usepackage{setspace}
\usepackage{graphicx}
\usepackage{amsfonts,amssymb}
\usepackage{caption}
\graphicspath{ {./figures/} }
\graphicspath{ {./Author/} }
\usepackage{amsmath}
\usepackage{lineno}
\usepackage{colortbl}
\usepackage{bbding}
\usepackage{multirow}
\usepackage{makecell}
\usepackage{blindtext}
\usepackage{xcolor}
\usepackage{color}
\usepackage{setspace}
\usepackage{makecell}
\usepackage{threeparttable}
\usepackage{hyperref}
\usepackage[section]{placeins}
\usepackage{cleveref}
\usepackage{hhline}
\usepackage{threeparttable}
\usepackage{tikz}
\usepackage{dsfont}
\usepackage{diagbox}

\hypersetup{
	colorlinks=true,
	linkcolor=cyan,
	filecolor=blue,      
	urlcolor=black,
	citecolor=green,
}
\newcommand{\subsubsubsection}[1]{\paragraph{#1}\mbox{}}
\DeclareMathAlphabet\mathbfcal{OMS}{cmsy}{b}{n}
\begin{document}
\title{Spatial-Spectral Diffusion Contrastive Representation Network for Hyperspectral Image Classification}
\author{Yimin Zhu, Linlin Xu, ~\IEEEmembership{Member,~IEEE}
\thanks{Corresponding author Linlin Xu is with the Department of Geomatics
Engineering, University of Calgary, Canada (email: lincoln.xu@ucalgary.ca)

Yimin Zhu is with the Department of Geomatics Engineering, University of Calgary, Canada (email: yimin.zhu@ucalgary.ca)
}}

\markboth{Journal of \LaTeX\ Class Files,~Vol.~18, No.~9, September~2020}%
{How to Use the IEEEtran \LaTeX \ Templates}

\maketitle

\begin{abstract}
Although efficient extraction of discriminative spatial-spectral features is critical for hyperspectral images classification (HSIC), it is difficult to achieve these features due to factors such as the spatial-spectral heterogeneity and noise effect. This paper presents a Spatial-Spectral Diffusion Contrastive Representation Network (DiffCRN), based on denoising diffusion probabilistic model (DDPM) combined with contrastive learning (CL) for HSIC, with the following characteristics. First, to improve spatial-spectral feature representation, instead of adopting the UNets-like structure which is widely used for DDPM, we design a novel staged architecture with spatial self-attention denoising module (SSAD) and spectral group self-attention denoising module (SGSAD) in DiffCRN with improved efficiency for spectral-spatial feature learning. Second, to improve unsupervised feature learning efficiency, we design new DDPM model with logarithmic absolute error (LAE) loss and CL that improve the loss function effectiveness and increase the instance-level and inter-class discriminability. Third, to improve feature selection, we design a learnable approach based on pixel-level spectral angle mapping (SAM) for the selection of time steps in the proposed DDPM model in an adaptive and automatic manner. Last, to improve feature integration and classification, we design an Adaptive weighted addition modul (AWAM) and Cross time step Spectral-Spatial Fusion Module (CTSSFM) to fuse time-step-wise features and perform classification. Experiments conducted on widely used four HSI datasets demonstrate the improved performance of the proposed DiffCRN over the classical backbone models and state-of-the-art GAN, transformer models and other pretrained methods. The source code and pre-trained model will be made available publicly. 

\end{abstract}


\IEEEpeerreviewmaketitle
\section{Introduction}
\IEEEPARstart{H}{yperspectral} image classification (HSIC) as a fundamental task in hyperspectral image (HSI) processing, aims to classify each pixel's in HSI to achieve semantic classification maps, which is critical to support various applications, e.g., land cover and crop mapping \cite{Lu2020_RS, Wang2021_AIR}, environmental monitoring \cite{Tong2014_JSTARS, Yang2017_TGRS}, mineral detection and mapping \cite{Murphy2014_TGRS, CharlotteA2011_IJRS} etc. Although HSIC is critical to support various environmental applications, achieving efficient HSIC is a challenging task, because it requires the extraction of discriminative spatial-spectral features that can capture subtle class signatures of different land cover and land use classes \cite{Bioucas2013_MGRS}. Therefore, many machine learning models have been proposed to improve the modeling of the complex spectral-spatial correlation in HSI in unsupervised \cite{Yu2022_LGRS}, semi-supervised \cite{Persello2014_TGRS} and supervised approaches \cite{Tuia2011_JSTSP}. 

Comparing with traditional machine learning methods, the deep learning (DL) methods have demonstrated improved spatial-spectral feature learning capability based on deep convolutional neural networks (CNNs) \cite{Yangxf2018_TGRS, Roy2020_LGRS, zhao2016_TGRS, Ben2018_TGRS, Zhong2018_TGRS}. Besides, other networks are exploited constantly to overcome the limitation of CNNs, such as transformers \cite{He2019_TGRS, He2020_TGRS, Hongdf2022_TGRS, Ibañez2022_TGRS}, graph neural networks \cite{Qin2019_LGRS, Hong2021GCN_TGRS}, capsule network \cite{Paoletti2019_TGRS, Paoletti2022_TGRS}, and multi-view learning \cite{Liubing2021_TGRS, Li2022_NC}. However, most of these approaches are supervised methods, which rely predominantly on training samples to learn the explicit mappings between semantic labels and spatial-spectral data, leading to drawbacks such as high labeling and data preparation cost, as well as poor generalization capability outside the discriminant space. As a result, some recent works \cite{Patel2020_GARSS, Jamshidpour2020_RS, Li2022_TGRS} tend to exploit semi-supervised learning combining manifold learning, active learning and pseudo-label method for HSIC with optimizing abundant unlabeled data to characterize spectral-spatial relations. 

Unsupervised feature extraction is another feasible approach to model the spatial-spectral correlation in HSI for HSIC. Autoencoder (AE) is a common unsupervised framework which provides a probability to an input are designed to extract features hierarchically by minimize the reconstruction error, which has been widely used for HSIC \cite{cheys2015_JSTARS, Romero2016_TGRS, Sun2017_LGRS, Zhangxr2017_LGRS, Koda2020_LGRS}. However, the network layers are completely connected which contribute to high computing resources. Some enhanced AE approaches are applied into HSIC, e.g., fully conv-deconv network with residual learning \cite{Mou2017_IGARSS}, and the hypergraph-structured autoencoder \cite{Cai2022_LGRS}. Considering that different classes may have different spectral shifts, domain adaptation (DA) is used to solve the problem of distribution alignment to further construct the spectral-spatial relationships for enhanced unsupervised HSIC \cite{Liuzx2021_TGRS}. Furthermore, the generative adversarial network (GAN) is introduced to train a deep learning-based feature extractor in an unsupervised manner \cite{Zhangmy2019_TGRS, Hangrl2020_TGRS, Yuwb2022_TGRS}. Although the above unsupervised feature learning frameworks can excavate implicit spectral-spatial relationships, GAN-based framework may lead to model collapse due to complex training procedure \cite{Thanh-Tung2020_IJCNN, xuhl2022_TGRS}.

Recently, the Denoising Diffusion Probabilistic Models (DDPMs) \cite{ho2020DDPM}, as unsupervised feature laerning approaches, have garnered significant research interest owing to their notable advantages over alternative approaches, including GAN-based methods \cite{Dhariwal2021_NIPS}. As generative models, DDPMs train a denoising autoencoder to learn the reverse of a Markovian diffusion process \cite{song2020improved}. Recently, DDPMs have demonstrated superiorities in vision tasks such as image generation \cite{ho2020DDPM, Rombach2022_CVPR, liu2022compositional, Yuanzq2023_TGRS}, image super-resolution \cite{Rombach2022_CVPR, Mengqy2023_TGRS, Chunghj2023_TMI}, segmentation \cite{baranchuk2022labelefficient, wu2023medsegdiff, kim2023diffusion, bandara2022ddpmcd, Graikos2022_NIPS, Brempong2022_CVPRW}, classification \cite{han2022card, yang2023diffmic, chen2023spectraldiff, zhou2023hyperspectral}. 

DDPMs have strong potential to improve HSIC by providing enhanced feature extraction from HSI due to the following reasons. First, DDPMs provide a principled probabilistic framework for feature learning, which can explicitly model the uncertainty and variability in HSI, and enable the extraction of improved spectral-spatial features for HSIC. Second, DDPMs can capture both high-level and low-level features that can enhance HSIC using proper time steps in DDPMs \cite{baranchuk2022labelefficient}. Third, DDPMs have more stable training mechanism and are less prone to model collapse compared to other deep unsupervised models, such as the GAN-based model \cite{baranchuk2022labelefficient}. Given the theoretical advantages of DDPMs, they have not been sufficiently studies for enhanced HSIC. Although some researchers have used DDPMs for HSIC \cite{bandara2022ddpmcd, chen2023spectraldiff, zhou2023hyperspectral}, there are still a lack of research on spatial-spectral DDPMs that are tailor designed to HSI characteristics and challengs. 

This paper therefore presents a new Spatial-Spectral Diffusion Contrastive Representation Network (DiffCRN), based on denoising diffusion probabilistic model (DDPM) combined with contrastive learning (CL) for HSIC, with the following contributions.

\begin{itemize}
\item[1)] To improve spatial-spectral feature representation in DiffCRN, instead of adopting the UNets-like structure which is widely used for DDPMs, we design a novel staged architecture with spatial self-attention denoising module (SSAD) and spectral group self-attention denoising module (SGSAD) in DiffCRN with improved efficiency for spectral-spatial feature learning.

\item[2)] To improve unsupervised feature learning efficiency, we design new spatial-spectral DiffCRN with logarithmic absolute error (LAE) loss and CL that improve the loss function effectiveness and increase the instance-level and inter-class discriminability.

\item[3)] To improve feature selection, we design a learnable approach based on pixel-level spectral angle mapping (SAM) for the selection of time steps in the proposed spatial-spectral DiffCRN in an adaptive and automatic manner.

\item[4)] To improve feature integration and classification, we design an Adaptive weighted addition modul (AWAM) and Cross time step Spectral-Spatial Fusion Module (CTSSFM) to fuse time-step-wise features and perform classification. 
\end{itemize}

Experiments conducted on widely used four HSI datasets demonstrate the improved performance of the proposed DiffCRN over the classical backbone models and state-of-the-art GAN, transformer models and other pretrained methods. The remainder of this paper is organized as follows. \autoref{section2} describes related work. In \autoref{section3}, our proposed DiffCRN is introduced in detail. \autoref{section4} discusses our experiment results. Some conclusions are drawn in \autoref{section6}.

\section{Related work} \label{section2}
In this section, we describe existing research directions relevant to our research based on traditional spectral-spatial feature extraction, as well as those based on deep learning, along with mainstream generative tasks and development context of DDPMs. Through a comprehensive review of related research, our aim is to establish a strong foundation for our proposed method, solidify our contribution, and apply it effectively in complex HSIC research.

\subsection{HSI classification based on supervised learning}

Traditional feature extraction approaches developed on statistical computations and linear algebra methods along with the canonical machine learning methods \cite{Prasad2008_TGRS, science2000_LLE, DU2003_PR}, which ignore the spatial relationships between each spectral pixels. Although researchers have proposed various methods which take spatial structure and texture information into consideration, e.g., the morphological profiles (MPs) \cite{Benediktsson2005_TGRS}, extended morphological profiles (EMPs) \cite{Plaza2005_TGRS_EMPS}, gray-level co-occurrence matrix (GLCM) \cite{Tsai2013_TGRS}, markov random field (MRF) \cite{Sun2015_TGRS}, invariant attribute profiles (IAPs) \cite{Hong2020_TGRS}. However, the above methods are criticized due to that the extracted shallow features lack representativeness, and the design of the hyperparameters are konwledge-driven.

During the last decade, the deep learning (DL) based methods have flourished in HSIC, many progressive DL-based data-driven networks have been widely utilized for supervised HSIC methods in an end-to-end manner, providing automatic feature learning from data \cite{chen_DBN2015_JSTARS, Li2019_TGRS, Audebert2019_TGRSM, Rasti2020_MGRS}, e.g., deep belief networks (DBNs) \cite{chen_DBN2015_JSTARS}. Particularly, HSIC benifits from CNNs-based models due to the ability to extract locally spatial contextual information and spectral variability information \cite{2DCNN, Zhong2017_IGRSS, Roy2020_LGRS}. Specifically, since the 3D convolution (3DCONV) can simultaneously investigate spectral-spatial relationships \cite{Hamida3dCNN, HybridSN}. For instance, Zhong et al. presented a spectral-spatial residual network (SSRN) \cite{SSRN}, which investigates the 3D spatial-spectral features by skip-connections to ensure the generalization ability. Furthermore, Tang et al. \cite{Tang2021_TGRS} factorized the mixed feature maps by their low and high-frequency information using a 3D Octave convolution model to realize the interaction of spectral-spatial information. However, these 3DCONV methods contain huge model parameters which require a large number of training samples \cite{Fu2023_TGRS}. Additionally, many researches from a sequential perspective with transformers \cite{Hongdf2022_TGRS, Ibañez2022_TGRS, SSTN} to design network for alleviating the deficiency of CNNs on mining and representing the sequence attributes of spectral signatures. For instance, Sun et al. \cite{SSFTT} constructed a spatial–spectral feature tokenization transformer (SSFTT) to capture spatial–spectral features and high-level semantic features. However, these works lack consideration of significant differences, such as the size and the number of basic elements between NLP and HSI. They still face quadratic computation complexities \cite{Liu2021_ICCV_Swin}, hence limited to modeling complex relations. According to \cite{park2022how}, their capture of long-range dependencies even hinders network optimization. 

\subsection{HSI classification based on unsupervised learning}

\subsubsection{\textbf{AE-based model}}
Unsupervised Feature extraction methods do not need labeled data and are not restricted to specific prerequisites. Typically, autoencoder (AE) is a common unsupervised framework. Kemker et al. \cite{Kemker2017_TGRS} proposed stacked convolutional autoencoder (SCAE) to identify deeper features and yielded the superior performance. \cite{Masci2011, Du2017_TCYB} used several autoencoders to learn a hierarchical feature representation, resulting in more discriminative features. The fully connected operators in AE were replaced by convolutional operators, so that the network can directly extract spectral–spatial joint features from cubes. Even so, they often involve complex training processes, which also needs numerous labeled samples in the classification stage.

\subsubsection{\textbf{GAN-based model}}
As a novel unsupervised classification scheme, the GANs have also been employed for HSIC. For instance, Zhan et al. \cite{Zhan2018_LGRS} developed a framework for HSIC using a 1-D GAN for HSIC (HSGAN) using discriminator features. Zhang et al. \cite{Zhangmy2019_TGRS} designed a deconvolutional generator and a 2-D CNN discriminator to learn spectral-spatial relationships of data sets and extract spatial-spectral features, respectively. Hang et al. \cite{Hangrl2020_TGRS} proposed a multitask generative adversarial network (MTGAN) which was consist of a generator network for hyperspectral imagery reconstruction and classification, and a discriminator network to discriminate between the real and fake (generated) data. Although GANs are powerful to energize unsupervised HSI classifiers, they always encounter the failures of model collapse and non-convergence. 

\subsection{Development of Denoising Diffusion Probabilistic Models}
DDPMs \cite{ho2020DDPM} have emerged as the new state-of-the-art family of deep generative models and have broken the long-time dominance of GANs, which has achieved great success in the field of natural image generation, natural language processing, temporal data modeling, multi-modal modeling. DDPMs are a family of probabilistic generative models that progressively destruct data by injecting noise, then reverse the former by learning transition kernels parameterized by deep neural networks. DDPMs are trained by optimizing the variational lower bound of the negative log-likelihood of the data, and it avoids mode collapse often encountered by GANs. Because of slow sample generation speed of DDPM, \cite{DDIM} presented denoising diffusion implicit models (DDIM), a class of non-Markovian diffusion processes, whose reverse process can be 10$\times$ to 50$\times$ faster than DDPMs. Dmitry et al. \cite{baranchuk2022labelefficient} demonstrated that the pretrained conditional DDPM can be served as a feature extractor for discriminative computer vision problems. To train a Neural Radiance Fields (NeRFs) with only a samll views, \cite{yang2023learning} found that unconditional diffusion prior can be treated as guidance for 3D generation and rendering. Furthermore, DDPMs have been successful on high-level semantic understanding tasks, i.e., image super resolution, natural and medical segmentation, classification. Although DDPMs have been widely applied to the community of natural image generation, and a small quantity on remote sensing fake sample generation \cite{Yuanzq2023_TGRS}, synthetic aperture radar despeckling \cite{Perera2023SARDDPM}, remote sensing change detection \cite{bandara2022ddpmcd}, seldom on the field of HSIC. Hence, the critical research question is how to adpat the latent feature representations of DDPMs for discriminative HSIC tasks, under the help of construction of raw HSI data distribution. Considering the theoretical advantages of DDPMs, some researchers have used DDPMs for HSIC \cite{bandara2022ddpmcd, chen2023spectraldiff, zhou2023hyperspectral}. However, there are still a lack of research on spatial-spectral DDPMs that are tailor designed to HSI characteristics and challenges. For example, efficient selection of time steps in DDPM in an adaptive learnable manner for HSIC has not been addressed, given the critical importance of time step selection \cite{baranchuk2022labelefficient, bandara2022ddpmcd, chen2023spectraldiff, zhou2023hyperspectral}. In \cite{bandara2022ddpmcd, chen2023spectraldiff}, the time steps are manually selected to characterize spectral-spatial features, and \cite{zhou2023hyperspectral} uses all of time steps, which might lead to redundancy.   

\section{Methodology} \label{section3}

\subsection{Problem Formulation}

\subsubsection{\textbf{Notation}} To facilitate reading and reduce ambiguity, we first define the common notation in \autoref{notation_tabel1}.

\begin{table}[htbp]
\Huge
\centering
\caption{A list of main notations used in this paper.}
\label{notation_tabel1}
\renewcommand{\arraystretch}{1.5}
\resizebox{0.5\textwidth}{!}{
\begin{tabular}{ll}
\toprule[4pt]
\Huge \textbf{Notations} & \Huge \textbf{Descriptions} \\ \hline
$H, W, C$ & \makecell[l]{height, width, number of spectral channels respectively} \\ \hline 
$\mathbfcal{X} \in \mathbb{R}^{H \times W \times C}$ & \makecell[l]{Hyperspectral image cube} \\ \hline
$\mathbfcal{H}_{\boldsymbol{0}} \in \mathbb{R}^{P \times P \times C}$ & \makecell[l]{cropped original spectral-spatial instance \\ using a $\emph{P} \times \emph{P}$ neighborhood} \\  \hline
\emph{T} & \makecell[l]{Total number Gaussian diffusion steps} \\ \hline
$\mathbfcal{H}_{\boldsymbol{t}} \in \mathbb{R}^{P \times P \times C}, t \in \{1, .., T\}$ &  \makecell[l]{Latent feature variable at the $t$-th timestep generated by \\ diffusion forward process, specially, $\mathbfcal{H}_{\boldsymbol{T}}$ is nearly \\ an isotropic Gaussian distribution}\\ \hline
$\boldsymbol{\epsilon}_{\theta}$ & \makecell[l]{Spatial-Spectral denoising network with parameters $\theta$} \\ \hline
$\mathbfcal{F}$ & Feature selection function \\ \hline
$\mathscr{A}_{\alpha}$ &  \makecell[l]{Contrastive Learning module $\mathscr{A}$ with parameters $\alpha$} \\ \hline
$\mathscr{C}_{\beta}$ &  \makecell[l]{Classification network $\mathscr{C}$ with parameters $\beta$} \\ \hline
$\boldsymbol{y}_i \in \{1,...,N\}, i \in \{1,.., \emph{H} \times \emph{W}\}$ & The class label of each pixel, $N$ represents the number of category \\ \toprule[4pt]
\end{tabular}
}
\end{table}

\subsubsection{\textbf{Spatial-Spectral Diffusion Representation Network}} Spatial-Spectral Diffusion Representation Network is latent variables model with the form \cite{Jascha2015, ho2020DDPM}
\begin{gather}
        p_{\theta}(\mathbfcal{H}_{\boldsymbol{0}}) \coloneqq \int p_{\theta} (\mathbfcal{H}_{\boldsymbol{0:T}}) d\mathbfcal{H}_{\boldsymbol{1:T}}\\
        p_{\theta} (\mathbfcal{H}_{\boldsymbol{0:T}}) \coloneqq p(\mathbfcal{H}_{\boldsymbol{T}}) \prod \limits_{t=1}^T p_{\theta}(\mathbfcal{H}_{\boldsymbol{t-1}}|\mathbfcal{H}_{\boldsymbol{t}}) \label{eq1}
\end{gather}
where $\mathbfcal{H}_{\boldsymbol{1}},...,\mathbfcal{H}_{\boldsymbol{t}},...,\mathbfcal{H}_{\boldsymbol{T}}$ are latents of the same dimensionality as the spectral-spatial instance sampled from a real HSI data distribution ${\mathbfcal{H}_{\boldsymbol{0}}} \sim {q(\mathbfcal{X})}$. The joint distribution $p_{\theta} (\mathbfcal{H}_{\boldsymbol{0:T}})$ is called the reverse process, and it is defined as a Markov chain with learned Gaussian transitions starting at $p(\mathbfcal{H}_{\boldsymbol{T}})=\mathcal{N}(\mathbfcal{H}_{\boldsymbol{T}};\boldsymbol{0},\boldsymbol{I})$. So, Spatial-Spectral Diffusion Representation Network enables to construct hyperspectral instance distribution with hundreds of spectrums by training the denoising network in the reverse process to estimate the noise added
in the spatial-spectral forward diffusion process, and then the desired image $\mathbfcal{H}_{\boldsymbol{0}}$ will be generated.

\begin{figure}[htbp]
    \centering
    \includegraphics[width=0.49\textwidth]{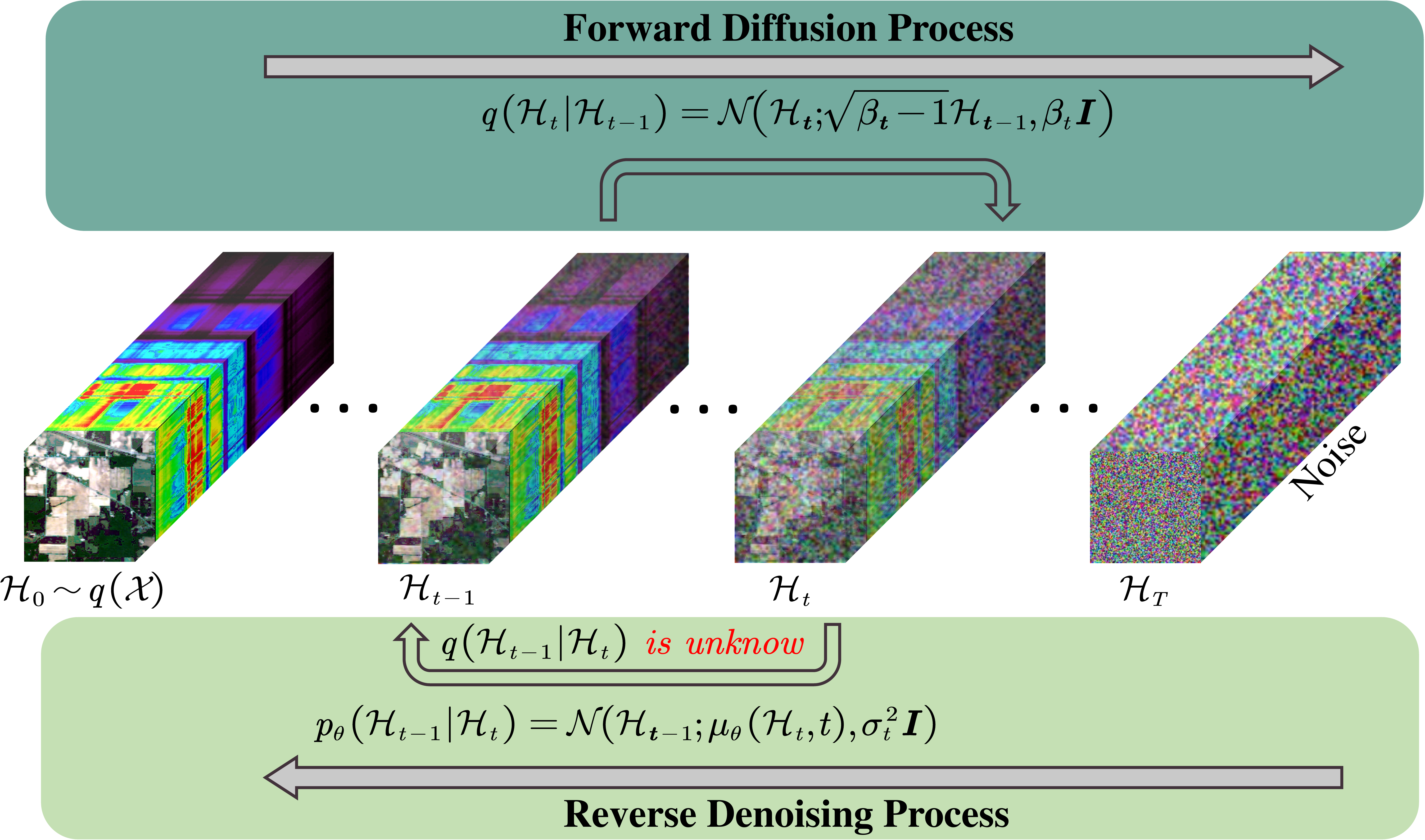}
    \caption{Denoising diffusion model forward and backward process.
$q(\mathbfcal{H}_{\boldsymbol{t}}|\mathbfcal{H}_{\boldsymbol{t-1}})$, $p_{\theta}(\mathbfcal{H}_{\boldsymbol{t-1}}|\mathbfcal{H}_{\boldsymbol{t}})$ represent noising adding forward process and
denoising backward process, respectively. The essential question is to estimate the conditional probability $q(\mathbfcal{H}_{\boldsymbol{t-1}}|\mathbfcal{H}_{\boldsymbol{t}})$.}
    \label{diffusion forward}
\end{figure}

\begin{figure*}[htbp]
    \centering
    \includegraphics[width=0.99\textwidth]{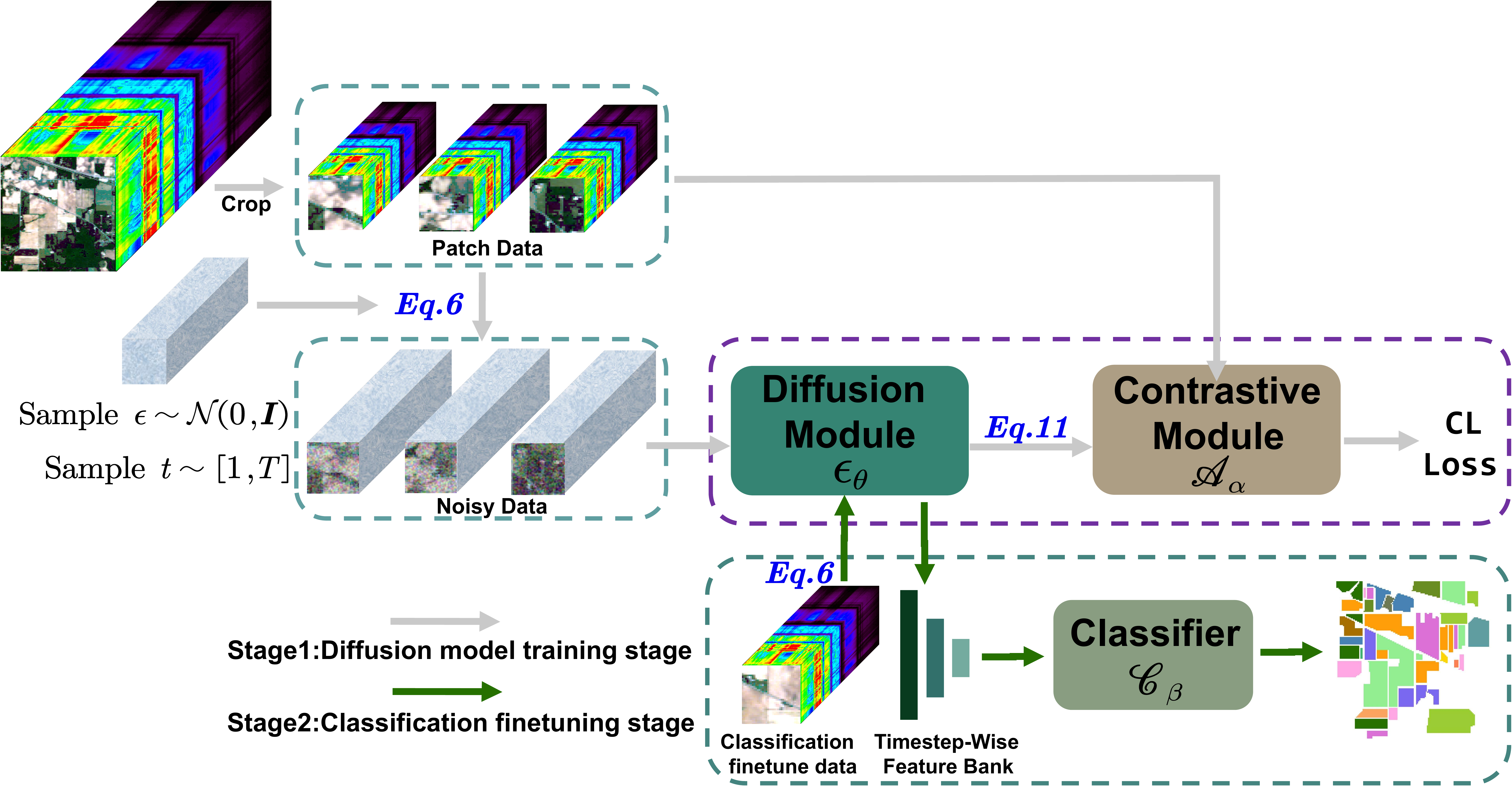}
    \caption{Overall framework of the proposed DiffCRN model. The method consists of two steps. \textbf{Step 1}: As show in gray line, we pretrain the diffusion adversarial representation network $\epsilon_{\theta}$ and $\mathscr{A}_{\alpha} $with cropped HSI patches in an unsupervised manner. \textbf{Step 2}: As show in green line, another cropped HSI patches are feeding into the pretrained diffusion adversarial representation network which parameters are freezed, then, the feature from different stages on specific time step $t$ are extracted. The extracted features on the same stage but different $t$ are weighted add. Finally, a classifier $\mathscr{C}_{\beta}$ is trained by using the five-timestep representations of limit labelled data to perform HSI classification.}
    \label{overall_architecture}
\end{figure*} 

\subsubsubsection{\textbf{Spatial-Spectral Diffusion Forward Process}} As mentioned above, by performing variational inference on a Markovian process with \emph{T} number of timesteps, diffusion models can convert the Gaussian noise distribution $\mathbfcal{H}_{\boldsymbol{T}} \sim \mathcal{N}(\boldsymbol{0},\boldsymbol{I})$ into the target distribution of $\mathbfcal{H}_{\boldsymbol{0}}$. 

The spatial-spectral diffusion forward process is inspired by non-equilibrium thermodynamics which can be viewed as a Markov chain that the noise is gradually added to $\mathbfcal{H}_{\boldsymbol{0}}$ to produce latent variables $\mathbfcal{H}_{\boldsymbol{1}}$ through $\mathbfcal{H}_{\boldsymbol{T}}$ at time \emph{t} with fixed variance schedule $\beta_t$, At timestep \emph{t}, the noisy spectral-spatial instance $\mathbfcal{H}_{\boldsymbol{t}}$ can be represented as follows:
\begin{gather} q(\mathbfcal{H}_{\boldsymbol{1}},...,\mathbfcal{H}_{\boldsymbol{T}}|\mathbfcal{H}_{\boldsymbol{0}}) \coloneqq \prod \limits_{t=1}^T q(\mathbfcal{H}_{\boldsymbol{t}}|\mathbfcal{H}_{\boldsymbol{t-1}}) \\
    q(\mathbfcal{H}_{\boldsymbol{t}}|\mathbfcal{H}_{\boldsymbol{t-1}}) \coloneqq \mathcal{N}(\mathbfcal{H}_{\boldsymbol{t}};\sqrt{\beta_{t}-1}\mathbfcal{H}_{\boldsymbol{t-1}}, \beta_{t}\boldsymbol{I}) \label{eq3}
\end{gather}
Through the reparameterization trick, the hyperspectral instance $\mathbfcal{H}_{\boldsymbol{t}}$ can be sampled directly at an arbitrary time step \emph{t} in closed form, the marginal distribution of intermediate latent variables $\mathbfcal{H}_{\boldsymbol{t}}$ given $\mathbfcal{H}_{\boldsymbol{0}}$ can be derived as:
\begin{gather}
q(\mathbfcal{H}_{\boldsymbol{t}}|\mathbfcal{H}_{\boldsymbol{0}}) = \mathcal{N}(\mathbfcal{H}_{\boldsymbol{t}};\sqrt{\bar{\alpha_{t}}}\mathbfcal{H}_{\boldsymbol{0}},(1-\bar{\alpha_{t}})\boldsymbol{I})\\ 
\mathbfcal{H}_{\boldsymbol{t}} = \sqrt{\bar{\alpha_{t}}}\mathbfcal{H}_{\boldsymbol{0}}+\sqrt{1-\bar{\alpha_{t}}}\epsilon \label{eq5}
\end{gather}
Here, $\alpha_{t} \coloneqq  1-\beta_{t}$, $\bar{\alpha_{t}} \coloneqq  \prod \limits_{s=1}^t \alpha_{s}$, $\epsilon \sim \mathcal{N}(0, \boldsymbol{I})$. In \autoref{eq3}, the magnitude of the added Gaussian noise decreases as the value of $\alpha_{t}$ increases, which also means that the informative of $\mathbfcal{H}_{\boldsymbol{0}}$ decrease as the time step \emph{t} increases in \autoref{eq5}.
\subsubsubsection{\textbf{Spatial-Spectral Diffusion Reverse Process}} The spatial-spectral diffusion reverse process, modeled by a neural network $\boldsymbol{\epsilon}_{\theta}$, aims to learn to
remove the degradation brought from the diffusion forward process, i.e. denoising. In each timestamp $t$ of the reverse process, the denoising operation is performed on the noisy spectral-spatial instance $\mathbfcal{H}_{\boldsymbol{t}}$ to obtain the previous instance $\mathbfcal{H}_{\boldsymbol{t-1}}$. In \autoref{eq1}, since $p(\mathbfcal{H}_{\boldsymbol{t-1}}|\mathbfcal{H}_{\boldsymbol{t}})$ is unknown and depends on the data distribution, we approximate it using a neural network $\boldsymbol{\epsilon}_{\theta}$ to predict the mean $\mu_{\theta}(\mathbfcal{H}_{\boldsymbol{t}}, t)$ of a Gaussian distribution as follows:
\begin{gather}
    p_{\theta}(\mathbfcal{H}_{\boldsymbol{t-1}}|\mathbfcal{H}_{\boldsymbol{t}}) \coloneqq \mathcal{N}(\mathbfcal{H}_{\boldsymbol{t-1}}; \mu_{\theta}(\mathbfcal{H}_{\boldsymbol{t}}, t), \sigma_{t}^{2}\boldsymbol{I})
    \label{eq8}
\end{gather}
where $\sigma_{t}^{2}$ is the variance of the conditional distribution $p(\mathbfcal{H}_{\boldsymbol{t-1}}|\mathbfcal{H}_{\boldsymbol{t}})$, following:
\begin{gather}
    \sigma_{t}^{2} = \frac{1-\bar{\alpha}_{t-1}}{1-\bar{\alpha}_{t}} \beta_{t}
\end{gather}
As reported by Ho et al. \cite{ho2020DDPM}, the learning objective for the above model is derived by considering the variational lower bound, and then, the mean $\mu_{\theta}(\mathbfcal{H}_{\boldsymbol{t}}, t)$ of the conditional distribution $p(\mathbfcal{H}_{\boldsymbol{t-1}}|\mathbfcal{H}_{\boldsymbol{t}})$ can be reparameterized
as follows:
\begin{gather}
    \mu_{\theta}(\mathbfcal{H}_{\boldsymbol{t}}, t) = \frac{1}{\sqrt{\alpha_{t}}}(\mathbfcal{H}_{\boldsymbol{t}} - \frac{\beta_{t}}{\sqrt{1-\bar{\alpha}_{t}}} \boldsymbol{\epsilon}_{\theta}(\mathbfcal{H}_{\boldsymbol{t}}, t))
\end{gather}
where $\boldsymbol{\epsilon}_{\theta}$ is the denoising network intended to predict added gaussion noise $\boldsymbol{\epsilon}$ from the noisy hyperspectral instance $\mathbfcal{H}_{\boldsymbol{t}}$ at time step \emph{t}. The forward and reverse processes are illustrated in \autoref{diffusion forward}.

\subsubsubsection{\textbf{Sampling}} To sample $\mathbfcal{H}_{\boldsymbol{t-1}} \sim p_{\theta}(\mathbfcal{H}_{\boldsymbol{t-1}}|\mathbfcal{H}_{\boldsymbol{t}})$ is to iteratively compute 
\begin{gather}
    \mathbfcal{H}_{\boldsymbol{t-1}} = \frac{1}{\sqrt{\alpha_{t}}}(\mathbfcal{H}_{\boldsymbol{t}} - \frac{\beta_{t}}{\sqrt{1-\bar{\alpha}_{t}}} \boldsymbol{\epsilon}_{\theta}(\mathbfcal{H}_{\boldsymbol{t}}, t)) + \sigma_{t} \boldsymbol{z}
    \label{sampling}
\end{gather}
where $\boldsymbol{z} \sim \mathcal{N}(\boldsymbol{0}, \boldsymbol{I})$, \emph{t = T,...,1}. 

According to Tweedie’s formula \cite{chung2023diffusion}, we can use $\hat{\mathbfcal{H}_{\boldsymbol{0}}} = \mathbfcal{H}_{0|t} \triangleq \mathbb{E} [\mathbfcal{H}_{0}| \mathbfcal{H}_{t}]$ to represents the estimation of $\mathbfcal{H}_{0}$ at timestep $t$ given $\mathbfcal{H}_{t}$, as follows:
\begin{gather}
   \hat{\mathbfcal{H}_{\boldsymbol{0}}} \simeq \frac{1}{\sqrt{\bar{\alpha}_{t}}}
   (\mathbfcal{H}_{\boldsymbol{t}} + (1-\bar{\alpha}_{t}) \boldsymbol{\epsilon}_{\theta}(\mathbfcal{H}_{\boldsymbol{t}}, t))
    \label{eq12}
\end{gather}

\begin{figure*}[htbp]
    \centering
    \includegraphics[width=0.99\textwidth]{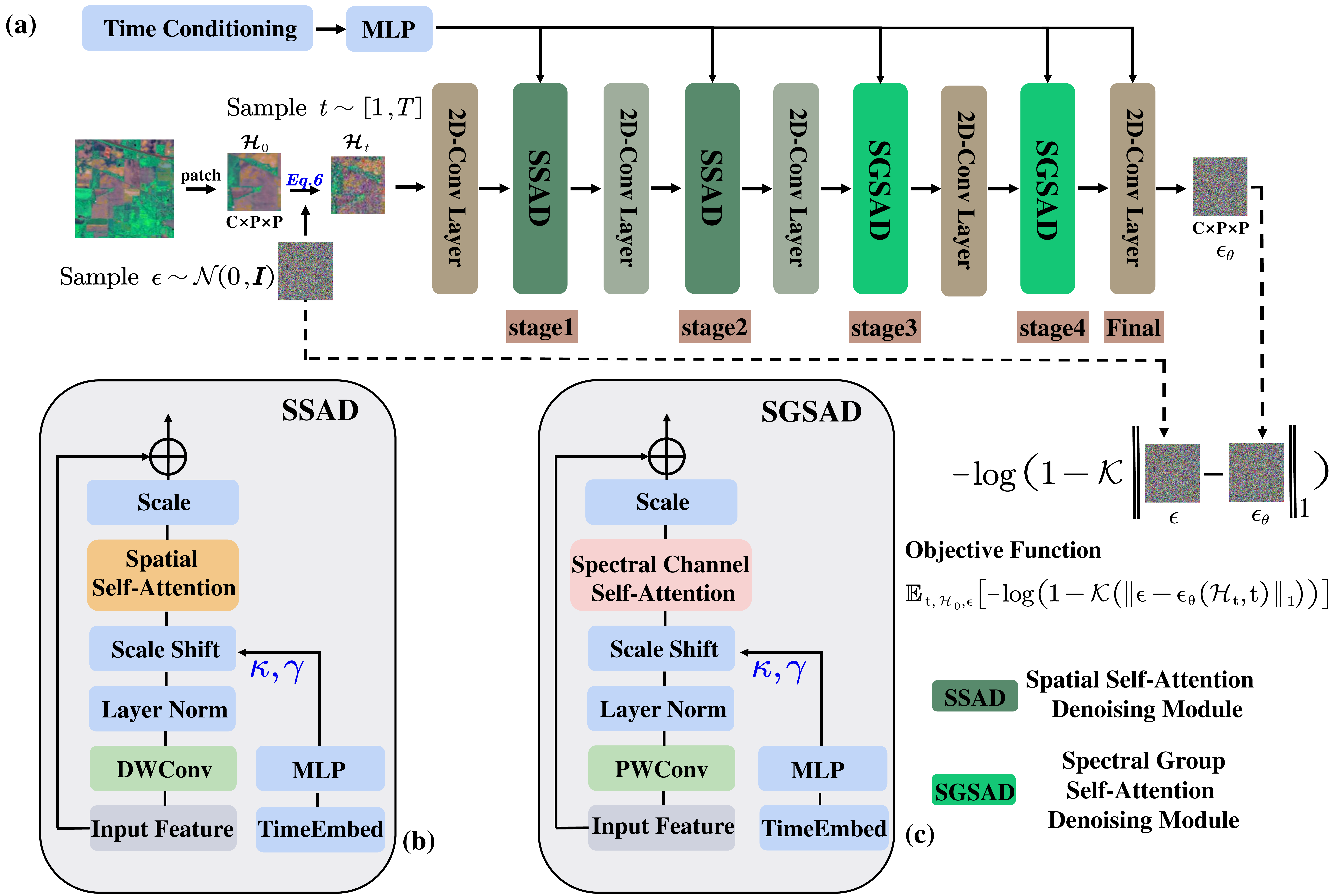}
    \caption{Stage architecture of the Spatial-Spectral Denoising Adversarial Representation Learning Network. Timestep embeddings is generated by MLP. DWConv and PWConv represent depthwise convolution and pointwise convolution respectively. (a) The network consists of two SSAD Modules and two SGSAD Modules. (b) The details of Spatial Self-Attention Denoising Module (SSAD) (c) The details of the Spectral Group Self-Attention Denoising Module (SGSAD).}
    \label{architecture}
\end{figure*}

\subsection{Architecture and Key Components of Spatial-Spectral Diffusion Contrastive Representation Network} 
\subsubsection{\textbf{Overall Architecture}}
Unlike U-Nets-like backbone architecture of the diffusion models such as \cite{bandara2022ddpmcd, chen2023spectraldiff, Perera2023SARDDPM, cao2023ddrf}, we introduce stage architecture with spatial-spectral dual self-attention denoising module, a novel architecture for DDPMs. \autoref{overall_architecture} shows an overview of the DiffCRN. In the \autoref{overall_architecture}, the proposed model consists of two stages. In the first stage, i.e., spatial-spectral diffusion contrastive representation part, the unlabeled training data $\{\mathbfcal{T}_{i}, \mathbfcal{H}_{\boldsymbol{0}}^{i}\}_{i=1}^{N}$ in training set $\mathbfcal{T}$ is fed into DifCRN after forward diffusion process \autoref{eq5}, then estimate \(\epsilon\) using network \(\boldsymbol{\epsilon}_{\theta}(\mathbfcal{H}_{\boldsymbol{t}}, t)\). According to \autoref{eq12}, we can roughly reconstruct $\hat{\mathbfcal{H}_{\boldsymbol{0}}}$, then, the pairs $\{\mathbfcal{H}_{\boldsymbol{0}},\hat{\mathbfcal{H}_{\boldsymbol{0}}}\}$ be regarded as two views and then fed into contrastive learning module $\mathscr{A}_{\alpha}$ for contrastive learning. 

\subsubsection{\textbf{Spatial Self-Attention Denosing Module (SSAD)}}

In addition to noisy hyperspectral instance embedding feature inputs, diffusion models sometimes process additional conditional information. Here, we consider the noise timesteps \emph{t}. The noisy hyperspectral instance $\mathbfcal{H}_{\boldsymbol{t}} \in \mathbb{R}^{C \times P \times P}$is firstly embeded by a 2D convolution layer with layer norm and GELU activation which is denoted as $\mathscr{F}(\mathbfcal{H}_{\boldsymbol{t}}^{C \times P \times P})$. For better denoising in spatial domain, and approximate information and distribution of spatial features, we intorduce depthwise convolution and spatial self-attention module (SSA) into SSAD, as shown in Figure \autoref{architecture}. We can express this process using the following equations:
\begin{align}
\begin{aligned}
     &\mathcal{A} = \text{LN}(\text{DWConv}(\mathscr{F}(\mathbfcal{H}_{\boldsymbol{t}}^{C \times P \times P}))) \\
     &\hat{\mathcal{A}} = \mathcal{A} \cdot (1 + \gamma) + \kappa \\
     &\hat{\mathcal{A}} = \mathscr{F}(\mathbfcal{H}_{\boldsymbol{t}}^{C \times P \times P}) + LayerScale \cdot SSA(\hat{\mathcal{A}})
\end{aligned}
\end{align}
where, $\gamma, \kappa$ are the corresponding scale and shift based on the input timestep conditioning to modulate $\mathcal{A}$, which can be represented as:
\begin{align}
    \begin{aligned}
        \gamma, \kappa = \mathsf{Chunk}(\text{MLP}(\text{TimeEmbeddings}))
    \end{aligned}
\end{align}
where $\mathsf{Chunk}$ is splitting the feature into two parts with equal
size along the channel dimension. \text{MLP} is implemented as a SiLU activation followed by a linear layer. Sinusoidal TimeEmbeddings is created as follows:
\begin{align}
    \text{TimeEmbeddings} = [\cos (\frac{t}{10000^{\frac{2i}{d}}}), \sin (\frac{t}{10000^{\frac{2i}{d}}})]
\end{align}
where, $t \in \mathbb{R}^{B}$, $B$ represents the batch size of timestep, $i=0,...,\frac{d}{2} - 1$, $d$ represents the dimension of embedding.
$[ \cdot ]$ represents the concatenation operation.
Spatial Self-Attention module (SSA) computes self-attention within local patch, as shown in \autoref{SSA-SGSA}. Then SSA can be represented by:
\begin{align}
\begin{aligned}
    SSA(\mathbfcal{Q}, \mathbfcal{K}, \mathbfcal{V}) &= Attention(\mathbfcal{Q}, \mathbfcal{K}, \mathbfcal{V}) = \sigma(\frac{\mathbfcal{Q} \mathbfcal{K}^{T}}{\sqrt{C}}) \mathbfcal{V}
\end{aligned}
\end{align}
where $\mathbfcal{Q}=\mathbfcal{X}\mathbfcal{W}^{Q}, \mathbfcal{K}=\mathbfcal{X}\mathbfcal{W}^{K}, \mathbfcal{V}=\mathbfcal{X}\mathbfcal{W}^{V} \in \mathbb{R}^{C\times P \times P}$ are patch queries, keys, and values. $\mathbfcal{W}^{Q}, \mathbfcal{W}^{K}, \mathbfcal{W}^{V}$ represent the projection weights for $\mathbfcal{Q},\mathbfcal{K},\mathbfcal{V}$. $\mathbfcal{X}, C$ represents the inputs and total number of spectral domain respectively. \(\sigma\) indicates \textit{Softmax} function.

\subsubsection{\textbf{Spectral Group Self-Attention Denosing Module (SGSAD)}}

The output of the second SSAD will firstly fedded into a 2D convolution layer then the SGSAD. SGSAD is a well-designed block for better denoising in spectral domain, and approximate information and distribution of spectral features. we intorduce pointwise convolution and spectral group self-attention (SGSA) into SGSAD, as shown in \autoref{architecture}. We can express this process using the following equations:
\begin{align}
\begin{aligned}
     &\mathcal{B} = \text{LN}(\text{PWConv}(\mathscr{F}(\mathbfcal{S}_{\boldsymbol{t}}^{C \times P \times P}))) \\
     &\hat{\mathcal{B}} = \mathcal{B} \cdot (1 + \gamma) + \kappa \\
     &\hat{\mathcal{B}} = \mathscr{F}(\mathbfcal{S}_{\boldsymbol{t}}^{C \times P \times P}) + LayerScale \cdot SGSA(\hat{\mathcal{B}})
\end{aligned}
\end{align}
where, $\mathbfcal{S}_{\boldsymbol{t}}^{C \times P \times P}$ is the output of the second spatial denoising module at \emph{t} time step. \autoref{architecture} indicates the detailed stage architecture of Spatial-Spectral Denoising Network.

\begin{figure}[h]
    \centering
    \includegraphics[width=0.49\textwidth]{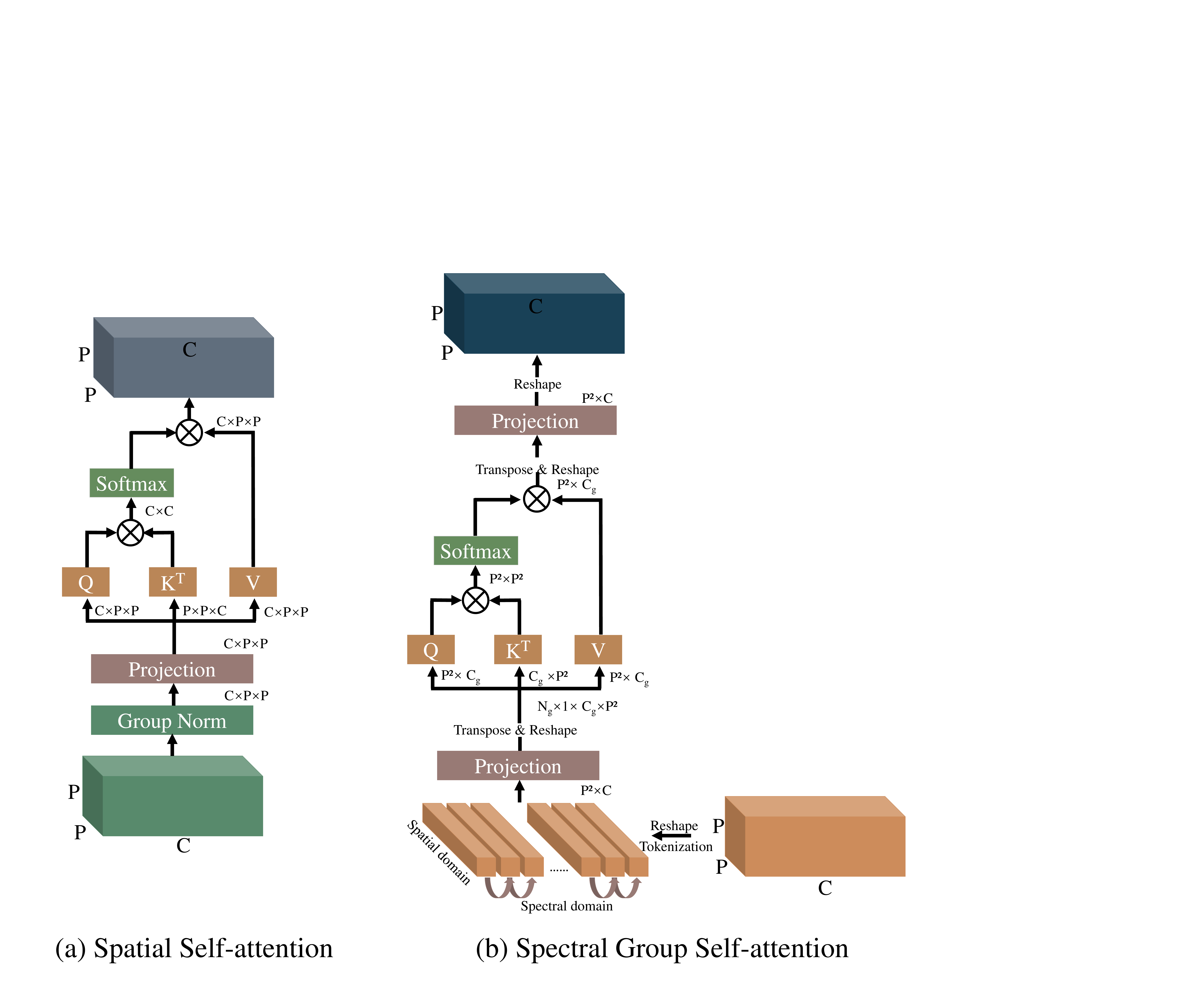}
    \caption{(a) Spatial Self-Attention module (SSA) computes self-attention within local patch. (b) Spectral Group Self-Attention module (SGSA) groups spectral tokens into multi groups. Attention is performed in each spectral group with an entire patch-level spectral as a token. In this work, we introduce SSA into SSAD to obtain local feature, and introduce SGSA into SGSAD to obtain global features. $P$, $C$, $N_g$, $C_g$ represent patch size, number of spectral domian, number of spectral groups and spectrums per group respectively.}
    \label{SSA-SGSA}
\end{figure}

In SGSA, we apply attention mechanisms on the transpose of patch-level tokens, the diemsion change form $\mathbb{R}^{C \times P \times P}$ to $\mathbb{R}^{P^{2} \times C}$, $P$ indicates the patch size. So, spectral tokens in SGSA interact with global spatial information on the spectral dimension. To reduce the computational complexity, we group spectral domain into $N_g$ groups and perform self-attention within each group, each group with dimension $C_g$. In this way, our spectral group attention is global, with patch-level tokens interacting across a group of channels, as shown in \autoref{SSA-SGSA}. Then SGSA can be represented by:
\begin{align}
\begin{aligned}
    SGSA(\mathbfcal{Q}, \mathbfcal{K}, \mathbfcal{V}) &= [head_{1}, head_{2}, ..., head_{N_g}]\\
    \text{where} \qquad head_{i} &= SGSA_{group}(\mathbfcal{Q}_i, \mathbfcal{K}_i, \mathbfcal{V}_i) = \sigma(\frac{\mathbfcal{Q}_i \mathbfcal{K}_{i}^{T}}{\sqrt{C_g}}) \mathbfcal{V}_{i}
\end{aligned}
\end{align}
where $\mathbfcal{Q}_i, \mathbfcal{K}_i, \mathbfcal{V}_i \in \mathbb{R}^{P^2 \times C_g}$ are grouped channel-wise patch-level queries, keys, and values.

\subsubsection{\textbf{Contrastive Learning Module}}

\begin{figure}[htbp]
    \centering
    \includegraphics[width=0.49\textwidth]{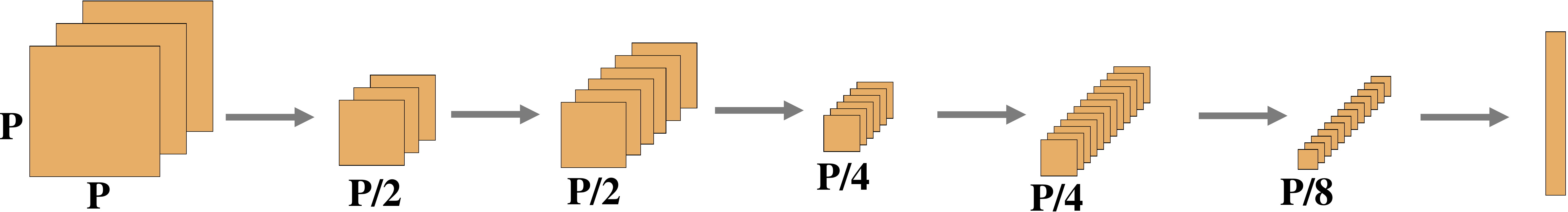}
    \caption{Contrastive Learning Module in DiffCRN.}
    \label{Discriminator}
\end{figure}

\autoref{Discriminator} shows the structure of the contrastive learning module in DiffCRN, which is comprised of six convolutional layers and a linear layer. Each convolutional layer includes a 2D convolution layer with 3 × 3 convolutional kernels, a layer normalization and a GELU activation. The numbers of kernels from the first to the sixth convolution layer are correspondingly designed as 96, 96, 128, 128, 256, and 256. The contrastive learning module regularly learn spatial downsampling and increase the feature dimension. The inputs of contrastive learning module are reconstructed $\hat{\mathbfcal{H}_{\boldsymbol{0}}}$ by \autoref{eq12} and raw hyperspectral instance $\mathbfcal{H}_{\boldsymbol{0}}$. After feeding inputs into six convolutional layers, an average pooling operation is adopted before linear layer. 

\subsection{Spatial-Spectral Diffusion Contrastive Representation Network Optimizing}
Now we need to define the form of the composite loss function, which contains three parts. Here, we explain as follows:

\subsubsection{\textbf{Spatial-Spectral Diffusion Loss}}
By maximizing its variational lower bound and it can turn into a simple supervised loss, i.e., Mean Squared Error (MSE), which is the preferred loss function in DDPMs \cite{ho2020DDPM}, i.e., \autoref{eq10}, also in some current works, including \cite{bandara2022ddpmcd, chen2023spectraldiff, Perera2023SARDDPM,wu2023medsegdiffv2, yang2023diffmic, kim2023diffusion, asiedu2022decoder}
\begin{gather}
    L_{simple}(\theta) = \mathbb{E}_{t, \mathbfcal{H}_{\boldsymbol{0}}, \boldsymbol{\epsilon}} \left[ ||\boldsymbol{\epsilon} - \boldsymbol{\epsilon}_{\theta}(\mathbfcal{H}_{\boldsymbol{t}}, t)||^{2}\right]
    \label{eq10}
\end{gather}
where \emph{t} is uniform between 1 and \emph{T}. While, the MSE treats all the predicted noise uniformly. However, due to the variance schedule, the noise level in diffusion forward process at an arbitrary time step \emph{t} is quite different. Inspired by \cite{Li2023}, we introduce a nonuniform loss function for diffusion model, named logarithmic absolute error (LAE), thus, we modify the \autoref{eq10} as follows:
\begin{gather}
    \mathbfcal{L}_{diff} = \mathbb{E}_{t, \mathbfcal{H}_{\boldsymbol{0}}, \boldsymbol{\epsilon}} [-\log (1 - \mathbfcal{K}(\left\| \boldsymbol{\epsilon} - \boldsymbol{\epsilon}_{\theta}(\mathbfcal{H}_{\boldsymbol{t}},t)\right\|_{1}))]
    \label{eq18}
\end{gather} 
Here, $\mathbfcal{K}$ is a clipping function, i.e., $\mathcal{K}(\cdot) \in [0,0.9999)$. The reason for adopting such a logarithm loss function is to put more emphasis on heavily penalizing the big gap between $\boldsymbol{\epsilon}$ and $\boldsymbol{\epsilon}_{\theta}(\mathbfcal{H}_{\boldsymbol{t}},t)$. In the following ablation study, we verify the effectiveness of LAE.

\subsubsection{\textbf{Reconstruction Loss}}
In view of the good properties of DDPMs in mathematical derivation, according to \autoref{eq12}, we can reconstruct $\hat{\mathbfcal{H}_{\boldsymbol{0}}}$ roughly. We use the similar form of \autoref{eq18} to measure the distance of $\hat{\mathbfcal{H}_{\boldsymbol{0}}}$ and $\mathbfcal{H}_{\boldsymbol{0}}$, which can be defined as:
\begin{gather}
    \mathbfcal{L}_{rec} = \mathbb{E}_{t, \mathbfcal{H}_{\boldsymbol{0}}, \boldsymbol{\epsilon}} [-\log (1 - \mathbfcal{K}(\left\| \hat{\mathbfcal{H}_{\boldsymbol{0}}} - \mathbfcal{H}_{\boldsymbol{0}}\right\|_{1}))]
    \label{eq_rec}
\end{gather}

\subsubsection{\textbf{Contrastive Learning Loss}}
Contrastive learning is a prominent approach in self-supervised learning that aims to learn useful representations from unlabelled data \cite{SimCLR}. Accordingly, in DiffCRN, the contrastive loss function is designed to maximize similarity between fast sampled fake hyperspectral imagery \(\hat{\mathbfcal{H}_{\boldsymbol{0}}}\) through \autoref{eq12} and the raw hyperspectral instance \(\mathbfcal{H}_{\boldsymbol{0}}\) while maximizing difference for two distinct samples. So, the \(\hat{\mathbfcal{H}_{\boldsymbol{0}}} \in {\mathbb{R}^ {B \times C \times H \times W}}\)  and 
 \(\mathbfcal{H}_{\boldsymbol{0}} \in {\mathbb{R}^ {B \times C \times H \times W}}\) are treated as two views for the given training sample. For the contrastive leaning task, \(\hat{\mathbfcal{H}_{\boldsymbol{0}}}\) and \(\mathbfcal{H}_{\boldsymbol{0}}\) are fed into the Contrastive Learning Module to get the latent feature \(\mathbf{F}_{\hat{\mathbfcal{H}_{\boldsymbol{0}}}}^{'} \in {\mathbb{R}^ {B \times 256}}\) and \(\mathbf{F}_{\mathbfcal{H}_{\boldsymbol{0}}} \in {\mathbb{R}^ {B \times 256}}\). A particularly successful contrastive loss, named InfoNCE, is adopted \cite{SimCLR}. It takes the following form:
 
\begin{align}
    \begin{aligned}
        \mathbfcal{L}_{con} = &\frac{1}{2B} \underset{k=1}{\overset{B}{\varSigma}} (\ell (2k-1, 2k) + \ell(2k, 2k-1)) \\
        \ell(i,j) = &-\log \frac{e^{\text{sim}(\mathbf{F}_{\hat{\mathbfcal{H}_{\boldsymbol{0}}}}^{'[i]}, \mathbf{F}_{\mathbfcal{H}_{\boldsymbol{0}}}^{[j]})/\tau}}{\underset{k=1}{\overset{2B}{\varSigma}} \mathds{1}_{[k \neq i]} e^{\text{sim}(\mathbf{F}_{\hat{\mathbfcal{H}_{\boldsymbol{0}}}}^{ '[i]}, \mathbf{F}_{\mathbfcal{H}_{\boldsymbol{0}}}^{[k]})/\tau}}
    \end{aligned}
\end{align}

Where \(\mathbf{F}_{\hat{\mathbfcal{H}_{\boldsymbol{0}}}}^{'[i]}\) and \(\mathbf{F}_{\mathbfcal{H}_{\boldsymbol{0}}}^{[j]}\) are two views
for the \(i\)th and \(j\)th sample. Here \text{sim}: \(\mathbb{R}^{d} \rightarrow \mathbb{R}\) is a similarity metric (e.g., cosine similarity). \(\tau\) is a hyperparameter controls the magnitude of the loss. \(B\) is the number of training samples in one mini-batch, and \(\mathds{1}_{[k \neq i]}\) is an indicator function, 1 if \(k \neq i\) and 0 otherwise.

\subsubsection{\textbf{Compound Loss Function}} Obviously, the proposed framework DiffCRN is a multitask learning network. It is necessary to define a weighted sum of these four loss considering the contributions of different loss components rather than using predefined or manually tuned weights. Inspired by \cite{Cipolla2018multitask}, we define the compound loss based on homoscedastic uncertainty of each task, which can learn various quantities and units in each task. We define as follows:
\begin{align}
    \begin{aligned}
    \mathbfcal{L} &= e^{-w_{diff}} \mathbfcal{L}_{diff} +  e^{-w_{rec}} \mathbfcal{L}_{rec} + 0.5 \mathbfcal{L}_{con} \\ &+ w_{diff} + w_{rec}
    \end{aligned}
\end{align}

\begin{figure}[htbp]
    \centering
    \includegraphics[width=0.49\textwidth]{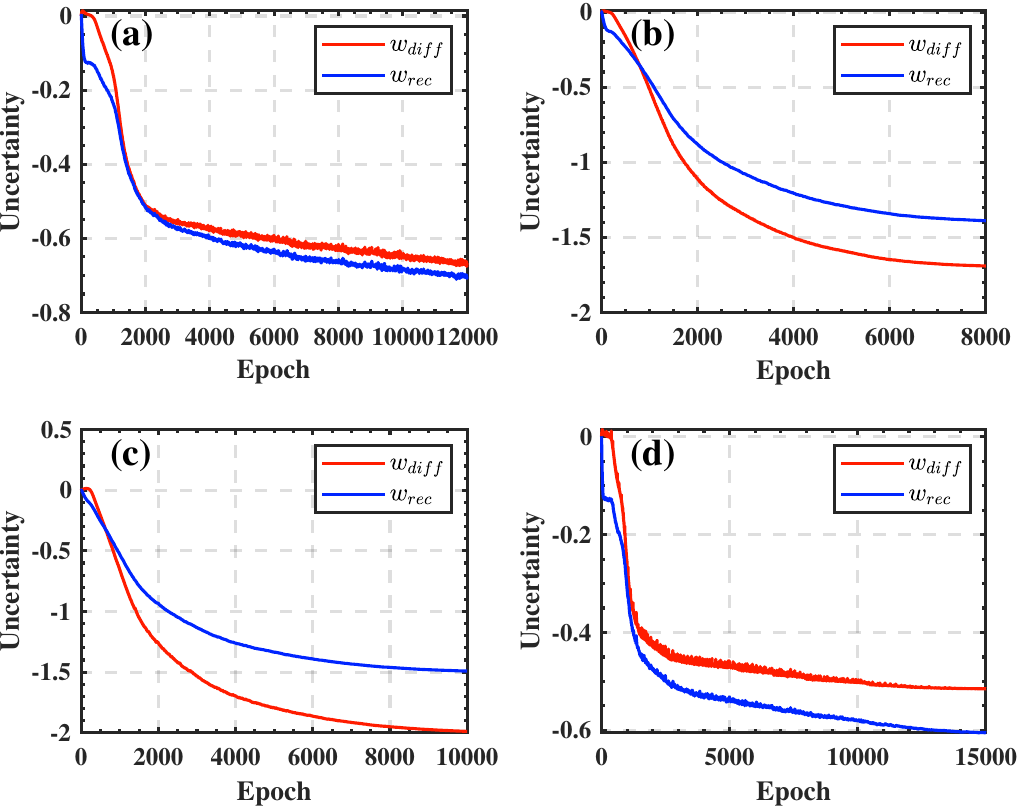}
    \caption{Uncertainty over spatial-spectral diffusion process loss (red) and reconstruction loss (blue). The task uncertainty converges as the increasing of epoch. (a) Indian pines (c) Pavia University (c) MUUFL (d) WHU-Hi-HongHu}
    \label{uncertainty}
\end{figure}

Where $w_{diff}=\log v_{diff}$, $w_{rec} = \log v_{rec}$, $v_{diff}$, $v_{rec}$ are the noise variance representing the uncertainty of the reconstruction task of \autoref{eq18} and \autoref{eq_rec}. During pretrainig the spatial-spectral diffusion model, we set $w_{diff}$ and $w_{rec}$ as the learning parameter of the model, and initialize they to 0. The last two terms $w_{diff}$ and $w_{rec}$ act as a regularizer to avoid overfitting. Notably, \autoref{uncertainty} shows the changes of \(w_{diff}\) and \(w_{rec}\) on specific dataset, i.e., task uncertainty. We can see that, as the increasing of epoch, \(w_{diff}\) and \(w_{rec}\) gradually become smaller and converges, that means the weight of $\mathbfcal{L}_{diff}$ and $\mathbfcal{L}_{rec}$ become bigger, resulting the main optimization object.

\subsubsection{\textbf{Pixel-Level Spectral Angle Mapping (SAM)}}
Unlike \cite{bandara2022ddpmcd, chen2023spectraldiff}, which taking feature representations of diffusion model at the manually tuned time steps, then feeding into classifier, and \cite{zhou2023hyperspectral} uses all of time steps, which might lead to redundancy. For dynamic adjustment and decision, it is essential to adjust which time step should be used iteratively during the phase of training. Hence, we propose a simple but efficient methodology based on learnable SAM to measure the similarity on pixel level, is given as:
\begin{align}
    &\mathbf{SAM}(\hat{\mathbfcal{H}_{\boldsymbol{0}}}, \mathbfcal{H}_{\boldsymbol{0}}) = \cos ^{-1} (\frac{<\hat{\mathbfcal{H}_{\boldsymbol{0}}} , \mathbfcal{H}_{\boldsymbol{0}}>}{\left\| \hat{\mathbfcal{H}_{\boldsymbol{0}}} \right\|_{2} \cdot \left\|  \mathbfcal{H}_{\boldsymbol{0}} \right\|_{2}} )\label{eq19}
\end{align}
where, $\hat{\mathbfcal{H}_{\boldsymbol{0}}} \in \mathbb{R}^{B \times C \times P \times P}$, $\mathbfcal{H}_{\boldsymbol{0}} \in \mathbb{R}^{B \times C \times P \times P}, \mathbf{SAM} \in \mathbb{R}^{B \times P \times P}$, here, $B$ is the number of hyperspectral instance, each instance corresponds to a time step \emph{t}. The smaller the SAM, the higher the similarity. The mean value of SAM can be obtained along the \(B\) dimension. In the following experiments, we take the time step corresponding to the top five most similar of SAM after averaging. In the following ablation study, we verify the effectiveness of SAM.

\subsection{Classification with Spectral-Spatial Diffusion Features}

\subsubsection{\textbf{Extracting Intermediate Representations From DiffCRN}}
\begin{figure}[htbp]
    \centering
    \includegraphics[width=0.49\textwidth]{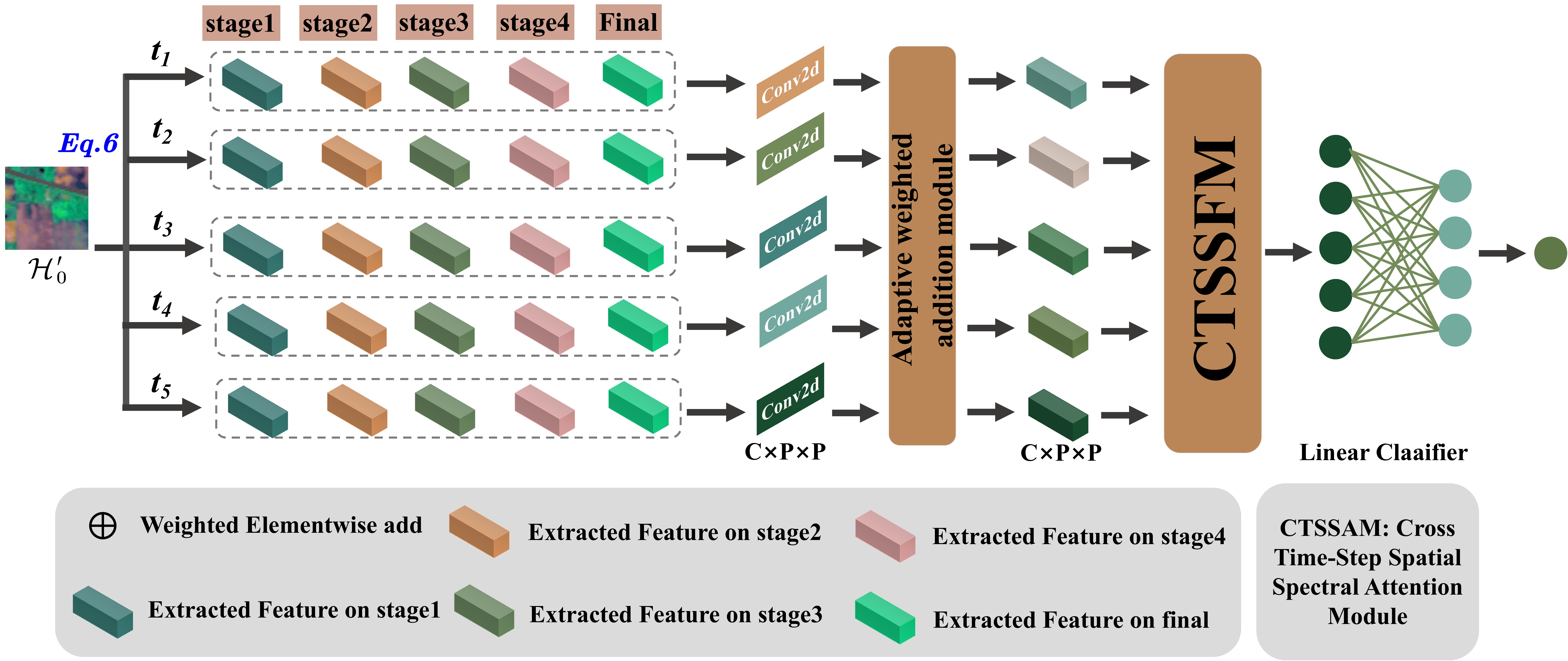}
    \caption{Classifier with spectral-spatial diffusion feature as inputs. The input data of classifier is the extracted features from diffusion model on specific time step, here we choose five time steps. Firstly, a convolutional layer with kernel size of 1 is applied to the extracted features to change to dimension. Then, the features maps are fed into AWAM and CTSSFM followed by a linear classifier to perform HSIC.}
    \label{Classifier_architecture}
\end{figure}

Given a hyperspectral instance, we first extract the intermediate feature representations from the pre-trained spectral-spatial diffusion model and feed these features as the inputs to the classifier $\mathscr{C}_{\beta}$, as shown in \autoref{Classifier_architecture}. To achieve multi-level representation, we retain the features from five different time step $t$, which is selected from top five the smallest value of $\mathbf{SAM}$. The above process can be expressed as:
\begin{align}
\begin{aligned}
    &\chi_{i} = \text{Conv2d}(\mathbfcal{F(\boldsymbol{\epsilon}_{\theta}}(\mathbfcal{H}_{\boldsymbol{t}, t})))\\
    &\tilde{\psi}=\text{CTSSFM}(\underset{s=1}{\overset{5}{\varSigma}}\text{AWAM}(\underset{j=1}{\overset{5}{\varSigma}}\chi_{j}))\\
    &\hat{y} = \text{FC}(\text{GAP}(\tilde{\psi}))
\end{aligned}
\end{align}
where, $\mathbfcal{F}(\cdot)$ is the extracted feature representations on different time step $t$. $\tilde{\psi}$ means the concatenated features. $s$ represents the stage index in DiffCRN, as shown in \autoref{architecture} (i.e., stage1, stage2, stage3, stage4, final), $j$ represents the j-th time index. $\hat{y}$ represents the predicted label corresponding to the hyperspectral instance.

\subsubsection{\textbf{Adaptive weighted addition module (AWAM)}}
The AWAM performs channel attention by exploring the inter-channel relationships present within the features, as shown in \autoref{module_in_Classifier}. Given a feature map \(F \in \mathbb{R}^{C \times H \times W}\), which is the element-wise addition of extracted feature maps from the same stage but different diffusion time step, as input. AWAM infers 1D channel attention map \(M(F) \in \mathbb{R}^{C \times 1 \times 1}\). The refined feature map \(F' \in \mathbb{R}^{C \times H \times W}\) is computed as:
\begin{gather}
    F' = (F \otimes M(F)) \otimes \text{Conv}_{1\times1}(F \otimes M(F))
\end{gather}
where \(\otimes\) denotes elements-wise multiplication. In \(M(F)\),
average pooling (AvgPool) and max pooling (MaxPool) operations are employed to aggregate spatial information. Subsequently, features are fed into a shared multi-layer perceptron (MLP) activated by sigmoid function. The feature map is obtained by combining the outputs of the shared MLP through element-wise summation. Then, 1D channel attention map \(M(F) \in \mathbb{R}^{C \times 1 \times 1}\) is broadcast along the spatial dimension to obtain the refined feature. 1D channel attention map \(M(F) \in \mathbb{R}^{C \times 1 \times 1}\) is calculated as follows:
\begin{gather}
    M(F) = \sigma (\text{MLP}(\text{AvgPool}(F)) + \text{MLP}(\text{MaxPool}(F)))
\end{gather}

\subsubsection{\textbf{Cross Time Step Spectral-Spatial Fusion Module (CTSSFM)}}
\begin{figure}[htbp]
    \centering
    \includegraphics[width=0.49\textwidth]{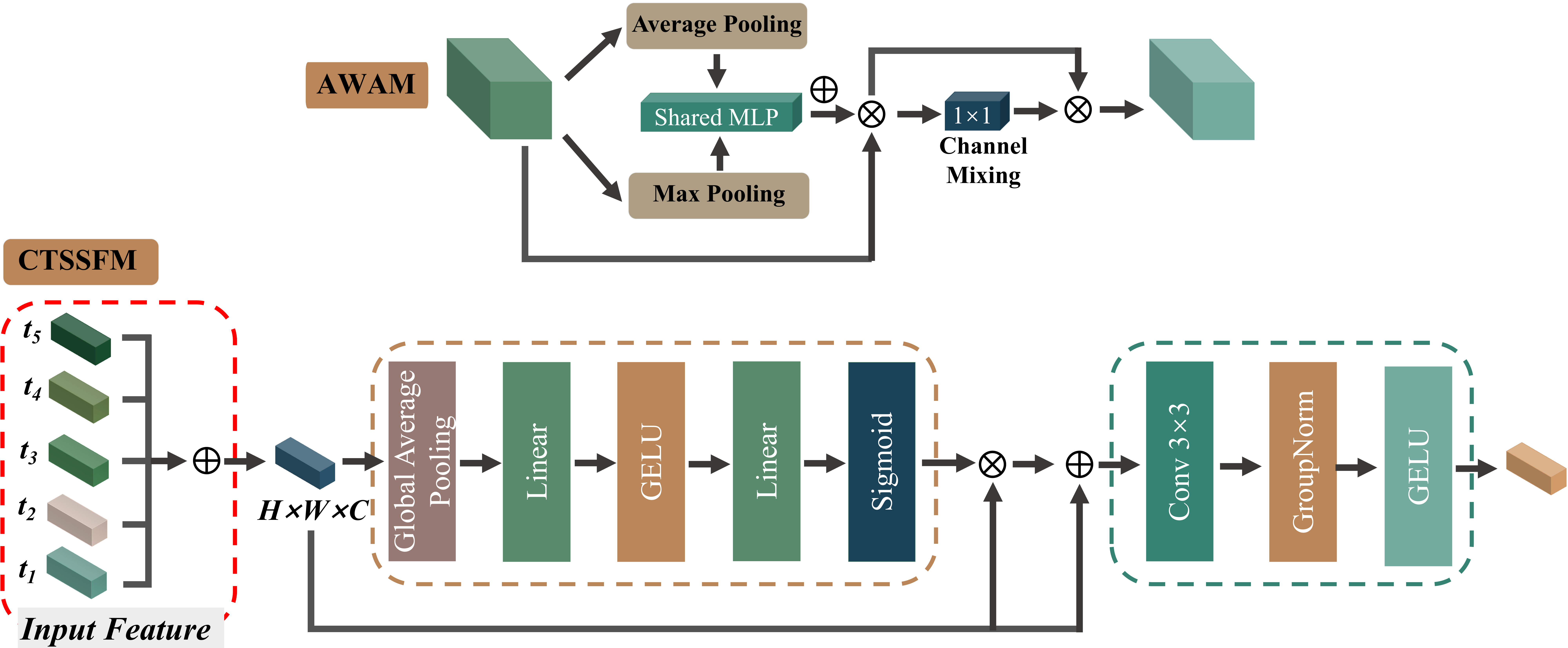}
    \caption{\textbf{Top}: Adaptive weighted addition module (AWAM). The input data of AWAM is sum of extracted of feature on the same stage but different time step after Conv1$\times$1. \textbf{Bottom}: Cross time step Spectral-Spatial Fusion Module (CTSSFM). The input feature of CTSSFM is the element-wise summation of the output of AWAM.}
    \label{module_in_Classifier}
\end{figure}

The detailed structure of CTSSFM is illustrated in \autoref{module_in_Classifier}. The input of CTSSFM is the element-wise summation of refined feature maps from different stages and diffusion time step, which is differ from AWAM. We use \(F \in \mathbb{R}^{C \times H \times W}\) to represent the sum of five given feature maps \(F_{i} \in \mathbb{R}^{C \times H \times W}, i \in 1,...,5\). CTSSFM performs
channel attention and spatial attention sequentially, then a 3D attention map \(M(F) \in \mathbb{R}^{C \times H \times W}\) is refined, follows:
\begin{gather}
    M(F) = \mathscr{F}_{spa}(F \oplus (\mathscr{F}_{spe}(F) \otimes F))
\end{gather}
where \(\oplus\) denotes elements-wise summation. We use global average pooling (GAP) in the spectral attention module \(\mathscr{F}_{spe}\) to get feature descriptor on the global spatial information. Subsequently, a one-dimensional convolution module is used to extract the channels of interest. Finally, we get the spectral attention vector an sigmoid function. In summary, \(\mathscr{F}_{spe}\) is calculated as follows:
\begin{gather}
    \mathscr{F}_{spe} = \sigma (\text{GAP}(\text{Conv}_{1 \times 1}(F)))
\end{gather}

In spatial attention module \(\mathscr{F}_{spa}\), fisrtly, we apply a \(3\times3\) convolution to maintain a large receptive field to aggregate the spatial context information, then a GroupNorm and GELU activation function are used. The calculation process is expressed as follows:
\begin{gather}
    \mathscr{F}_{spa} = \text{GELU}(\text{GruopNorm}(\text{Conv}_{3 \times 3}(*))
\end{gather}
where * is the output of \(\mathscr{F}_{spe}\). In the following ablation study, we verify the effectiveness of AWAM and CTSSFM.

\subsubsection{\textbf{Classification Loss}}
In the second stage, i.e., classification task, we adopt the cross-entropy loss, which can be written as
\begin{align}
    \begin{aligned}
        \mathbfcal{L}_{cls} &= \frac{1}{N} \underset{i=1}{\overset{N}{\varSigma}} [
        y_{i} \log \mathscr{C}_{\beta}(\mathbfcal{F}(\boldsymbol{\epsilon}_{\theta}(\mathbfcal{H}_{\boldsymbol{t}}, t))) \\& - (1-y_{i}) \log \mathscr{C}_{\beta}(1 - \mathbfcal{F}(\boldsymbol{\epsilon}_{\theta}(\mathbfcal{H}_{\boldsymbol{t}}, t))) ]
    \end{aligned}
\end{align}
where $N$ denotes the number of labeled cubes, $y_{i}$ denotes the class label of the $i$th cube in classification training set $\mathbfcal{L}$. 

\section{Experimental results and analysis} \label{section4}

\subsection{Datasets Description}

To evaluate the performance of the proposed method, four classical HSI datasets are adopated, i.e., Indian Pines (IN)\footnote{\url{https://www.ehu.eus/ccwintco/index.php/Hyperspectral_Remote_Sensing_Scenes}}, Pavia University (PU) \footnote{\url{https://www.ehu.eus/ccwintco/index.php/Hyperspectral_Remote_Sensing_Scenes}}, WHU-Hi-HongHu\footnote{\url{http://rsidea.whu.edu.cn/resource_WHUHi_sharing.htm}}, and MUUFL\footnote{\url{https://github.com/GatorSense/MUUFLGulfport}}.

\begin{figure}[htbp]
\centering
\subfigure[Indian Pines]{
\includegraphics[width=0.22\textwidth]{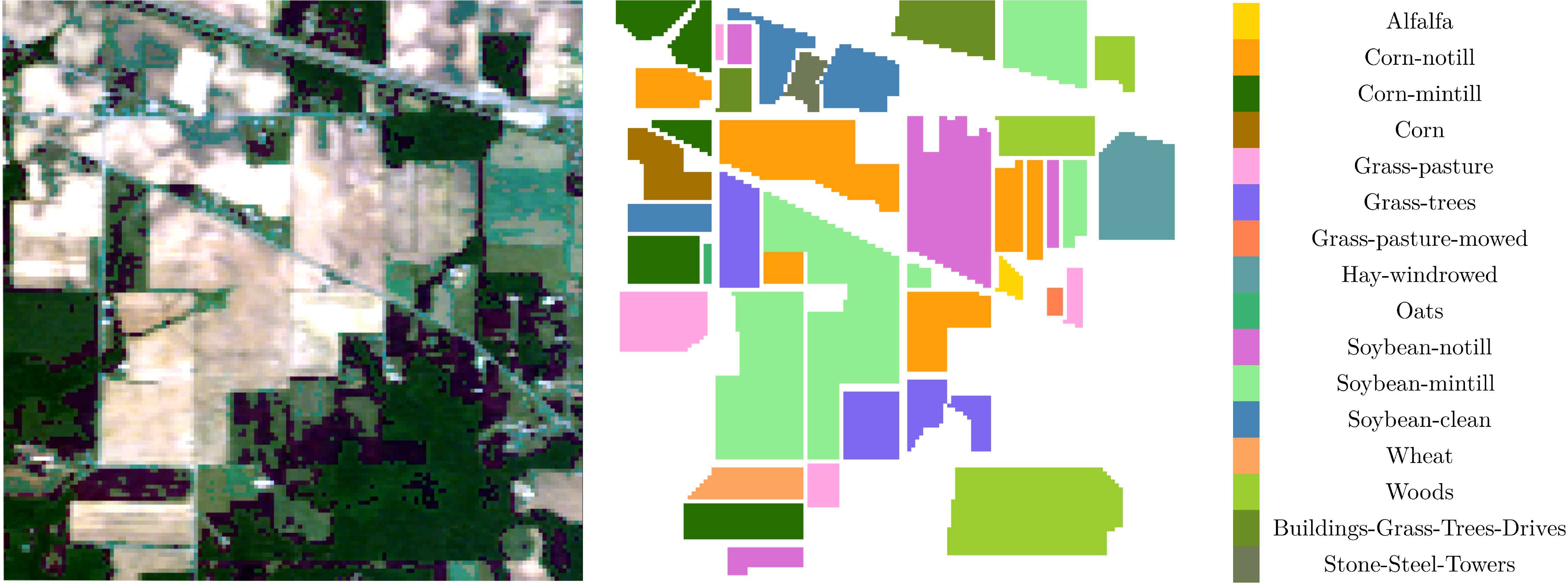}
}
\subfigure[Pavia University]{
\includegraphics[width=0.22\textwidth]{figures/dataset_brief/PU_brief.pdf}
}
\subfigure[WHU-Hi-HongHu]{
\includegraphics[width=0.22\textwidth]{figures/dataset_brief/HongHu_brief.pdf}
}
\subfigure[MUUFL]{
\includegraphics[width=0.22\textwidth]{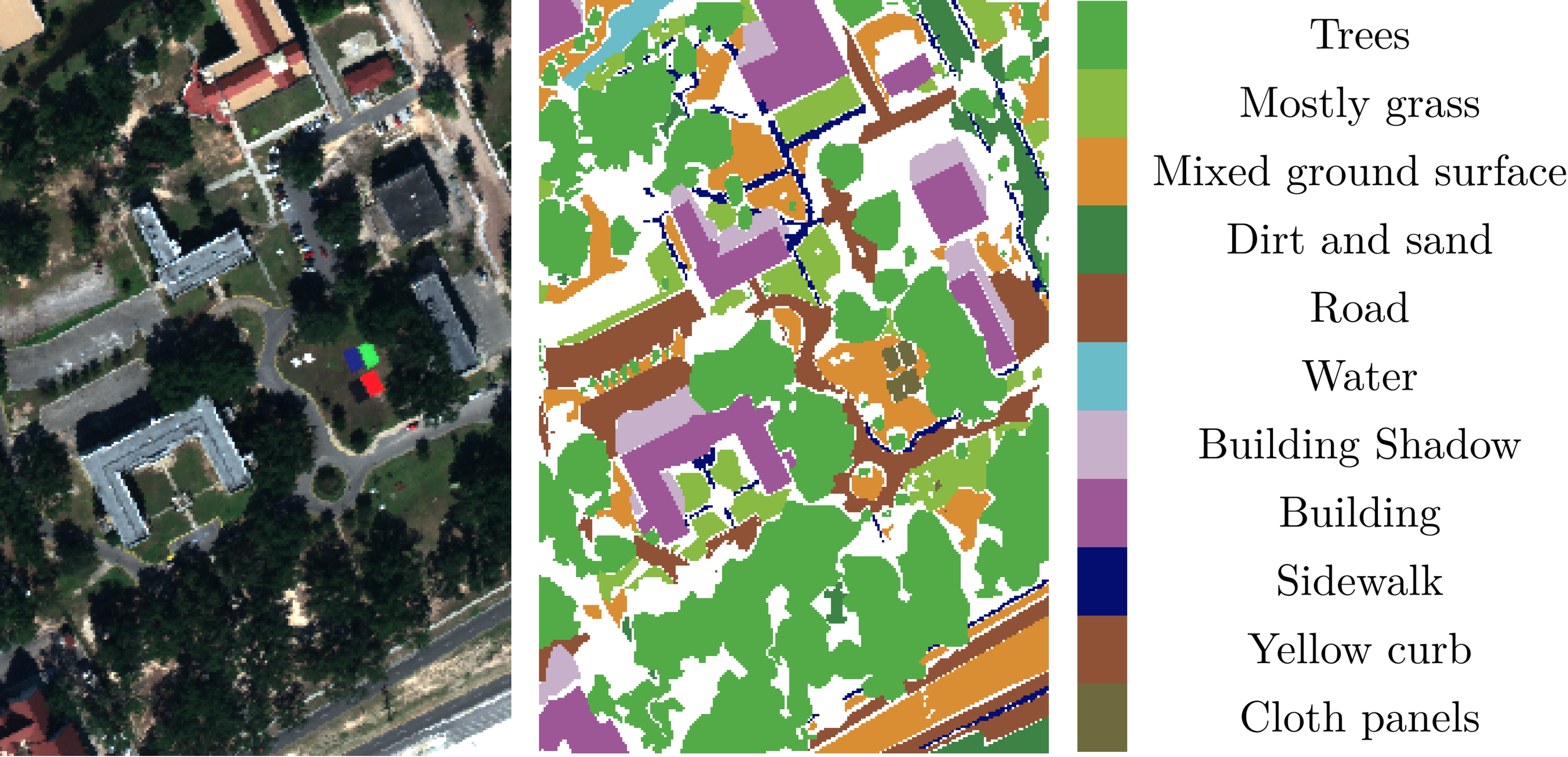}
}
\caption{The pseudocolor image, the corresponding ground-truth map and the legend of the four datasets.}
\label{dataset}
\end{figure}

\subsubsection{\textbf{Indian Pines Data}}  This dataset was collected by the AVIRIS sensor over Northwestern Indiana, USA. This data consists of 145 × 145 pixels at a ground sampling distance (GSD) of 20 m and 220 spectral
bands covering the wavelength range of 400–2500 nm with a
10-m spectral resolution. In the experiment, 24 water-absorption
bands and noise bands were removed, and 200 bands were
selected. There are 16 mainly investigated categories in this studied scene. The class name and the number of training and testing are listed in \autoref{tabel_sample_num}. Exp1 and Exp2 represents 10 \% of the samples and 20 samples of per class are randomly selected for training, the resting for testing, respectively. \autoref{dataset} (a) shows the false-color map and ground-truth map.

\subsubsection{\textbf{Pavia University Data}} It was acquired by the ROSIS sensor over Pavia University and its surroundings, Pavia, Italy. This dataset has 103 spectral bands ranging from 430 to 860 nm. Its spatial resolution is 1.3 m, and its image size is 610 × 340. Nine land-cover categories are covered. The class name and the number of training and testing are listed in \autoref{tabel_sample_num}. Exp1 and Exp2 represents 3 \% of the samples and 20 samples of per class are randomly selected for training, the resting for testing, respectively. \autoref{dataset} (b) shows the false-color map and ground-truth map.

\subsubsection{\textbf{WHU-Hi-HongHu Data}} This dataset was acquired on November 20, 2017, by Headwall Nano-Hyperspec imaging sensor equipped on a DJI Matrice 600 Pro UAV platform over the area of Honghu City, Hubei province, China, with a spatial resolution of 0.043 m, image size of 940 $\times$ 475, and 270 bands in the range of from 400 to 1000 nm. 22 land-cover categories are covered. The class name and the number of training and testing are listed in \autoref{tabel_sample_num}. Exp1 and Exp2 represents 0.05 \% of the samples and 20 samples of per class are randomly selected for training, the resting for testing, respectively. \autoref{dataset} (c) shows the false-color map and ground-truth map.

\subsubsection{\textbf{MUUFL Data}} It was obtained in November 2010 around the region of the University of Southern Mississippi Gulf Park, Long Beach, MS, USA, by using the ROSIS sensor. It is made up of 325 × 220 pixels along with 72 spectral bands. The first and last eight bands are deleted owing to noise, resulting in 64 bands in total. There are 53687 ground-truth pixels with 11 different types of classes for urban land cover. The class name and the number of training and testing are listed in \autoref{tabel_sample_num}. Exp1 and Exp2 represents 3 \% of the samples and 20 samples of per class are randomly selected for training, the resting for testing, respectively. \autoref{dataset} (d) shows the false-color map and ground-truth map. 

\begin{table*}[htbp]
\Huge
\centering
\caption{LAND-COVER TYPES, TRAINING AND TESTING SAMPLE NUMBERS OF THE INDIAN PINES DATASET, THE PAVIA UNIVERSITY DATASET, THE WHU-HI-HONGHU DATASET AND THE MUUFL DATASET.}
\resizebox{\textwidth}{!}{
\begin{threeparttable}
\begin{tabular}{cccc|ccc|ccc|ccc}
\hline
\multicolumn{1}{c|}{\multirow{3}{*}{No.}} &
  \multicolumn{3}{c|}{Indian Pines} &
  \multicolumn{3}{c|}{Pavia University} &
  \multicolumn{3}{c|}{WHU-Hi-HongHu} &
  \multicolumn{3}{c}{MUUFL} \\ \cline{2-13} 
\multicolumn{1}{c|}{} &
  \multirow{2}{*}{Class} &
  Training &
  Testing &
  \multirow{2}{*}{Class} &
  Training &
  Testing &
  \multirow{2}{*}{Class} &
  Training &
  Testing &
  \multirow{2}{*}{Class} &
  Training &
  Testing \\
\multicolumn{1}{c|}{} &
   &
  Exp1/Exp2\tnote{1} &
  Exp1/Exp2 &
   &
  Exp1/Exp2 &
  Exp1/Exp2 &
   &
  Exp1/Exp2 &
  Exp1/Exp2 &
   &
  Exp1/Exp2 &
  Exp1/Exp2 \\ \hline \specialrule{0em}{3pt}{3pt} \hline
\multicolumn{1}{c|}{1} &
  Alfalfa &
  5/20 &
  41/26 &
  Asphlat &
  199/20 &
  6432/6611 &
  Red roof &
  70/50 &
  13971/13991 &
  Trees &
  697/20 &
  22549/23226 \\
\multicolumn{1}{c|}{2} &
  Corn-notill &
  143/20 &
  1285/1408 &
  Meadows &
  559/20 &
  18090/18629 &
  Road &
  18/50 &
  3494/3462 &
  Mostlt grass &
  128/20 &
  4142/4250 \\
\multicolumn{1}{c|}{3} &
  Corn-mintill &
  83/20 &
  747/810 &
  Gravel &
  63/20 &
  2036/2079 &
  Bare soil &
  109/50 &
  21712/21771 &
  Mixed ground surface &
  206/20 &
  6676/6862 \\
\multicolumn{1}{c|}{4} &
  Corn &
  24/20 &
  213/217 &
  Trees &
  92/20 &
  2972/3044 &
  Cotton &
  816/50 &
  162469/163235 &
  Dirt and sand &
  55/20 &
  1771/1806 \\
\multicolumn{1}{c|}{5} &
  Grass-pasture &
  48/20 &
  435/463 &
  Painted metal sheets &
  40/20 &
  1305/1325 &
  Cotton firewood &
  31/50 &
  61876168 &
  Road &
  201/20 &
  6486/6667 \\
\multicolumn{1}{c|}{6} &
  Grass-trees &
  73/20 &
  657/710 &
  Bare Soil &
  151/20 &
  4878/5009 &
  Rape &
  223/50 &
  44334/44507 &
  Water &
  14/20 &
  452/446 \\
\multicolumn{1}{c|}{7} &
  Grass-pasture-mowed &
  3/20 &
  25/8 &
  Bitumen &
  40/20 &
  1290/1310 &
  Chinese cabbage &
  121/50 &
  23982/24053 &
  Building Shadow &
  67/20 &
  2166/2213 \\
\multicolumn{1}{c|}{8} &
  Hay-windrowed &
  48/20 &
  430/458 &
  Self-Blocking Bricks &
  110/20 &
  3572/3662 &
  Pakchoi &
  20/50 &
  4034/4004 &
  Building &
  187/20 &
  6053/6220 \\
\multicolumn{1}{c|}{9} &
  Oats &
  2/15 &
  18/5 &
  Shadows &
  28/20 &
  919/972 &
  Cabbage &
  54/50 &
  10765/10769 &
  Sidewalk &
  42/20 &
  1343/1365 \\
\multicolumn{1}{c|}{10} &
  Soybean-notill &
  97/20 &
  875/952 &
   &
   &
   &
  Tuber mustard &
  62/50 &
  12332/8904 &
  Yellow curb &
  5/20 &
  178/163 \\
\multicolumn{1}{c|}{11} &
  Soybean-mintill &
  246/20 &
  2209/2435 &
   &
   &
   &
  Brassica parachinensis &
  55/50 &
  10960/10965 &
  Cloth panels &
  8/20 &
  261/249 \\
\multicolumn{1}{c|}{12} &
  Soybean-clean &
  59/20 &
  534/573 &
   &
   &
   &
  Brassica chinensis &
  45/50 &
  8909/8904 &
   &
   &
   \\
\multicolumn{1}{c|}{13} &
  Wheat &
  21/20 &
  184/185 &
   &
   &
   &
  Small Brassica chinensis &
  113/50 &
  22394/22457 &
   &
   &
   \\
\multicolumn{1}{c|}{14} &
  Woods &
  127/20 &
  1138/1245 &
   &
   &
   &
  Lactuca sativa &
  37/50 &
  7319/7212 &
   &
   &
   \\
\multicolumn{1}{c|}{15} &
  Building-Grass-Trees-Drives &
  39/20 &
  347/366 &
   &
   &
   &
  Celtuce &
  5/50 &
  997/952 &
   &
   &
   \\
\multicolumn{1}{c|}{16} &
  Stone-Steel-Towers &
  9/20 &
  84/73 &
   &
   &
   &
  Film covered lettuce &
  36/50 &
  7226/7212 &
   &
   &
   \\
\multicolumn{1}{c|}{17} &
   &
   &
   &
   &
   &
   &
  Romaine lettuce &
  15/50 &
  2995/2960 &
   &
   &
   \\
\multicolumn{1}{c|}{18} &
   &
   &
   &
   &
   &
   &
  Carrot &
  16/50 &
  3201/3167 &
   &
   &
   \\
\multicolumn{1}{c|}{19} &
   &
   &
   &
   &
   &
   &
  White radish &
  44/50 &
  8668/8662 &
   &
   &
   \\
\multicolumn{1}{c|}{20} &
   &
   &
   &
   &
   &
   &
  Garlic sprout &
  17/50 &
  3469/3436 &
   &
   &
   \\
\multicolumn{1}{c|}{21} &
   &
   &
   &
   &
   &
   &
  Broad bean &
  7/50 &
  1321/1278 &
   &
   &
   \\
\multicolumn{1}{c|}{22} &
   &
   &
   &
   &
   &
   &
  Tree &
  20/50 &
  4020/3990 &
   &
   &
   \\ \hline \specialrule{0em}{3pt}{3pt} \hline
\multicolumn{2}{c}{Total} &
  1027/315 &
  9222/9934 &
   &
  1282/180 &
  41494/42596 &
   &
  1934/1100 &
  384759/385593 &
   &
  1610/220 &
  52077/53467 \\ \hline
\end{tabular}
\begin{tablenotes}
        \footnotesize
        \Huge
        \item[1] EXP1/EXP2 means two experiments with different number of training samples. In EXP2, we choose SSRN, MTGAN, SSFTT, and SC-SS-MTr as comparative methods.
\end{tablenotes}

\end{threeparttable}
}
\label{tabel_sample_num}
\end{table*}

\subsection{Experimental Settings}

\subsubsection{\textbf{Implementation Details and Training Details}}
Our proposed DiffCRN model is implemented on the PyTorch 1.10.2 platform using a workstation with Intel(R) Xeon(R) CPU E5-2640 v4, 256-GB RAM, and an NVIDIA GeForce RTX 2080 Ti 11-GB GPU. Training, and testing samples are extracted as 3D cubes. The Adam optimizer is adopted with learning rate of 1e-4. In the pre-training stage, we set batch size of 1024, 640, 480, 640 and patch size of 7, 9, 11, 9 for Indian pines, Pavia University, WHU-Hi-HongHu and MUUFL dataset respectively. The number of epochs is set to 12000, 8000, 15000, 15000 for these datasets respectively. In the classification stage, We set batch size of 64, 64, 128, 128 for four datasets respectively. The number of epoch is set to 300 for all datasets. We calculate the results by averaging the results of ten repeated experiments with different training sample selected randomly.

\subsubsection{\textbf{Comparing With Other Methods}} In order to compare the effectiveness of the proposed DiffCRN, several representative algorithms were selected for the control experiments, including three conventional classifiers (i.e., SVM, KNN, RF), CNNs-based networks (i.e., 1D-CNN \cite{1DCNN1}, 2D-CNN \cite{2DCNN}, 3D-CNN \cite{Hamida3dCNN}, SSRN \cite{SSRN}, HybridSN \cite{HybridSN}, SS-ConvNeXt \cite{SS-ConvNeXt}), GAN-based network (i.e., MTGAN \cite{Hangrl2020_TGRS}, ADGAN \cite{WangTGRS_2021}), transformer-based networks (i.e., SSFTT \cite{SSFTT}, SSTN \cite{SSTN}), unsupervised method (i.e., SC-SS-MTr \cite{Huang2023_TGRS}):
\begin{enumerate}[i)]

\item The SSRN introduces the ResNet structure into the designed spatial 3-D CNN module and spectral 3-D CNN module and extracted rich spatial and spectral features. 

\item The HybridSN consists of 3-D CNN and 2-D CNN layers. An image patch sample passes through three 3-D CNN layers and one 2-D CNN layer successively to obtain a spectral–spatial joint feature.

\item The SS-ConvNeXt is our previous work published in 2023. A spectral-ConvNext and a Spatial-ConvNeXt block are designed to capture spectral and spatial information, respectively.

\item The MTGAN is a multitask generative adversarial network using encode-decoder to generate fake image, and using a discriminator to discriminate the input sample coming from the real distribution or the reconstructed one. The extracted features in encoder-decoder are used to make classification.

\item The ADGAN consists of a generator and a discriminator. The generator consists of four 2D transposed convolution blocks with Tanh layer, each 2D transposed convolution blocks consists of ConvTransposed2d layer, BN layer, and a ReLU activation function. The discriminator consists of five 2D CNN blocks and a softmax layer.

\item The SSFTT is a spatial-spectral Transformer that utilized the advantages of hybrid CNN and Transformer with a gaussian weighted feature tokenizer module.

\item The SSTN is a spectral–spatial transformer network, which consists of spatial attention and spectral association modules, to overcome the constraints of convolution kernels.

\item The SC-SS-MTr is a spectral-spatial masked transformer with supervised and contrastive learning, which pretrains a vanilla Transformer via reconstruction from masked inputs then fine-tune for HSIC. 

\end{enumerate}

\subsubsection{\textbf{Evaluation Metrics}}
To quantitatively evaluate the proposed method and other compared methods, we choose following commonly used metrics, i.e., Overall Classification Accuracy (\textbf{\textit{OA}} $\uparrow$), Average Classification Accuracy (\textbf{\textit{AA}} $\uparrow$), Category Accuracy (\textbf{\textit{CA}} $\uparrow$), and Kappa Coefficient (\textbf{$k$} $\uparrow$), Mean Intersection over Union (\textbf{\textit{MIoU}} $\uparrow$), Frequency Weighted Intersection over Union (\textbf{\textit{FWIoU}} $\uparrow$) and standard deviation ($\sigma$) of classwise accuracy. Among them, the MIoU, and FWIoU can be computed by:
\begin{equation}
\left\{
\begin{aligned}
&MIoU = \frac{1}{K+1} \underset{i=0}{\overset{K}{\varSigma}} \frac{N_{ii}}{\underset{j=0}{\overset{K}{\varSigma}}N_{ij} + \underset{j=0}{\overset{K}{\varSigma}}N_{ji} - N_{ii}}\\
&FWIoU = \frac{1}{\underset{i=0}{\overset{K}{\varSigma}} \underset{j=0}{\overset{K}{\varSigma}} N_{ij}} \underset{i=0}{\overset{K}{\varSigma}} \frac{N_{ii}}{\underset{j=0}{\overset{K}{\varSigma}}N_{ij} + \underset{j=0}{\overset{K}{\varSigma}}N_{ji} - N_{ii}}\\
\end{aligned}
\right.
\end{equation}
where $N \in \mathbb{R}^{K \times K}$ is the confusion matrix. \textbf{\textit{MIoU}} is used to compare the similarity between the model prediction results and the true labels. \textbf{\textit{FWIoU}} is the weighted IoU. Each experiment was repeated ten times and the results were averaged.

\subsection{Performance Analysis}

\subsubsection{\textbf{Numerical Evaluation}}
A quantitative assessment of classification performance is presented in \autoref{table_IN_result}. The best classification values are colored shadow and optimal standard deviation displayed in bold. Our proposed DiffCRN introduces the diffusion model to HSI classification, considering extracted features from various diffusion time step $t$ based on SAM, resulting in powerful ability to model complex relations. The results show that the proposed DiffCRN is superior to all other techniques in terms of \textit{OA}, \textit{AA}, \textit{$k$}, \textit{FWIoU}, \textit{MIoU}, and represent better performance on classwise accuracy simultaneously. 

\subsubsubsection{\textbf{Performance on Indian Pines}}\autoref{table_IN_result} shows the quantitative comparison results of the Indian Pines dataset. Benefiting from the pretrained DiffCRN, the performance of DiffCRN is higher than other conventional classifiers, classic backbone, GAN based networks and transformered based methods as well as unsupervised network with pretraining. For instance, DiffCRN is 6.7\%,  11.8\%, 3.9\%, 10.8\%, 6.1\% and 14.1\% better than SSRN, MTGAN, ADGAN, SSFTT, SSTN and SC-SS-MTr in terms of MIoU. The results show that DiffCRN can obtain label domain classification results with higher precision masks that have better raster reversibility. In terms of AA, HybridSN and SSRN provide better performance than conventional classifier, e.g., RF. Transformer based network SSFTT can not provide better performance than unsupervised network with pretraining method SC-SS-MTr. The proposed DiffCRN exhibits a performance that is superior to SC-SS-MTr, with an increase of approximately 1.7\% in AA. Further more, under the circumstance of samll categories, e.g., class 7 (Grass-pasture-mowed) and class 9 (Oats), the SC-SS-MTr achieves the poor performance, only 8.89\% accuracy on class 9, although it is better in OA. On the contrary, the proposed DiffCRN performs better on challenging classes (small-sample classes).
    
\subsubsubsection{\textbf{Performance on Pavia University}} \autoref{table_PU_result} compares the experimental results to various classes on the Pavia University dataset. From \autoref{table_PU_result}, DiffCRN achieves the best accuracy (OA of 99.33\%, AA of 99.26\%, and  $k$ of 99.31\%, FWIoU of 98.81\% and MIoU of 98.60\%) among the methods. The classification performance of SSRN, SSTN and SC-SS-MTr are close on AA. Experimental results show that using pretrained diffusion model to conduct classification task successfully achieving high accuracy. The OA obtained by DiffCRN on the Pavia University dataset outperforms the ones of SC-SS-MTr, SSFTT, MTGAN, SSRN by 0.58 percentage points, 10.75 percentage points, 0.67 percentage points and 1.12 percentage points, respectively.

\subsubsubsection{\textbf{Performance on WHU-Hi-HongHu}} \autoref{table_HongHu} summarizes the comparative results on the WHU-Hi-HongHu dataset. We can see from \autoref{dataset} that the target distribution is more complex than Indian Pines and Pavia University dataset. Notably, the region is planted with different cultivars of the same crop type; for example, Chinese cabbage/cabbage (class 7/9) and and brassica chinensis/small brassica chinensis (class 12/13). Not surprisingly, conventional classifier fails to complete this complex agricultural scene classification task with many classes of crops. We also see that obvious noticeable rise with the direction of classic backbone (e.g., HybridSN, SSRN), GAN based methods (e.g., MTGAN), transformer based methods (e.g. SSTN) and SC-SSS-MTr in terms of OA. Just as we talked about the cultivars of the same crop type, the accuracy obtained by DiffCRN on Chinese cabbage/cabbage with outperforms the SSRN, MTGAN, SC-SS-MTr by 8.25\%/5.29\%, 0.99\%/0.67\%, 2.42\%/2.17\%, respectively.

\subsubsubsection{\textbf{Performance on MUUFL}} The quantitative results are listed in\autoref{table_MUUFL_result}. Both KNN, RF and SVM outperform the GAN based network ADGAN. The DiffCRN shows better accuracy than that of all other techniques, including GAN based, transformer based and pre-training based approaches, with OA, AA, $k$, FWIoU and MIoU of 93.44 ± 0.28\%, 82.77 ± 1.08\%, and 91.56 ± 0.36\%, 88.44±0.44\% and 76.26±1.06\%, respectively.  

\begin{figure}[htbp]
\centering
\subfigure[Indian Pines]{
\includegraphics[width=0.22\textwidth]{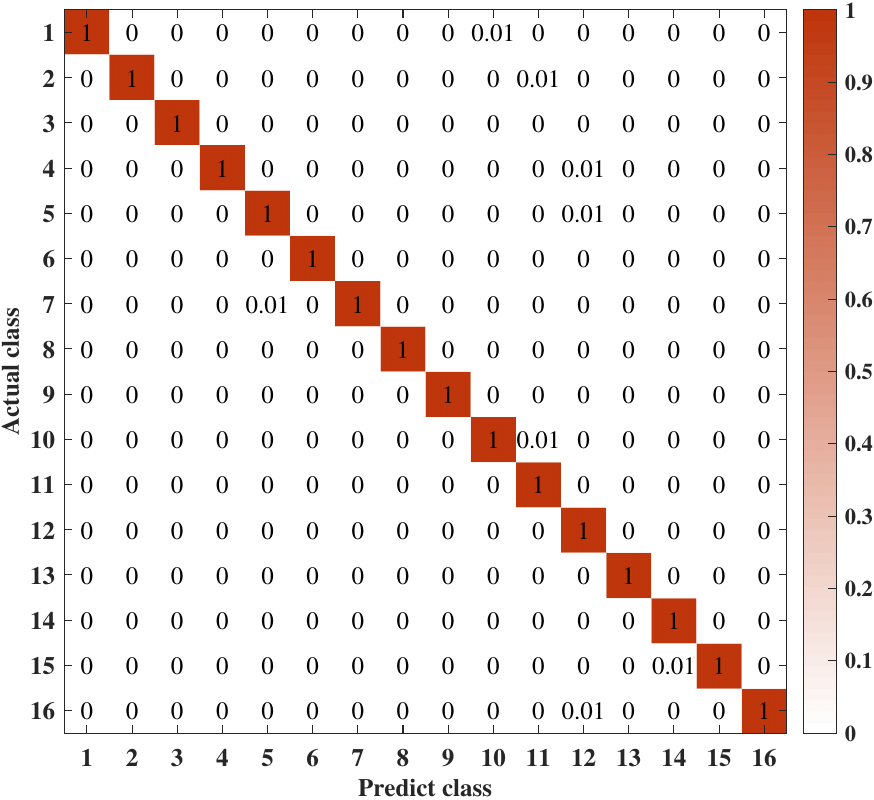}
}
\subfigure[Pavia University]{
\includegraphics[width=0.22\textwidth]{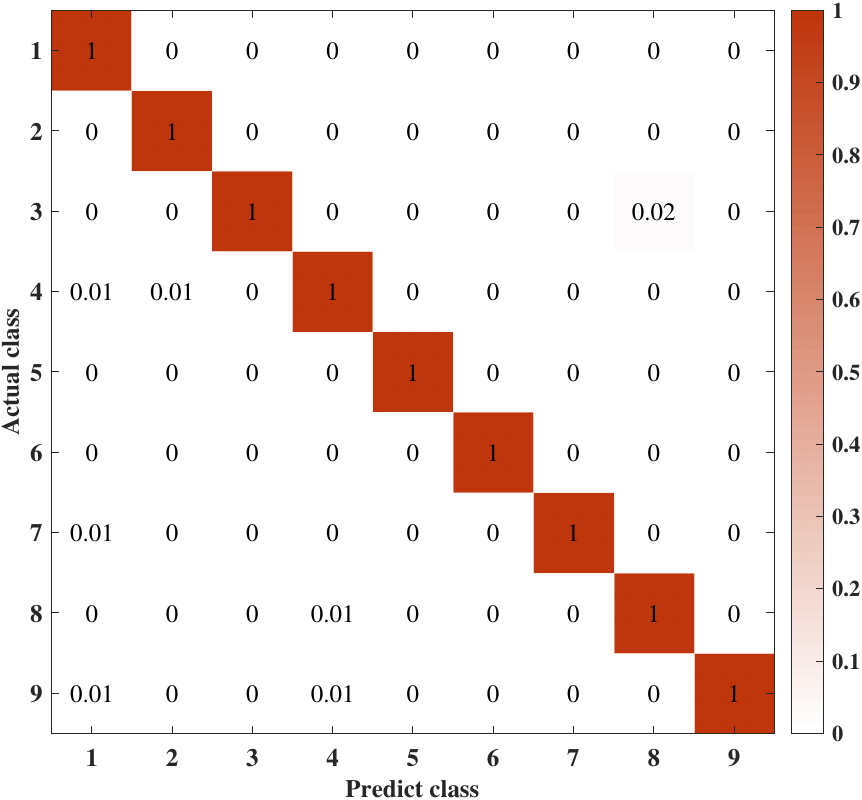}
}
\subfigure[WHU-Hi-HongHu]{
\includegraphics[width=0.22\textwidth]{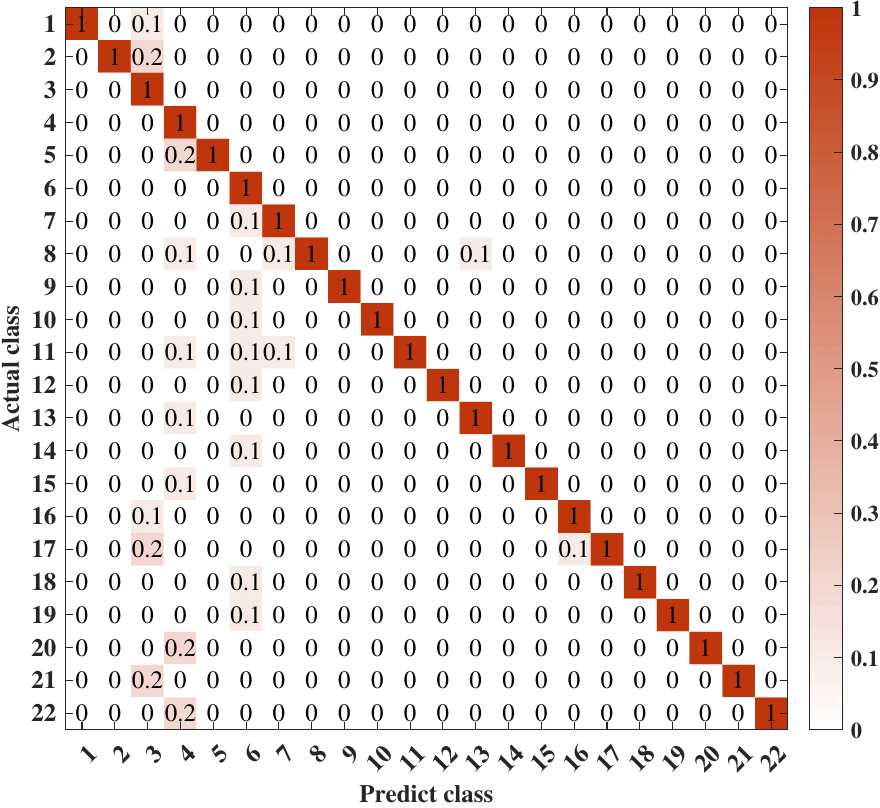}
}
\subfigure[MUUFL]{
\includegraphics[width=0.22\textwidth]{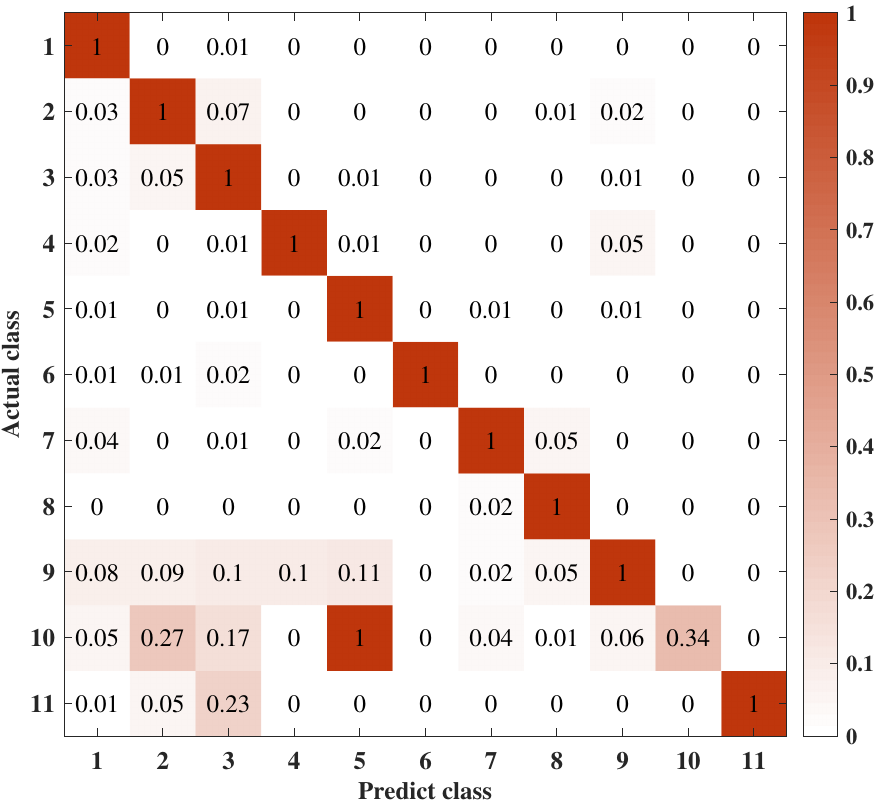}
}
\caption{Confusion matrix of Proposed DiffCRN on four dataset. $N_{ij}$ of confusion matrix means the $i$-th class is classified to the $j$-th class. Each row represents the class predicted, each col represents the true class. Results on Exp1.}
\label{confusionmatrix}
\end{figure}

Furthermore, the classification confusion matrix of DiffCRN is shown in \autoref{confusionmatrix}. From the diagonal elements, we can see that the number of each class varies slightly, except class 10 (raod) in MUUFL dataset.

\begin{table*}[htbp]
\Huge
\centering
\caption{Quantitative performance of different classification methods in terms of OA, AA, $k$, FWIoU, MIoU, as well as the accuracies for each class on the Indian Pines dataset with 10 \% training sample. The best results are colored shadow and optimal standard deviation in bold. Results on Exp1.}
\resizebox{\textwidth}{!}{
\begin{tabular}{c|c|ccc|cccccc|cc|cc|c|c}
\hline
\multirow{2}{*}{\makecell[c]{Class \\ No.}} &
  \multirow{2}{*}{Color} &
  \multicolumn{3}{c|}{Conventional Classifier} &
  \multicolumn{6}{c|}{CNNs Based networks} &
  \multicolumn{2}{c|}{GAN Based Networks} &
  \multicolumn{2}{c|}{Transformer Based Networks} &
  \makecell[c]{Unsupervised \\ Network} &
  \multirow{2}{*}{\textbf{DiffCRN}} \\ \cline{3-16}
 &
   &
  KNN &
  RF &
  SVM &
  1D-CNN \cite{1DCNN1} &
  2D-CNN \cite{2DCNN} &
  3D-CNN \cite{Hamida3dCNN} &
  HybridSN \cite{HybridSN}&
  SSRN \cite{SSRN}&
  SS-ConvNeXt \cite{SS-ConvNeXt}&
  MTGAN \cite{Hangrl2020_TGRS}&
  ADGAN \cite{WangTGRS_2021}&
  SSFTT \cite{SSFTT}&
  SSTN \cite{SSTN}&
  SC-SS-MTr \cite{Huang2023_TGRS}&
   \\ \hline \specialrule{0em}{3pt}{3pt} \hline
\multicolumn{1}{c|}{1} &
  \cellcolor[RGB]{255, 212, 0} &
  11.43±9.38 &
  7.86±9.12 &
  15.24±6.85 &
  58.78±29.66 &
  76.59±10.36 &
  81.71±18.27 &
  88.78±5.66 &
  90.98±9.13 &
  86.05±12.16 &
  84.63±9.89 &
  97.87±2.75 &
  94.18±7.55 &
  94.39±5.02 &
  91.95±6.82 &
  \cellcolor[RGB]{251, 228, 213}98.65±\textbf{1.30} \\
\multicolumn{1}{c|}{2} &
  \cellcolor[RGB]{255, 158, 9} &
  57.17±4.36 &
  64.70±1.92 &
  67.15±2.99 &
  64.79±6.14 &
  86.05±5.22 &
  86.89±2.94 &
  95.60±1.46 &
  97.54±1.02 &
  97.88±1.24 &
  97.20±1.40 &
  96.33±1.49 &
  94.90±2.99 &
  98.01±1.27 &
  97.28±1.35 &
  \cellcolor[RGB]{251, 228, 213}99.36±\textbf{0.32} \\
\multicolumn{1}{c|}{3} &
  \cellcolor[RGB]{38, 112, 0} &
  45.18±2.13 &
  54.61±2.20 &
  55.07±2.69 &
  57.40±8.06 &
  84.99±3.99 &
  81.97±4.44 &
  94.04±2.32 &
  96.88±2.48 &
  98.54±2.06 &
  95.90±3.13 &
  97.71±2.95 &
  92.16±4.24 &
  92.05±5.24 &
  97.54±1.35 &
  \cellcolor[RGB]{251, 228, 213}99.01±\textbf{0.90} \\
\multicolumn{1}{c|}{4} &
  \cellcolor[RGB]{166, 112, 0} &
  19.02±4.49 &
  27.99±7.99 &
  34.72±5.36 &
  58.08±8.49 &
  85.31±7.80 &
  84.90±5.00 &
  90.52±6.67 &
  97.70±2.46 &
  \cellcolor[RGB]{251, 228, 213}99.38±\textbf{0.75} &
  96.85±2.77 &
  98.36±1.76 &
  94.80±3.00 &
  96.24±2.30 &
  94.98±3.60 &
  99.13±1.19 \\
\multicolumn{1}{c|}{5} &
  \cellcolor[RGB]{255, 166, 227} &
  81.45±2.39 &
  83.06±4.11 &
  78.85±4.07 &
  72.87±7.95 &
  82.99±4.71 &
  90.59±3.86 &
  95.01±2.32 &
  95.13±2.78 &
  94.53±2.65 &
  95.61±3.42 &
  95.11±4.72 &
  95.33±2.92 &
  94.34±4.11 &
  94.29±2.32 &
  \cellcolor[RGB]{251, 228, 213}98.30±\textbf{1.38} \\
\multicolumn{1}{c|}{6} &
  \cellcolor[RGB]{123, 104, 238} &
  96.01±1.52 &
  95.65±1.69 &
  94.67±1.99 &
  90.20±5.28 &
  95.28±2.51 &
  95.35±4.16 &
  99.60±0.39 &
  99.25±0.74 &
  98.82±0.67 &
  98.48±1.12 &
  97.34±2.59 &
  96.66±3.27 &
  98.66±0.66 &
  99.56±0.53 &
  \cellcolor[RGB]{251, 228, 213}99.87±\textbf{0.14} \\
\multicolumn{1}{c|}{7} &
  \cellcolor[RGB]{255, 127, 80} &
  36.15±24.84 &
  2.31±\textbf{3.72} &
  9.62±5.51 &
  64.00±23.40 &
  67.20±17.67 &
  82.67±14.10 &
  95.20±4.92 &
  76.40±32.25 &
  95.45±5.67 &
  12.00±29.45 &
  85.50±12.46 &
  84.92±22.10 &
  92.80±11.12 &
  36.00±32.10 &
  \cellcolor[RGB]{251, 228, 213}98.29±4.54 \\
\multicolumn{1}{c|}{8} &
  \cellcolor[RGB]{95, 158, 160} &
  95.54±1.23 &
  98.10±0.97 &
  98.35±1.05 &
  94.28±5.69 &
  98.35±0.96 &
  99.46±1.12 &
  99.70±0.69 &
  99.53±1.47 &
  99.98±0.07 &
  99.95±0.10 &
  96.62±0.53 &
  99.62±0.56 &
  99.95±0.15 &
  99.95±0.09 &
  \cellcolor[RGB]{251, 228, 213}100.00±\textbf{0.00} \\
\multicolumn{1}{c|}{9} &
  \cellcolor[RGB]{60, 179, 113} &
  6.11±11.84 &
  6.11±8.05 &
  11.11±9.80 &
  63.53±20.15 &
  74.71±20.94 &
  74.02±18.17 &
  91.18±5.72 &
  55.29±39.24 &
  90.71±11.19 &
  61.76±43.87 &
  83.82±17.72 &
  77.83±16.66 &
  93.53±7.04 &
  8.89±7.11 &
  \cellcolor[RGB]{251, 228, 213}99.16±\textbf{2.22} \\
\multicolumn{1}{c|}{10} &
  \cellcolor[RGB]{218, 112, 214} &
  69.39±3.39 &
  65.82±5.49 &
  62.69±2.67 &
  67.79±10.40 &
  84.41±3.79 &
  79.03±4.41 &
  94.31±2.23 &
  96.19±1.52 &
  97.12±1.60 &
  95.43±2.60 &
  95.21±2.63 &
  92.21±4.95 &
  96.02±1.86 &
  95.71±1.82 &
  \cellcolor[RGB]{251, 228, 213}98.01±\textbf{0.69} \\
\multicolumn{1}{c|}{11} &
  \cellcolor[RGB]{144, 238, 144} &
  70.10±1.99 &
  86.57±1.13 &
  74.03±1.69 &
  79.17±8.07 &
  90.27±1.95 &
  88.15±3.56 &
  95.92±2.43 &
  98.29±1.14 &
  98.95±0.69 &
  98.44±1.10 &
  98.76±1.07 &
  97.52±1.25 &
  98.11±1.11 &
  \cellcolor[RGB]{251, 228, 213}99.48±\textbf{0.19} &
  98.43±0.47 \\
\multicolumn{1}{c|}{12} &
  \cellcolor[RGB]{70, 130, 180} &
  28.03±5.95 &
  45.58±4.98 &
  49.89±3.09 &
  42.34±10.33 &
  76.07±4.86 &
  79.59±4.99 &
  93.63±1.98 &
  97.97±1.70 &
  97.94±1.87 &
  95.10±3.07 &
  98.46±2.49 &
  90.48±9.01 &
  97.70±1.56 &
  96.53±1.37 &
  \cellcolor[RGB]{251, 228, 213}99.30±\textbf{0.66} \\
\multicolumn{1}{c|}{13} &
  \cellcolor[RGB]{250, 164, 96} &
  81.41±2.53 &
  92.81±1.82 &
  92.16±3.41 &
  95.41±2.89 &
  94.86±6.67 &
  93.33±5.68 &
  99.73±0.58 &
  99.68±0.58 &
  99.62±0.64 &
  98.76±2.18 &
  98.24±1.65 &
  96.59±6.15 &
  99.84±0.51 &
  97.83±3.77 &
  \cellcolor[RGB]{251, 228, 213}99.85±\textbf{0.41} \\
\multicolumn{1}{c|}{14} &
  \cellcolor[RGB]{154, 205, 50} &
  92.92±1.80 &
  95.21±2.22 &
  89.51±3.15 &
  94.47±3.10 &
  96.77±1.72 &
  96.11±1.82 &
  99.32±0.53 &
  99.57±0.43 &
  98.43±3.24 &
  99.12±1.62 &
  \cellcolor[RGB]{251, 228, 213}99.89±\textbf{0.10} &
  98.67±1.47 &
  99.37±0.59 &
  99.68±0.36 &
  99.58±0.37 \\
\multicolumn{1}{c|}{15} &
  \cellcolor[RGB]{107, 142, 35} &
  15.14±3.11 &
  39.28±5.99 &
  52.24±6.96 &
  44.24±10.57 &
  86.97±8.85 &
  81.53±9.74 &
  95.73±2.66 &
  98.01±2.61 &
  97.99±2.39 &
  98.10±1.56 &
  97.19±3.82 &
  83.48±12.18 &
  90.61±9.10 &
  98.47±1.26 &
  \cellcolor[RGB]{251, 228, 213}99.79±\textbf{0.43} \\
\multicolumn{1}{c|}{16} &
  \cellcolor[RGB]{110, 121, 85} &
  80.00±2.62 &
  79.05±5.92 &
  69.76±8.77 &
  91.67±3.55 &
  97.02±3.38 &
  81.45±7.79 &
  98.45±2.69 &
  97.38±2.68 &
  97.65±3.70 &
  92.50±5.24 &
  88.39±11.57 &
  89.93±15.59 &
  97.26±1.78 &
  70.60±13.97 &
  \cellcolor[RGB]{251, 228, 213}99.66±\textbf{0.58} \\ \hline \specialrule{0em}{3pt}{3pt} \hline
\multicolumn{2}{c|}{AA (\%) $\uparrow$} &
  56.19±1.88 &
  59.04±1.38 &
  59.69±1.27 &
  71.19±2.76 &
  86.12±6.59 &
  86.05±2.24 &
  95.42±0.92 &
  93.49±4.01 &
  96.82±1.30 &
  88.80±3.60 &
  95.48±1.55 &
  92.46±3.65 &
  96.18±1.43 &
  96.13±2.31 &
  \cellcolor[RGB]{251, 228, 213}99.26±\textbf{0.37} \\
\multicolumn{2}{c|}{OA (\%) $\uparrow$} &
  68.20±0.76 &
  74.85±1.00 &
  74.63±0.48 &
  73.75±0.80 &
  88.83±1.57 &
  87.90±1.11 &
  96.18±0.76 &
  97.81±0.66 &
  98.20±0.52 &
  97.20±0.50 &
  97.63±0.90 &
  95.12±1.76 &
  97.15±0.88 &
  97.33±0.40 &
  \cellcolor[RGB]{251, 228, 213}99.33±\textbf{0.16} \\
\multicolumn{2}{c|}{$k$*100 $\uparrow$} &
  63.52±0.88 &
  73.85±1.07 &
  70.93±0.55 &
  70.52±0.81 &
  88.40±1.63 &
  87.34±1.21 &
  96.07±0.81 &
  97.74±0.69 &
  98.11±0.58 &
  97.08±0.52 &
  97.94±0.85 &
  94.79±1.95 &
  96.94±0.96 &
  96.95±0.46 &
  \cellcolor[RGB]{251, 228, 213}99.31±\textbf{0.16} \\
\multicolumn{2}{c|}{FWIoU (\%) $\uparrow$} &
  30.92±2.33 &
  41.80±2.13 &
  41.82±2.06 &
  60.07±0.85 &
  81.91±2.27 &
  80.52±1.64 &
  93.44±1.28 &
  96.15±1.14 &
  96.81±0.93 &
  95.05±0.85 &
  96.48±1.39 &
  91.57±3.02 &
  95.06±1.37 &
  95.30±0.75 &
  \cellcolor[RGB]{251, 228, 213}98.81±\textbf{0.28} \\
\multicolumn{2}{c|}{MIoU (\%) $\uparrow$} &
  34.48±2.22 &
  43.11±1.71 &
  43.65±1.65 &
  59.35±2.09 &
  79.46±2.48 &
  77.69±2.12 &
  91.58±1.52 &
  91.88±4.45 &
  94.88±1.89 &
  86.72±3.74 &
  94.72±1.76 &
  87.76±4.63 &
  92.54±2.52 &
  84.55±2.83 &
  \cellcolor[RGB]{251, 228, 213}98.60±\textbf{0.55} \\ \hline
\end{tabular}
}
\label{table_IN_result}
\end{table*}

\begin{table*}[htbp]
\Huge
\centering
\caption{Quantitative performance of different classification methods in terms of OA, AA, $k$, FWIoU, MIoU, as well as the accuracies for each class on the Pavia University dataset with 3 \% training sample. The best results are colored shadow and optimal standard deviation in bold. Results on Exp1.}
\resizebox{\textwidth}{!}{
\begin{tabular}{c|c|ccc|cccccc|cc|cc|c|c}
\hline
\multirow{2}{*}{\makecell[c]{Class \\ No.}} &
  \multirow{2}{*}{Color} &
  \multicolumn{3}{c|}{Conventional Classifier} &
  \multicolumn{6}{c|}{CNNs Based networks} &
  \multicolumn{2}{c|}{GAN Based Networks} &
  \multicolumn{2}{c|}{Transformer Based Networks} &
  \makecell[c]{Unsupervised \\ Network} &
  \multirow{2}{*}{\textbf{DiffCRN}} \\ \cline{3-16}
\multicolumn{1}{c|}{} &
   &
  KNN &
  RF &
  SVM &
  1D-CNN \cite{1DCNN1} &
  2D-CNN \cite{2DCNN} &
  3D-CNN \cite{Hamida3dCNN} &
  HybridSN \cite{HybridSN}&
  SSRN \cite{SSRN}&
  SS-ConvNeXt \cite{SS-ConvNeXt}&
  MTGAN \cite{Hangrl2020_TGRS}&
  ADGAN \cite{WangTGRS_2021}&
  SSFTT \cite{SSFTT}&
  SSTN \cite{SSTN}&
  SC-SS-MTr \cite{Huang2023_TGRS}&
   \\ \hline \specialrule{0em}{3pt}{3pt} \hline
\multicolumn{1}{c|}{1} &
  \cellcolor[RGB]{255, 173, 189} &
  84.90±1.81 &
  89.04±1.40 &
  88.79±1.77 &
  84.17±3.93 &
  95.30±1.26 &
  95.08±0.96 &
  96.22±3.78 &
  98.74±0.63 &
  98.79±1.60&
  98.68±0.75 &
  71.35±18.54 &
  86.53±29.32 &
  98.20±0.96 &
  99.35±0.44 &
  \cellcolor[RGB]{251, 228, 213}99.83±\textbf{0.20} \\
\multicolumn{1}{c|}{2} &
  \cellcolor[RGB]{73, 182, 120} &
  96.35±1.52 &
  96.93±0.60 &
  95.65±0.61 &
  95.48±1.83 &
  98.09±0.70 &
  98.63±0.49 &
  99.22±0.68 &
  99.74±0.15 &
  99.73±0.43&
  99.79±0.12 &
  94.08±5.23 &
  94.84±13.20 &
  96.58±1.52 &
  99.92±0.07 &
  \cellcolor[RGB]{251, 228, 213}99.98±\textbf{0.02} \\
\multicolumn{1}{c|}{3} &
  \cellcolor[RGB]{255, 153, 56} &
  57.30±2.62 &
  53.19±4.16 &
  68.54±3.25 &
  59.47±11.73 &
  77.83±4.32 &
  86.72±5.28 &
  84.33±6.05 &
  90.69±5.71 &
  94.66±2.77&
  97.16±2.89 &
  65.31±50.62 &
  87.57±28.90 &
  97.23±1.61 &
  96.76±1.43 &
  \cellcolor[RGB]{251, 228, 213}98.64±\textbf{1.06} \\
\multicolumn{1}{c|}{4} &
  \cellcolor[RGB]{49, 107, 185} &
  75.50±3.96 &
  84.09±2.94 &
  85.13±3.30 &
  85.90±4.41 &
  97.97±0.58 &
  95.44±1.42 &
  98.07±1.07 &
  97.30±0.82 &
  96.21±0.94&
  97.44±\textbf{0.23} &
  58.87±22.54 &
  84.86±30.53 &
  96.09±0.68 &
  97.40±1.29 &
  \cellcolor[RGB]{251, 228, 213}99.17±0.38 \\
\multicolumn{1}{c|}{5} &
  \cellcolor[RGB]{255, 54, 31} &
  99.03±0.32 &
  97.62±0.44 &
  91.38±2.76 &
  99.10±0.83 &
  99.93±0.14 &
  99.36±1.54 &
  \cellcolor[RGB]{251, 228, 213}100.00±\textbf{0.00} &
  99.89±0.15 &
  99.86±0.13&
  99.59±0.30 &
  92.82±11.32 &
  99.39±1.22 &
  99.92±0.12 &
  99.75±0.38 &
  \cellcolor[RGB]{251, 228, 213}100.00±\textbf{0.00} \\
\multicolumn{1}{c|}{6} &
  \cellcolor[RGB]{91, 66, 162} &
  45.92±4.74 &
  51.65±1.99 &
  82.67±2.47 &
  50.62±7.93 &
  91.17±1.76 &
  89.67±3.75 &
  95.58±1.95 &
  99.50±0.49 &
  99.70±0.41&
  99.73±0.21 &
  84.95±14.71 &
  79.68±41.77 &
  99.51±1.08 &
  99.84±0.23 &
  \cellcolor[RGB]{251, 228, 213}99.96±\textbf{0.08} \\
\multicolumn{1}{c|}{7} &
  \cellcolor[RGB]{130, 67, 36} &
  74.50±4.01 &
  69.89±6.10 &
  79.01±3.65 &
  74.03±13.41 &
  81.49±3.49 &
  85.16±5.83 &
  97.87±1.52 &
  96.11±3.74 &
  98.61±1.05&
  95.97±2.50 &
  56.61±36.41 &
  88.78±31.35 &
  98.53±1.66 &
  97.84±1.37 &
  \cellcolor[RGB]{251, 228, 213}99.85±\textbf{0.22} \\
\multicolumn{1}{c|}{8} &
  \cellcolor[RGB]{180, 190, 190} &
  79.29±4.09 &
  85.12±2.90 &
  79.24±2.85 &
  81.91±5.93 &
  92.02±1.74 &
  88.77±3.88 &
  92.87±3.46 &
  97.02±2.35 &
  97.41±1.05&
  98.83±0.79 &
  82.70±24.22 &
  77.01±40.15 &
  97.75±0.81 &
  \cellcolor[RGB]{251, 228, 213}99.05±\textbf{0.20} &
  98.92±0.72 \\
\multicolumn{1}{c|}{9} &
  \cellcolor[RGB]{102, 211, 216} &
  99.72±0.10 &
  99.39±0.44 &
  75.95±8.76 &
  \cellcolor[RGB]{251, 228, 213}99.74±\textbf{0.12} &
  98.84±1.36 &
  91.99±2.73 &
  98.27±0.94 &
  98.97±1.25 &
  97.23±0.78 &
  96.61±1.68 &
  43.09±26.39 &
  87.61±20.54 &
  97.66±1.57 &
  89.81±5.30 &
  98.31±0.73 \\ \hline \specialrule{0em}{3pt}{3pt} \hline 
\multicolumn{2}{c|}{AA (\%) $\uparrow$} &
  79.17±0.55 &
  80.77±1.00 &
  82.93±1.30 &
  81.16±5.57 &
  92.52±1.70 &
  92.31±0.37 &
  95.83±2.16 &
  97.55±1.70 &
  98.02±0.47 &
  98.20±1.05 &
  72.20±16.97 &
  87.36±26.33 &
  97.94±1.11 &
  97.75±0.73 &
  \cellcolor[RGB]{251, 228, 213}99.41±\textbf{0.21} \\
\multicolumn{2}{c|}{OA (\%) $\uparrow$} &
  83.41±0.41 &
  85.53±0.37 &
  88.80±0.57 &
  84.37±0.71 &
  94.88±0.33 &
  94.82±0.84 &
  96.93±0.73 &
  98.58±0.33 &
  98.79±0.37 &
  \multicolumn{1}{c}{99.03±0.21} &
  \multicolumn{1}{c|}{71.35±18.54} &
  88.96±19.20 &
  97.46±0.80 &
  99.12±0.50 &
  \cellcolor[RGB]{251, 228, 213}99.70±\textbf{0.10} \\
\multicolumn{2}{c|}{$k$*100 (\%) $\uparrow$} &
  77.39±0.59 &
  80.96±0.50 &
  85.11±0.77 &
  78.97±1.05 &
  93.40±0.41 &
  93.31±1.09 &
  96.03±0.96 &
  98.16±0.43 &
  98.43±0.50 &
  \multicolumn{1}{c}{98.76±0.27} &
  \multicolumn{1}{c|}{94.08±5.23} &
  85.77±24.64 &
  96.70±1.04 &
  98.86±0.27 &
  \cellcolor[RGB]{251, 228, 213}99.61±\textbf{0.13} \\
\multicolumn{2}{c|}{FWIoU (\%) $\uparrow$} &
  63.42±0.75 &
  65.62±1.49 &
  71.58±2.21 &
  73.70±1.10 &
  90.80±.55 &
  90.63±1.41 &
  94.38±1.24 &
  94.33±.62 &
  97.70±0.71 &
  \multicolumn{1}{c}{98.16±0.39} &
  \multicolumn{1}{c|}{65.31±50.62} &
  85.33±23.28 &
  95.73±1.18 &
  98.31±0.40 &
  \cellcolor[RGB]{251, 228, 213}99.42±\textbf{0.19} \\
\multicolumn{2}{c|}{MIoU (\%) $\uparrow$} &
  66.00±0.61 &
  67.62±1.49 &
  72.04±2.11 &
  71.68±1.15 &
  87.75±0.90 &
  87.91±2.17 &
  92.65±1.44 &
  95.96±1.01 &
  96.63±1.01 &
  \multicolumn{1}{c}{96.65±0.79} &
  \multicolumn{1}{c|}{58.87±22.54} &
  82.34±26.08 &
  92.38±1.57 &
  96.29±1.19 &
  \cellcolor[RGB]{251, 228, 213}99.00±\textbf{0.33} \\ \hline
\end{tabular}
}
\label{table_PU_result}
\end{table*}


\begin{table*}[htbp]
\Huge
\centering
\caption{Quantitative performance of different classification methods in terms of OA, AA, $k$, FWIoU, MIoU, as well as the accuracies for each class on the WHU-Hi-HongHu dataset with 0.5 \% training sample. The best results are colored shadow and optimal standard deviation in bold. Results on Exp1.}
\resizebox{\textwidth}{!}{
\begin{tabular}{cc|ccc|cccccc|cc|cc|c|c}
\hline
\multicolumn{1}{c|}{\multirow{2}{*}{Class No.}} &
  \multirow{2}{*}{Color} &
  \multicolumn{3}{c|}{Conventional Classifier} &
  \multicolumn{6}{c|}{CNNs Based networks} &
  \multicolumn{2}{c|}{GAN Based Networks} &
  \multicolumn{2}{c|}{Transformer Based Networks} &
  \makecell[c]{Unsupervised \\ Network} &
  \multirow{2}{*}{\textbf{DiffCRN}} \\ \cline{3-16}
\multicolumn{1}{c|}{} &
   &
  KNN &
  RF &
  SVM &
  1D-CNN \cite{1DCNN1} &
  2D-CNN \cite{2DCNN} &
  3D-CNN \cite{Hamida3dCNN} &
  HybridSN \cite{HybridSN}&
  SSRN \cite{SSRN}&
  SS-ConvNeXt \cite{SS-ConvNeXt}&
  MTGAN \cite{Hangrl2020_TGRS}&
  ADGAN \cite{WangTGRS_2021}&
  SSFTT \cite{SSFTT}&
  SSTN \cite{SSTN}&
  SC-SS-MTr \cite{Huang2023_TGRS}&
   \\ \hline \specialrule{0em}{3pt}{3pt} \hline
\multicolumn{1}{c|}{1} &
  \cellcolor[RGB]{144, 73, 65} &
  84.16±1.72 &
  82.14±2.45 &
  82.34±1.80 &
  89.19±2.68 &
  96.18±1.43 &
  95.30±1.01 &
  94.73±2.82 &
  95.49±1.68 &
  \cellcolor[RGB]{251, 228, 213}98.29±\textbf{0.69} &
  97.16±1.12 &
  96.60±2.04 &
  92.08±2.96 &
  95.45±2.26 &
  96.96±1.29 &
  97.40±1.34 \\
\multicolumn{1}{c|}{2} &
  \cellcolor[RGB]{235, 210, 123} &
  67.44±8.56 &
  56.90±6.97 &
  49.84±9.47 &
  70.37±6.65 &
  84.83±5.59 &
  71.20±6.65 &
  65.75±13.28 &
  71.84±9.50 &
  \cellcolor[RGB]{251, 228, 213}90.49±3.94 &
  85.71±7.21 &
  79.73±8.58 &
  72.12±7.23 &
  88.96±6.26 &
  77.92±7.37 &
  88.46±\textbf{2.34} \\
\multicolumn{1}{c|}{3} &
  \cellcolor[RGB]{144, 145, 226} &
  87.24±1.92 &
  89.36±2.05 &
  82.88±1.89 &
  87.79±1.91 &
  91.45±2.07 &
  91.51±2.05 &
  92.24±1.85 &
  91.49±1.21 &
  95.48±2.50 &
  96.68±1.34 &
  98.35±3.20 &
  91.27±2.10 &
  94.02±1.26 &
  94.58±1.53 &
  \cellcolor[RGB]{251, 228, 213}97.04±\textbf{1.11} \\
\multicolumn{1}{c|}{4} &
  \cellcolor[RGB]{173, 201, 68} &
  98.56±0.22 &
  98.21±0.21 &
  97.77±0.27 &
  97.70±0.62 &
  99.32±0.29 &
  99.35±0.27 &
  98.85±0.36 &
  99.22±0.21 &
  99.77±0.18 &
  99.72±0.14 &
  \cellcolor[RGB]{251, 228, 213}99.93±0.53 &
  98.61±0.62 &
  99.17±0.67 &
  99.74±0.07 &
  99.89±\textbf{0.07} \\
\multicolumn{1}{c|}{5} &
  \cellcolor[RGB]{93, 147, 205} &
  14.70±4.36 &
  17.38±3.76 &
  31.63±4.30 &
  45.67±9.92 &
  76.91±6.10 &
  75.94±3.46 &
  77.04±7.92 &
  85.41±3.19 &
  92.04±2.55 &
  92.91±2.38 &
  35.55±8.06 &
  56.82±9.56 &
  86.11±8.46 &
  \cellcolor[RGB]{251, 228, 213}93.63±2.30 &
  93.41±\textbf{1.82} \\
\multicolumn{1}{c|}{6} &
 \cellcolor[RGB]{232, 119, 106}  &
  84.93±1.46 &
  86.24±0.86 &
  85.98±0.83 &
  89.26±2.09 &
  94.75±1.89 &
  93.37±1.56 &
  93.85±1.55 &
  96.67±1.22 &
  99.01±0.49 &
  99.24±0.45 &
  99.08±1.96 &
  94.95±1.66 &
  97.36±3.24 &
  97.86±1.04 &
  \cellcolor[RGB]{251, 228, 213}99.44±\textbf{0.41} \\
\multicolumn{1}{c|}{7} &
  \cellcolor[RGB]{96, 220, 147} &
  63.08±2.42 &
  72.44±1.79 &
  64.94±3.88 &
  71.58±3.47 &
  80.47±3.19 &
  79.17±5.28 &
  79.85±4.80 &
  88.05±4.30 &
  95.99±1.75 &
  95.31±1.41 &
  93.54±3.64 &
  85.67±3.23 &
  93.27±2.69 &
  93.88±1.56 &
  \cellcolor[RGB]{251, 228, 213}96.30±\textbf{1.33} \\
\multicolumn{1}{c|}{8} &
 \cellcolor[RGB]{153, 190, 226}  &
  2.54±\textbf{0.73} &
  1.61±1.15 &
  12.08±3.28 &
  7.06±4.31 &
  18.81±6.39 &
  30.27±7.50 &
  33.39±9.25 &
  28.94±9.75 &
  74.13±6.99 &
  71.57±8.44 &
  60.18±2.88 &
  41.58±9.25 &
  66.67±6.39 &
  58.48±5.59 &
  \cellcolor[RGB]{251, 228, 213}76.55±5.72 \\
\multicolumn{1}{c|}{9} &
 \cellcolor[RGB]{249, 186, 119}  &
  68.04±1.62 &
  86.55±2.23 &
  88.85±2.01 &
  93.24±1.79 &
  95.77±2.28 &
  93.06±2.32 &
  94.61±1.48 &
  93.71±3.60 &
  \cellcolor[RGB]{251, 228, 213}99.55±\textbf{0.42} &
  98.33±0.90 &
  92.75±2.72 &
  94.84±2.07 &
  98.71±0.50 &
  96.83±1.52 &
  99.00±0.75 \\
\multicolumn{1}{c|}{10} &
 \cellcolor[RGB]{180, 128, 206}  &
  16.91±2.36 &
  26.19±3.24 &
  45.23±2.15 &
  49.23±7.06 &
  75.00±4.02 &
  77.11±7.61 &
  68.27±5.53 &
  74.04±6.26 &
  96.40±1.62 &
  95.40±2.17 &
  76.77±6.03 &
  76.62±5.83 &
  90.85±2.74 &
  92.91±\textbf{1.48} &
  \cellcolor[RGB]{251, 228, 213}96.49±1.65 \\
\multicolumn{1}{c|}{11} &
 \cellcolor[RGB]{67, 136, 0}  &
  13.87±1.85 &
  20.97±3.98 &
  37.08±3.08 &
  38.20±5.87 &
  64.78±9.19 &
  64.42±6.39 &
  65.41±9.85 &
  78.02±7.09 &
  93.52±2.55 &
  91.24±2.61 &
  38.04±6.54 &
  71.20±9.20 &
  86.28±4.93 &
  89.07±3.53 &
  \cellcolor[RGB]{251, 228, 213}94.85±\textbf{1.12} \\
\multicolumn{1}{c|}{12} &
 \cellcolor[RGB]{124, 124, 124}  &
  43.11±\textbf{2.53} &
  48.98±2.97 &
  35.68±4.69 &
  51.42±6.36 &
  68.08±5.10 &
  67.69±7.74 &
  52.71±6.16 &
  63.70±7.45 &
  88.83±3.14 &
  85.19±5.73 &
  72.88±8.64 &
  59.49±9.75 &
  81.32±6.01 &
  80.85±4.34 &
  \cellcolor[RGB]{251, 228, 213}90.45±3.72 \\
\multicolumn{1}{c|}{13} &
 \cellcolor[RGB]{136, 128, 117}  &
  53.16±2.81 &
  67.92±2.54 &
  62.05±3.03 &
  63.73±5.66 &
  77.29±2.76 &
  78.13±4.14 &
  72.68±6.10 &
  79.34±3.40 &
  92.59±2.29 &
  93.37±2.02 &
  68.65±4.76 &
  75.00±8.94 &
  90.72±4.22 &
  89.50±\textbf{1.52} &
  \cellcolor[RGB]{251, 228, 213}94.28±1.75 \\
\multicolumn{1}{c|}{14} &
 \cellcolor[RGB]{73, 53, 32}  &
  37.89±5.18 &
  1.74±1.99 &
  48.21±4.24 &
  62.67±4.26 &
  78.55±4.96 &
  75.05±7.46 &
  76.71±8.23 &
  81.87±7.61 &
  93.95±2.07 &
  72.05±40.65 &
  95.80±4.90 &
  83.36±7.00 &
  94.88±2.63 &
  92.76±2.57 &
  \cellcolor[RGB]{251, 228, 213}97.97±\textbf{0.89} \\
\multicolumn{1}{c|}{15} &
  \cellcolor[RGB]{61, 176, 134} &
  5.93±4.13 &
  83.62±4.58 &
  26.67±9.69 &
  24.47±15.43 &
  20.12±11.61 &
  52.82±14.58 &
  70.52±10.43 &
  59.40±17.70 &
  87.51±2.87 &
  30.61±42.16 &
  12.93±16.62 &
  59.31±18.08 &
  87.08±12.81 &
  78.12±11.71 &
  \cellcolor[RGB]{251, 228, 213}97.74±\textbf{1.68} \\
\multicolumn{1}{c|}{16} &
 \cellcolor[RGB]{125, 125, 0}  &
  79.20±6.52 &
  45.71±10.13 &
  65.97±5.18 &
  81.51±6.65 &
  91.46±9.66 &
  88.17±3.60 &
  87.61±4.20 &
  93.07±3.28 &
  97.27±1.45 &
  98.30±1.57 &
  93.55±6.97 &
  89.83±2.47 &
  92.72±6.05 &
  96.71±1.73 &
  \cellcolor[RGB]{251, 228, 213}98.21±\textbf{1.17} \\
\multicolumn{1}{c|}{17} &
 \cellcolor[RGB]{103, 52, 152}  &
  39.01±10.08 &
  6.22±4.11 &
  39.69±7.76 &
  51.21±9.64 &
  72.22±13.89 &
  77.10±9.66 &
  72.38±13.60 &
  80.58±8.05 &
  \cellcolor[RGB]{251, 228, 213}98.17±\textbf{1.67} &
  68.54±39.21 &
  88.046±12.91 &
  84.86±7.93 &
  93.97±7.60 &
  94.58±2.84 &
  91.12±3.39 \\
\multicolumn{1}{c|}{18} &
 \cellcolor[RGB]{43, 21, 8}  &
  6.77±2.75 &
  6.22±4.11 &
  39.15±7.24 &
  28.87±11.01 &
  57.35±12.95 &
  60.13±12.23 &
  69.10±10.90 &
  55.67±18.15 &
  \cellcolor[RGB]{251, 228, 213}93.55±\textbf{2.26} &
  80.81±10.49 &
  76.97±13.49 &
  73.34±14.18 &
  90.18±5.72 &
  88.42±2.34 &
  93.15±2.48 \\
\multicolumn{1}{c|}{19} &
 \cellcolor[RGB]{205, 176, 137}  &
  22.01±3.36 &
  57.54±3.45 &
  60.52±7.14 &
  74.48±4.59 &
  85.68±5.54 &
  77.37±9.34 &
  79.68±7.47 &
  88.33±4.05 &
  93.78±2.79 &
  95.74±2.17 &
  38.71±3.93 &
  84.84±2.93 &
  91.43±4.06 &
  93.96±1.75 &
  \cellcolor[RGB]{251, 228, 213}95.99±\textbf{0.87} \\
\multicolumn{1}{c|}{20} &
 \cellcolor[RGB]{238, 150, 189}  &
  15.99±6.08 &
  29.28±6.57 &
  49.64±6.38 &
  9.23±6.53 &
  86.04±5.22 &
  65.00±12.03 &
  60.01±10.24 &
  70.79±13.22 &
  94.41±2.05 &
  91.54±2.40 &
  39.60±11.57 &
  56.12±10.08 &
  91.50±3.97 &
  89.27±2.97 &
  \cellcolor[RGB]{251, 228, 213}95.83±\textbf{1.13} \\
\multicolumn{1}{c|}{21} &
 \cellcolor[RGB]{111, 106, 62}  &
  2.74±2.81 &
  0.31±\textbf{0.37} &
  5.46±2.41 &
  9.23±5.47 &
  30.40±14.14 &
  36.23±7.04 &
  30.07±9.48 &
  26.42±13.38 &
  74.55±14.13 &
  67.24±18.96 &
  33.23±14.21 &
  26.55±9.99 &
  67.75±14.14 &
  48.35±11.04 &
 \cellcolor[RGB]{251, 228, 213} 87.13±3.70\\
\multicolumn{1}{c|}{22} &
 \cellcolor[RGB]{133, 198, 124}  &
  13.02±5.54 &
  20.45±9.51 &
  31.81±5.55 &
  38.88±9.48 &
  71.87±6.04 &
  66.06±7.21 &
  67.76±4.60 &
  73.67±5.89 &
  92.46±3.39 &
  89.81±7.36 &
  72.98±6.69 &
  68.55±8.48 &
  86.91±8.01 &
  91.43±2.58 &
  \cellcolor[RGB]{251, 228, 213}95.77±\textbf{1.94} \\ \hline \specialrule{0em}{3pt}{3pt} \hline
\multicolumn{2}{c|}{AA (\%) $\uparrow$} &
  41.83±0.68 &
  47.60±1.00 &
  51.98±1.34 &
  58.62±2.35 &
  73.56±2.29 &
  73.38±1.66 &
  72.87±1.46 &
  75.94±2.84 &
  92.81±0.86 &
  86.20±2.15 &
  71.09±2.12 &
  74.41±2.59 &
  89.33±1.25 &
  87.99±0.97 &
  \cellcolor[RGB]{251, 228, 213}94.39±\textbf{0.39} \\
\multicolumn{2}{c|}{OA (\%) $\uparrow$} &
  74.52±0.23 &
  78.35±0.25 &
  78.55±0.38 &
  81.93±0.41 &
  89.62±0.63 &
  88.95±0.58 &
  88.08±0.54 &
  90.72±0.88 &
  97.11±0.23 &
  95.96±0.82 &
  89.30±1.67 &
  89.22±1.13 &
  95.14±0.70 &
  95.53±0.31 &
  \cellcolor[RGB]{251, 228, 213}97.68±\textbf{0.16} \\
\multicolumn{2}{c|}{$k$*100 (\%) $\uparrow$} &
  66.75±0.33 &
  72.15±0.36 &
  72.56±0.51 &
  77.05±0.58 &
  86.91±0.79 &
  86.06±0.73 &
  84.97±0.69 &
  88.28±1.12 &
  96.36±0.29 &
  94.90±1.04 &
  86.72±1.76 &
  86.40±1.41 &
  93.89±0.88 &
  94.35±0.39 &
  \cellcolor[RGB]{251, 228, 213}97.08±\textbf{0.20} \\
\multicolumn{2}{c|}{FWIoU (\%) $\uparrow$} &
  20.69±0.61 &
  26.59±0.89 &
  30.93±1.17 &
  72.27±0.68 &
  83.03±0.91 &
  82.08±0.74 &
  80.80±0.76 &
  84.56±1.26 &
  94.59±0.43 &
  92.74±1.23 &
  82.90±1.43 &
  82.04±1.66 &
  91.31±1.09 &
  91.96±0.60 &
  \cellcolor[RGB]{251, 228, 213}95.60±\textbf{0.28} \\
\multicolumn{2}{c|}{MIoU (\%) $\uparrow$} &
  22.65±\textbf{0.67} &
  27.91±1.03 &
  33.25±1.28 &
  47.11±1.57 &
  63.91±2.37 &
  63.12±1.81 &
  61.31±1.70 &
  66.31±2.95 &
  88.33±0.82 &
  79.68±3.38 &
  63.44±2.51 &
  64.23±3.01 &
  81.61±1.94 &
  81.49±1.34 &
  \cellcolor[RGB]{251, 228, 213}90.19±\textbf{0.67} \\ \hline
\end{tabular}
}
\label{table_HongHu}
\end{table*}

\begin{table*}[htbp]
\Huge
\centering
\caption{Quantitative performance of different classification methods in terms of OA, AA, $k$, FWIoU, MIoU, as well as the accuracies for each class on the MUUFL dataset with 3 \% training sample. The best results are colored shadow and optimal standard deviation in bold. Results on Exp1.}
\resizebox{\textwidth}{!}{
\begin{tabular}{c|c|ccc|cccccc|cc|cc|c|c}
\hline
\multirow{2}{*}{\makecell[c]{Class \\ No.}} &
  \multirow{2}{*}{Color} &
  \multicolumn{3}{c|}{Conventional Classifier} &
  \multicolumn{6}{c|}{CNNs Based networks} &
  \multicolumn{2}{c|}{GAN Based Networks} &
  \multicolumn{2}{c|}{Transformer Based Networks} &
  \makecell[c]{Unsupervised \\ Network} &
  \multicolumn{1}{c}{\multirow{2}{*}{\textbf{DiffCRN}}} \\ \cline{3-16}
\multicolumn{1}{c|}{} &
   &
  \multicolumn{1}{c}{KNN} &
  RF &
  SVM &
  1D-CNN \cite{1DCNN1} &
  2D-CNN \cite{2DCNN} &
  3D-CNN \cite{Hamida3dCNN} &
  HybridSN \cite{HybridSN}&
  SSRN \cite{SSRN}&
  SS-ConvNeXt \cite{SS-ConvNeXt}&
  MTGAN \cite{Hangrl2020_TGRS}&
  ADGAN \cite{WangTGRS_2021}&
  SSFTT \cite{SSFTT}&
  SSTN \cite{SSTN}&
  SC-SS-MTr \cite{Huang2023_TGRS}&
  \multicolumn{1}{c}{} \\ \hline \specialrule{0em}{3pt}{3pt} \hline
\multicolumn{1}{c|}{1} &
  \cellcolor[RGB]{83, 171, 72} &
  92.25±0.34 &
  93.22±0.39 &
  91.27±0.56 &
  94.99±0.55 &
  96.76±0.38 &
  96.49±1.02 &
  95.04±1.18 &
  95.54±1.55 &
  95.75±0.33 &
  96.99±\textbf{0.30} &
  93.17±5.87 &
  96.36±0.91 &
  95.86±0.91 &
  \cellcolor[RGB]{251, 228, 213}97.76±0.33 &
  97.72±0.34 \\
\multicolumn{1}{c|}{2} &
  \cellcolor[RGB]{137, 186, 67} &
  71.75±3.04 &
  71.04±3.82 &
  65.73±1.86 &
  71.96±5.81 &
  75.72±4.92 &
  74.12±4.45 &
  82.00±3.84 &
  86.55±4.18 &
  85.90±0.15 &
  \cellcolor[RGB]{251, 228, 213}88.21±\textbf{1.40} &
  55.77±19.93 &
  77.72±11.79 &
  83.50±5.36 &
  85.34±2.62 &
  85.75±2.43 \\
\multicolumn{1}{c|}{3} &
  \cellcolor[RGB]{217, 142, 52} &
  74.15±2.05 &
  77.22±2.22 &
  70.66±3.02 &
  78.63±3.87 &
  79.73±3.95 &
  78.63±3.11 &
  84.04±2.42 &
  87.99±3.08 &
  87.72±2.25 &
  89.47±1.57 &
  75.10±10.72 &
  83.15±2.43 &
  82.62±3.74 &
  \cellcolor[RGB]{251, 228, 213}91.56±\textbf{1.09} &
  88.38±1.47 \\
\multicolumn{1}{c|}{4} &
  \cellcolor[RGB]{60, 131, 69} &
  71.53±3.84 &
  73.14±2.28 &
  59.92±14.82 &
  82.79±2.86 &
  81.69±4.05 &
  77.71±4.55 &
  90.82±4.45 &
  90.28±3.77 &
  90.26±1.59 &
  92.83±2.33 &
  78.94±5.05 &
  84.52±8.65 &
  84.49±6.93 &
  \cellcolor[RGB]{251, 228, 213}94.68±2.17 &
  91.15±\textbf{1.91} \\
\multicolumn{1}{c|}{5} &
  \cellcolor[RGB]{144, 82, 54} &
  90.66±0.94 &
  88.29±2.00 &
  84.98±1.18 &
  90.60±1.27 &
  91.15±1.80 &
  87.31±1.67 &
  91.39±1.42 &
  95.17±0.79 &
  93.64±1.08 &
  95.32±\textbf{0.46} &
  65.81±26.94 &
  85.17±19.80 &
  91.86±2.47 &
  86.04±0.78 &
  \cellcolor[RGB]{251, 228, 213}95.57±0.81 \\
\multicolumn{1}{c|}{6} &
  \cellcolor[RGB]{105, 188, 200} &
  73.42±\textbf{1.42} &
  72.01±3.90 &
  41.61±17.05 &
  78.72±2.32 &
  84.25±5.90 &
  79.96±9.69 &
  90.33±2.89 &
  93.16±7.00 &
  94.31±2.23 &
  87.09±8.75 &
  41.42±25.13 &
  84.67±13.53 &
  92.17±5.16 &
  87.30±5.99 &
  \cellcolor[RGB]{251, 228, 213}96.72±1.95 \\
\multicolumn{1}{c|}{7} &
  \cellcolor[RGB]{199, 176, 201} &
  63.88±2.91 &
  64.83±4.01 &
  54.57±3.37 &
  65.78±3.18 &
  83.21±3.72 &
  80.02±3.04 &
  78.82±4.45 &
  86.69±3.09 &
  84.62±3.22 &
  89.65±2.87 &
  72.79±8.39 &
  71.15±19.90 &
  84.94±3.52 &
  \cellcolor[RGB]{251, 228, 213}96.72±\textbf{1.53} &
  85.73±2.11 \\
\multicolumn{1}{c|}{8} &
  \cellcolor[RGB]{157, 87, 150} &
  76.81±1.13 &
  80.43±1.87 &
  82.72±1.55 &
  86.45±1.39 &
  92.53±1.19 &
  86.70±2.26 &
  95.22±1.05 &
  96.80±0.91 &
  96.07±0.63 &
  96.72±1.18 &
  84.41±9.79 &
  93.30±3.91 &
  95.11±0.96 &
  96.72±0.66 &
  \cellcolor[RGB]{251, 228, 213}97.28±\textbf{0.57} \\
\multicolumn{1}{c|}{9} &
  \cellcolor[RGB]{3, 14, 112} &
  34.56±4.94 &
  41.18±5.05 &
  38.60±9.11 &
  51.44±5.84 &
  57.48±4.17 &
  42.58±6.93 &
  56.06±8.77 &
  57.97±11.28 &
  36.76±6.36 &
  48.73±3.51 &
  16.02±9.90 &
  45.80±3.31 &
  40.17±10.38 &
  60.66±\textbf{4.15} &
  \cellcolor[RGB]{251, 228, 213}69.10±4.56 \\
\multicolumn{1}{c|}{10} &
  \cellcolor[RGB]{144, 82, 54} &
  2.70±2.84 &
  5.73±4.77 &
  \cellcolor[RGB]{251, 228, 213}42.64±17.36 &
  37.58±8.90 &
  15.45±3.45 &
  10.17±6.98 &
  23.26±9.48 &
  19.83±11.52 &
  2.20±2.94 &
  3.44±4.94 &
  1.12±\textbf{2.51} &
  11.07±5.09 &
  7.13±7.01 &
  17.46±6.86 &
  21.11±6.12 \\
\multicolumn{1}{c|}{11} &
  \cellcolor[RGB]{111, 106, 62} &
  71.11±8.41 &
  78.85±11.07 &
  26.40±11.97 &
  72.41±\textbf{7.88} &
  56.44±15.50 &
  46.32±11.97 &
  77.85±19.65 &
  77.13±15.92 &
  71.26±15.35 &
  68.92±14.57 &
  19.77±15.76 &
  68.35±8.54 &
  77.59±11.06 &
  81.85±8.84 &
  \cellcolor[RGB]{251, 228, 213}81.99±9.51 \\ \hline \specialrule{0em}{3pt}{3pt} \hline
\multicolumn{2}{c|}{AA (\%) $\uparrow$} &
  65.71±\textbf{0.75} &
  67.81±1.36 &
  59.92±5.03 &
  73.76±1.53 &
  74.13±4.46 &
  69.09±1.11 &
  78.62±1.94 &
  80.65±5.74 &
  76.23±1.98 &
  77.94±2.04 &
  54.97±7.36 &
  72.84±6.68 &
  75.95±1.89 &
  80.65±1.46 &
  \cellcolor[RGB]{251, 228, 213}82.77±1.08 \\
\multicolumn{2}{c|}{OA (\%) $\uparrow$} &
  82.49±\textbf{0.20} &
  83.54±0.32 &
  80.51±1.29 &
  86.32±0.29 &
  89.06±0.64 &
  86.74±0.57 &
  89.96±0.56 &
  92.07±0.78 &
  91.09±0.56 &
  92.85±0.24 &
  79.05±6.35 &
  88.15±4.75 &
  89.89±0.58 &
  92.93±0.33 &
  \cellcolor[RGB]{251, 228, 213}93.44±0.28 \\
\multicolumn{2}{c|}{$k$*100 (\%) $\uparrow$} &
  76.83±\textbf{0.25} &
  78.85±0.41 &
  74.30±1.68 &
  82.01±0.40 &
  85.92±0.85 &
  82.86±0.69 &
  87.15±0.71 &
  89.85±1.00 &
  88.42±0.77 &
  90.81±0.30 &
  76.53±9.05 &
  84.43±7.07 &
  86.88±0.78 &
  90.96±0.44 &
  \cellcolor[RGB]{251, 228, 213}91.56±0.36 \\
\multicolumn{2}{c|}{FWIoU (\%) $\uparrow$} &
  44.37±1.66 &
  48.50±2.00 &
  37.98±5.56 &
  77.08±0.51 &
  81.64±1.00 &
  78.18±0.66 &
  83.17±0.81 &
  86.40±1.23 &
  84.54±0.93 &
  87.39±0.38 &
  71.19±8.91 &
  80.09±7.49 &
  83.04±0.94 &
  87.61±0.55 &
  \cellcolor[RGB]{251, 228, 213}88.44±\textbf{0.44} \\
\multicolumn{2}{c|}{MIoU (\%) $\uparrow$} &
  46.71±1.48 &
  50.50±2.07 &
  39.91±4.56 &
  64.04±1.72 &
  65.97±1.84 &
  60.41±\textbf{0.96} &
  69.16±1.50 &
  73.18±3.09 &
  68.25±1.77 &
  70.92±1.78 &
  50.37±7.35 &
  64.13±6.95 &
  66.01±2.75 &
  74.51±1.34 &
  \cellcolor[RGB]{251, 228, 213}76.26±1.06 \\ \hline
\end{tabular}
}
\label{table_MUUFL_result}
\end{table*}

\begin{figure*}[htbp]
    \centering
    \includegraphics[width=0.99\textwidth]{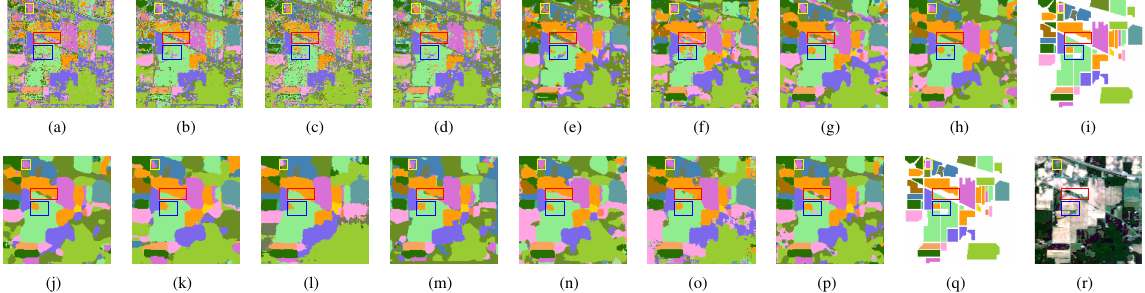}
    \\ \quad
    \centerline{\includegraphics[width=0.99\textwidth]{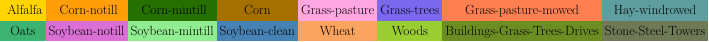}}
    \caption{Classification maps of the Indian Pines dataset. (a) KNN (OA=68.20\%). (b) RF (OA=74.85\%). (c) SVM (OA=74.63\%). (d) 1D-CNN (OA=73.75\%). (e) 2D-CNN (OA=88.83\%). (f) 3D-CNN (OA=87.90\%). (g) HybridSN (OA=96.18\%). (h) SSRN (OA=97.81\%). (i) Ground Truth map. (j) SS-ConvNeXt (OA=98.20\%). (k) MTGAN (OA=97.20\%). (l) ADGAN (OA=97.63\%). (m) SSFTT (OA=95.12\%). (n) SSTN (OA=97.15\%). (o) SC-SS-MTr (OA=97.33\%). (p) DiffCRN (OA=99.33\%). (q) label-domain classification map of DiffCRN. (r) RGB Composite Image. Results on Exp1. The box in red, blue, yellow shows the region of interest, which represents the category boundary preserving capability of different methods.}
    \label{IN_classification_map}
\end{figure*}

\subsubsection{\textbf{Visual Evaluation}}
\begin{figure*}[htbp]
    \centering
    \includegraphics[width=0.99\textwidth]{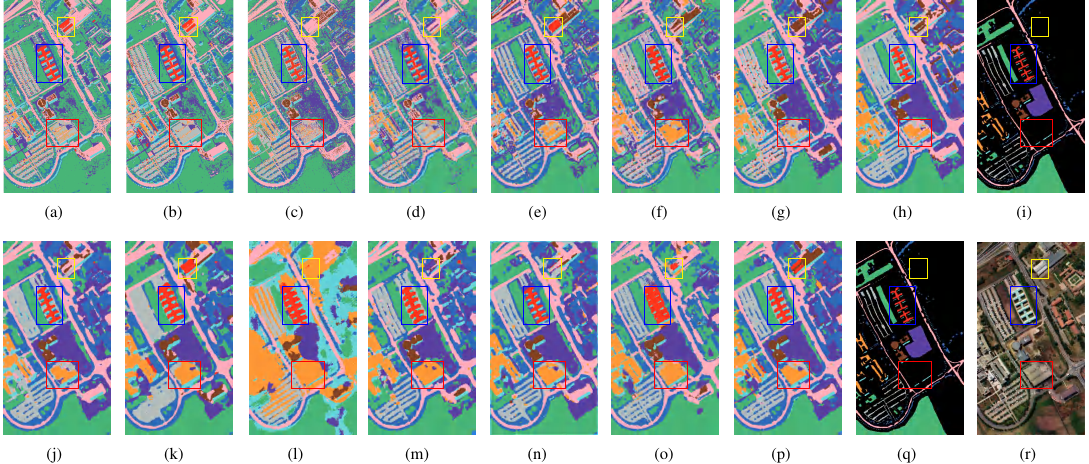}
    \\ \quad
    \centerline{\includegraphics[width=0.99\textwidth]{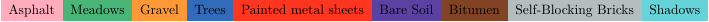}}
    \caption{Classification maps of the Pavia University dataset. (a) KNN (OA=83.41\%). (b) RF (OA=85.53\%). (c) SVM (OA=88.80\%). (d) 1D-CNN (OA=84.37\%). (e) 2D-CNN (OA=94.88\%). (f) 3D-CNN (OA=94.82\%). (g) HybridSN (OA=96.93\%). (h) SSRN (OA=98.58\%). (i) Ground Truth map. (j) SS-ConvNeXt (OA=98.79\%).  (k) MTGAN (OA=99.03\%). (l) ADGAN (OA=71.35\%). (m) SSFTT (OA=88.96\%). (n) SSTN (OA=97.46\%). (o) SC-SS-MTr (OA=99.12\%). (p) DiffCRN (OA=99.70\%). (q) label-domain classification map of DiffCRN. (r) RGB Composite Image. Results on Exp1. The box in red, blue, yellow shows the region of interest which represents the category boundary preserving capability of different methods.}
\label{PU_classification_map}
\end{figure*}

\begin{figure*}[htbp]
    \centering
    \includegraphics[width=0.99\textwidth]{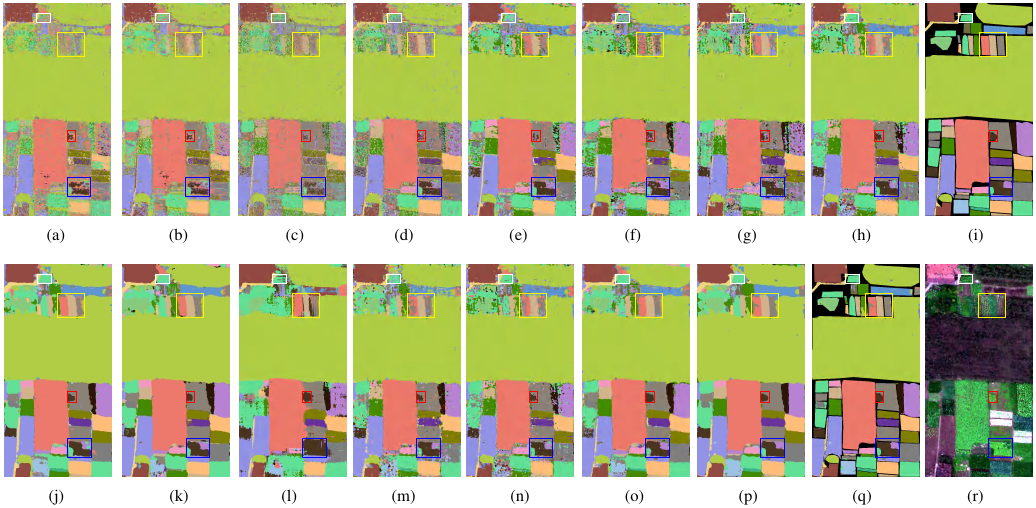}
    \\ \quad
    \centerline{\includegraphics[width=0.99\textwidth]{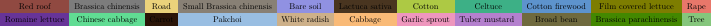}}
    \caption{Classification maps of the WHU-Hi-HongHu dataset. (a) KNN (OA=74.52\%). (b) RF (OA=78.35\%). (c) SVM (OA=78.55\%). (d) 1D-CNN (OA=81.93\%). (e) 2D-CNN (OA=89.62\%). (f) 3D-CNN (OA=88.95\%). (g) HybridSN (OA=88.08\%). (h) SSRN (OA=90.72\%). (i) Ground Truth map. (j) SS-ConvNeXt (OA=97.11\%). (k) MTGAN (OA=95.96\%). (l) ADGAN (OA=89.30\%). (m) SSFTT (OA=89.22\%). (n) SSTN (OA=95.14\%). (o) SC-SS-MTr (OA=95.53\%). (p) DiffCRN (OA=97.68\%). (q) label-domain classification map of DiffCRN. (r) RGB Composite Image. Results on Exp1. The box in red, blue, yellow and white shows the region of interest which represents the category boundary preserving capability of different methods.}
    \label{HH_classification_map}
\end{figure*}

\begin{figure*}[htbp]
    \centering
    \includegraphics[width=0.99\textwidth]{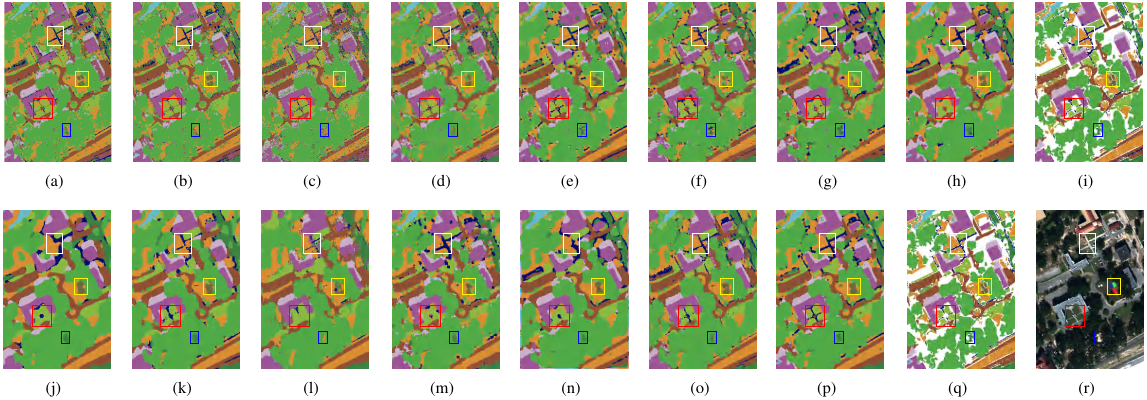}
    \\ \quad
    \centerline{\includegraphics[width=0.99\textwidth]{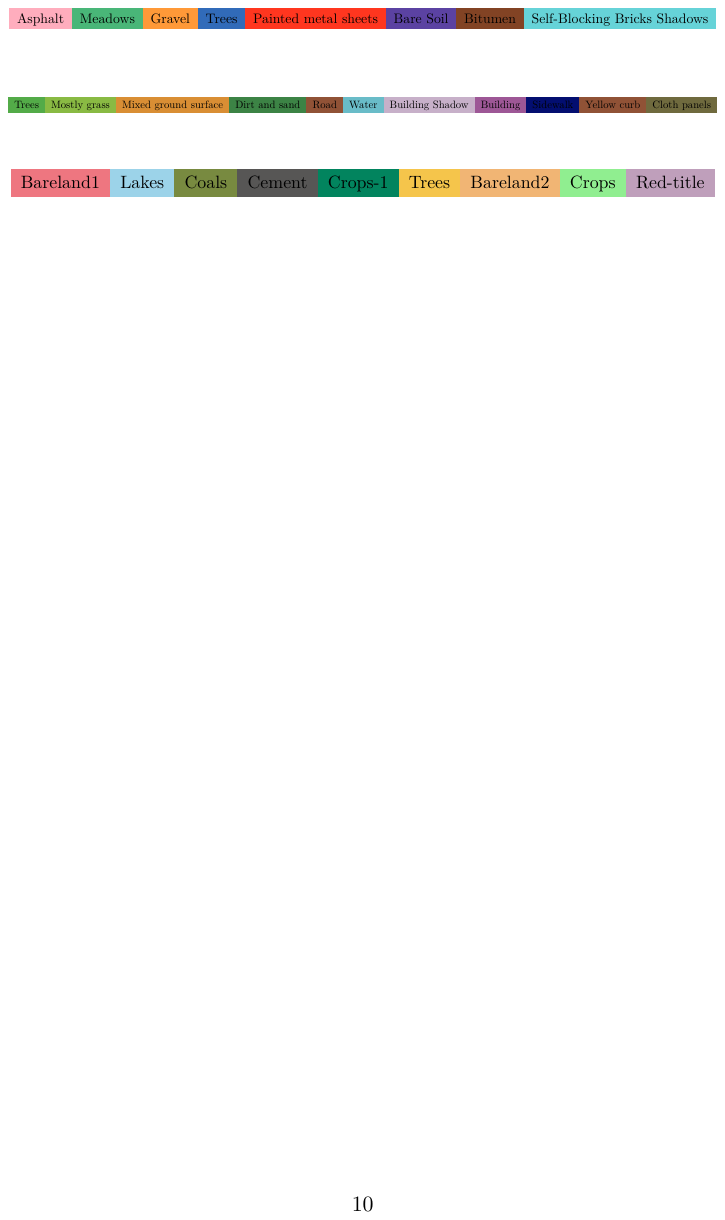}}
    \caption{Classification maps of the MUUFL dataset. (a) KNN (OA=82.49\%). (b) RF (OA=83.54\%). (c) SVM (OA=80.51\%). (d) 1D-CNN (OA=86.32\%). (e) 2D-CNN (OA=89.06\%). (f) 3D-CNN (OA=86.74\%). (g) HybridSN (OA=89.46\%). (h) SSRN (OA=92.07\%). (i) Ground Truth map. (j)SS-ConvNeXt (OA=91.09\%). (k) MTGAN (OA=92.85\%). (l) ADGAN (OA=79.05\%). (m) SSFTT (OA=88.15\%). (n) SSTN (OA=89.89\%). (o) SC-SS-MTr (OA=92.93\%). (p) DiffCRN (OA=93.44\%). (q) label-domian classification map of DiffCRN. (r) RGB Composite Image. Results on Exp1. The box in red, blue, yellow and white shows the region of interest which represents the category boundary preserving capability of different methods.}
\label{MUUFL_classification_map}
\end{figure*}

From the visualization perspective, \autoref{IN_classification_map} - \autoref{MUUFL_classification_map} give the image domain classification results of different approaches on the four datasets. Our goal is to perform a qualitative evaluation of the compared methods. Overall, the visual effects produced by DiffCRN are better than other methods. Conventional classifiers, such as KNN, RF, and SVM, and 1D-CNN provide classification maps with salt and pepper noise around the boundary areas because they only exploit spectral information. Classic backbone network such as HybridSN and SSRN improve this phenomenon and achieve better visual performance. MTGAN and ADGAN provide smoother boundary due to the encoder-decoder architecture, resulting in lossing spatial information. Transformer based methods, e.g., SSTN can extract more abstract information in sequential representation,
so it provides better classification maps than classic backbone network. The classification map generated by SC-SS-MTr provides discrete raster grid points maybe due to the masked image modeling mechanism of SC-SS-MTr in pre-training stage resulting in some error reconstruction of raster. Refernece to RGB composite image, as shown in box in red, blue, yellow and white, DiffCRN characterizes texture and edge details better than other techniques.

\begin{figure*}[htbp]
  \centerline{
  {\includegraphics[width=0.15\textwidth]{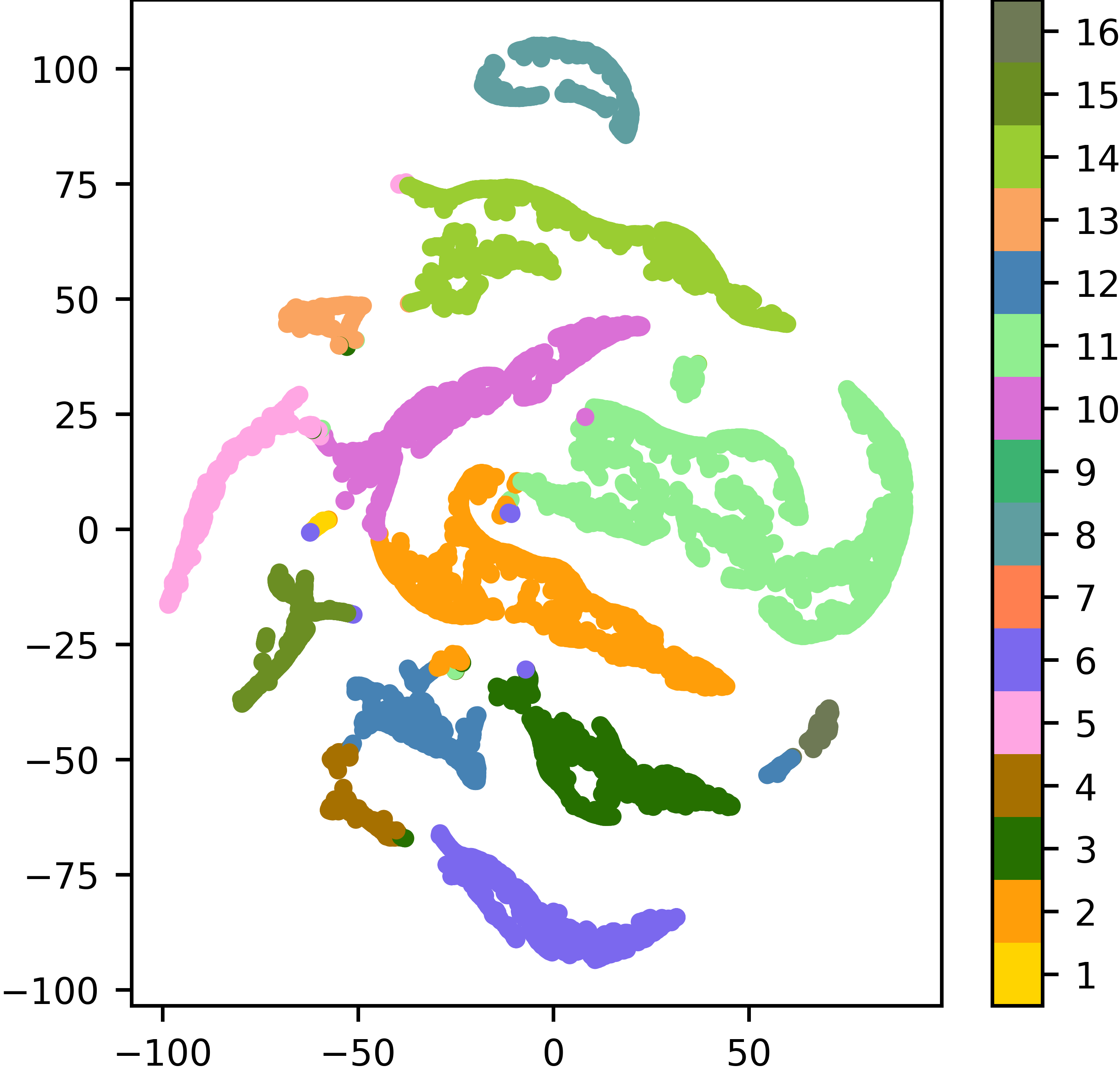}}
  {\includegraphics[width=0.15\textwidth]{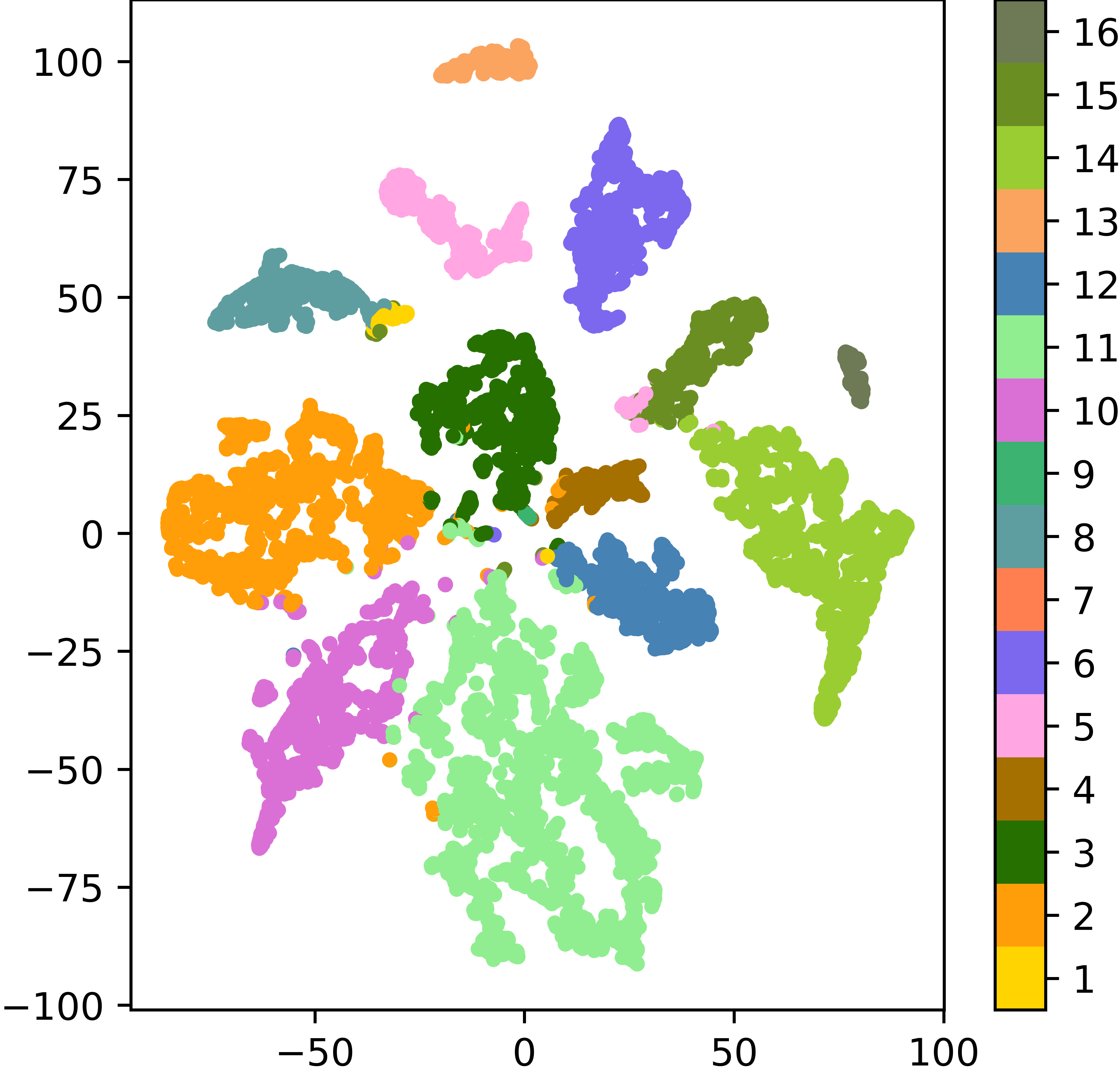}}
  {\includegraphics[width=0.15\textwidth]{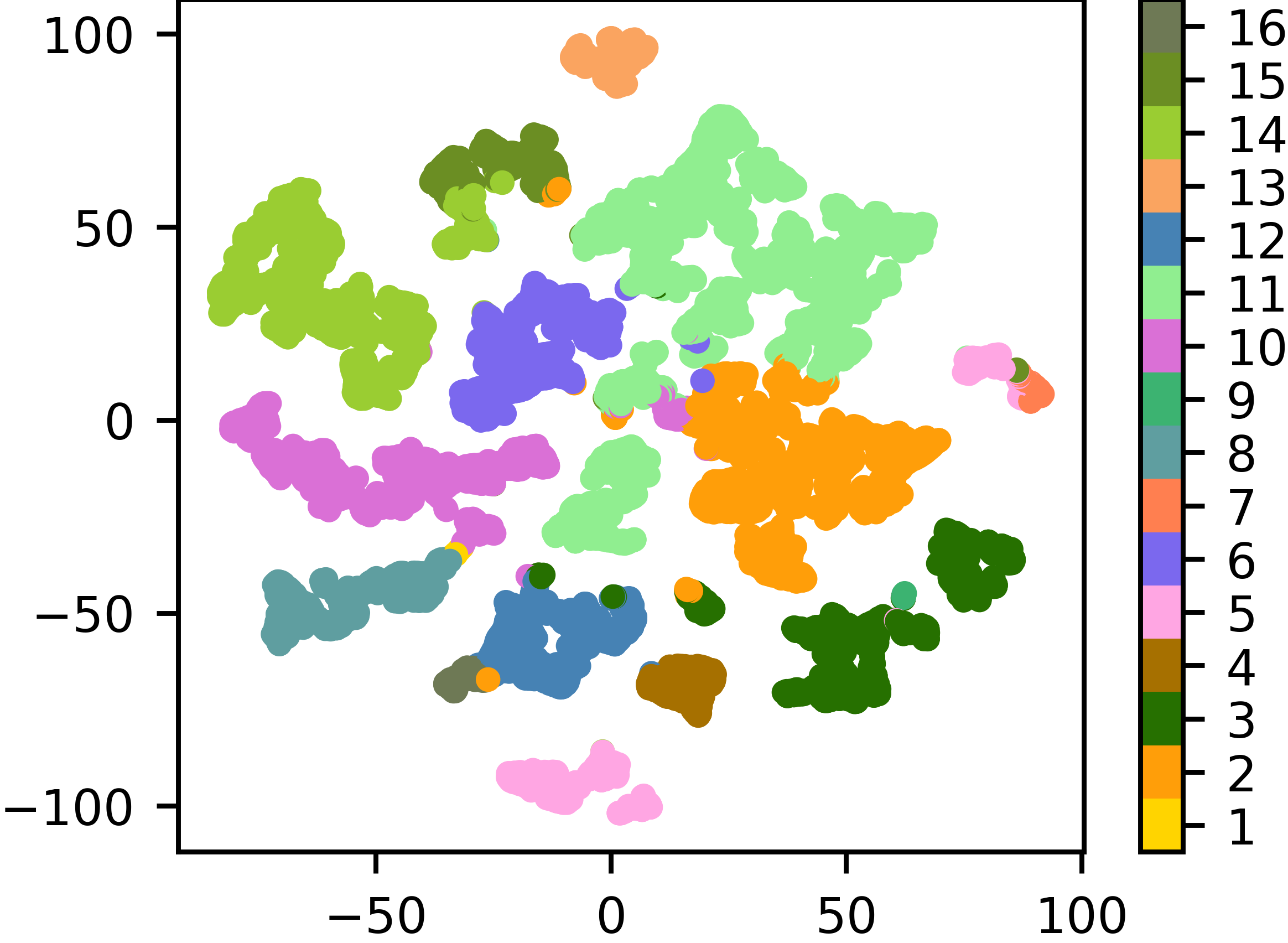}}
  {\includegraphics[width=0.15\textwidth]{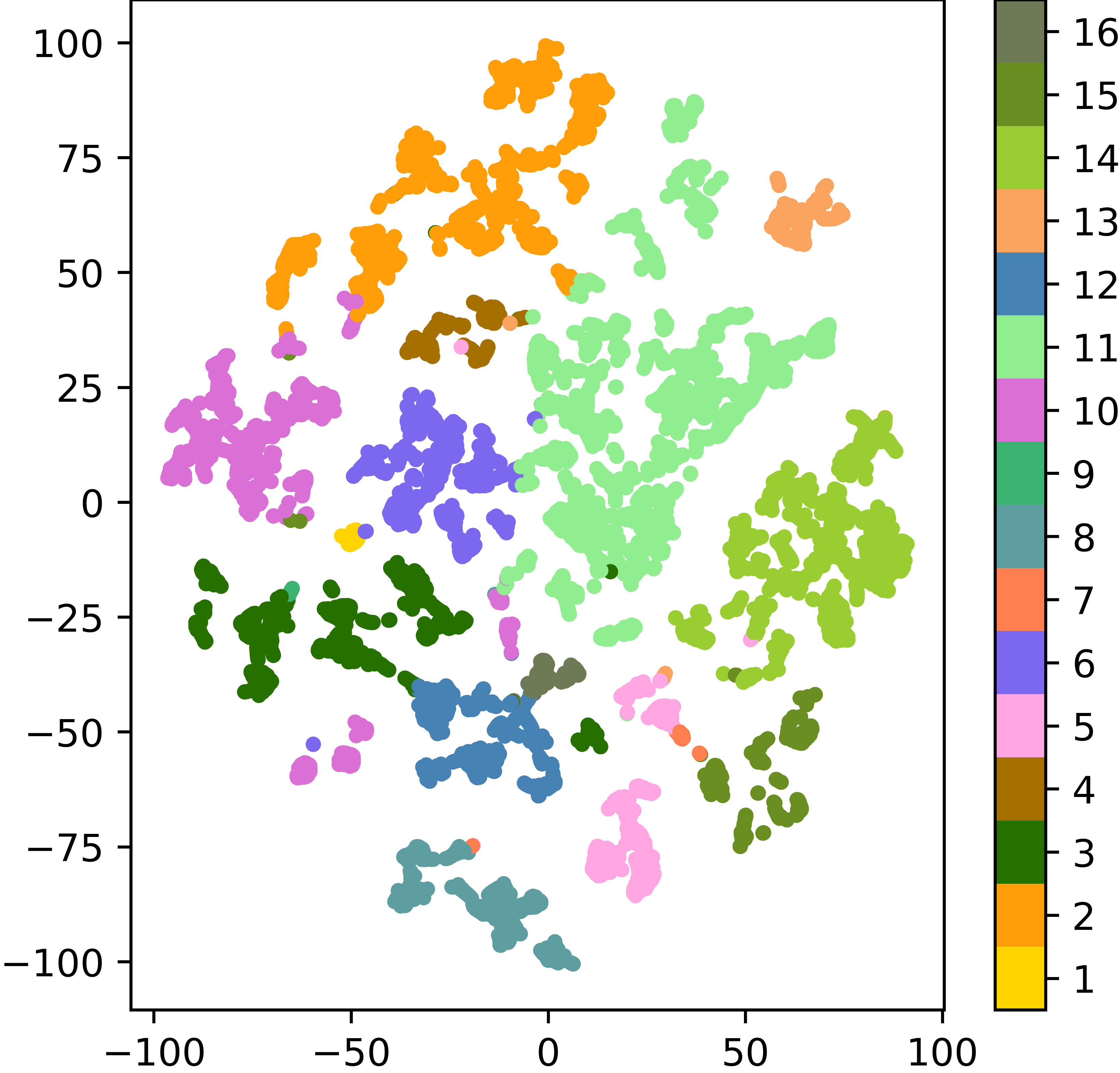}}
  {\includegraphics[width=0.15\textwidth]{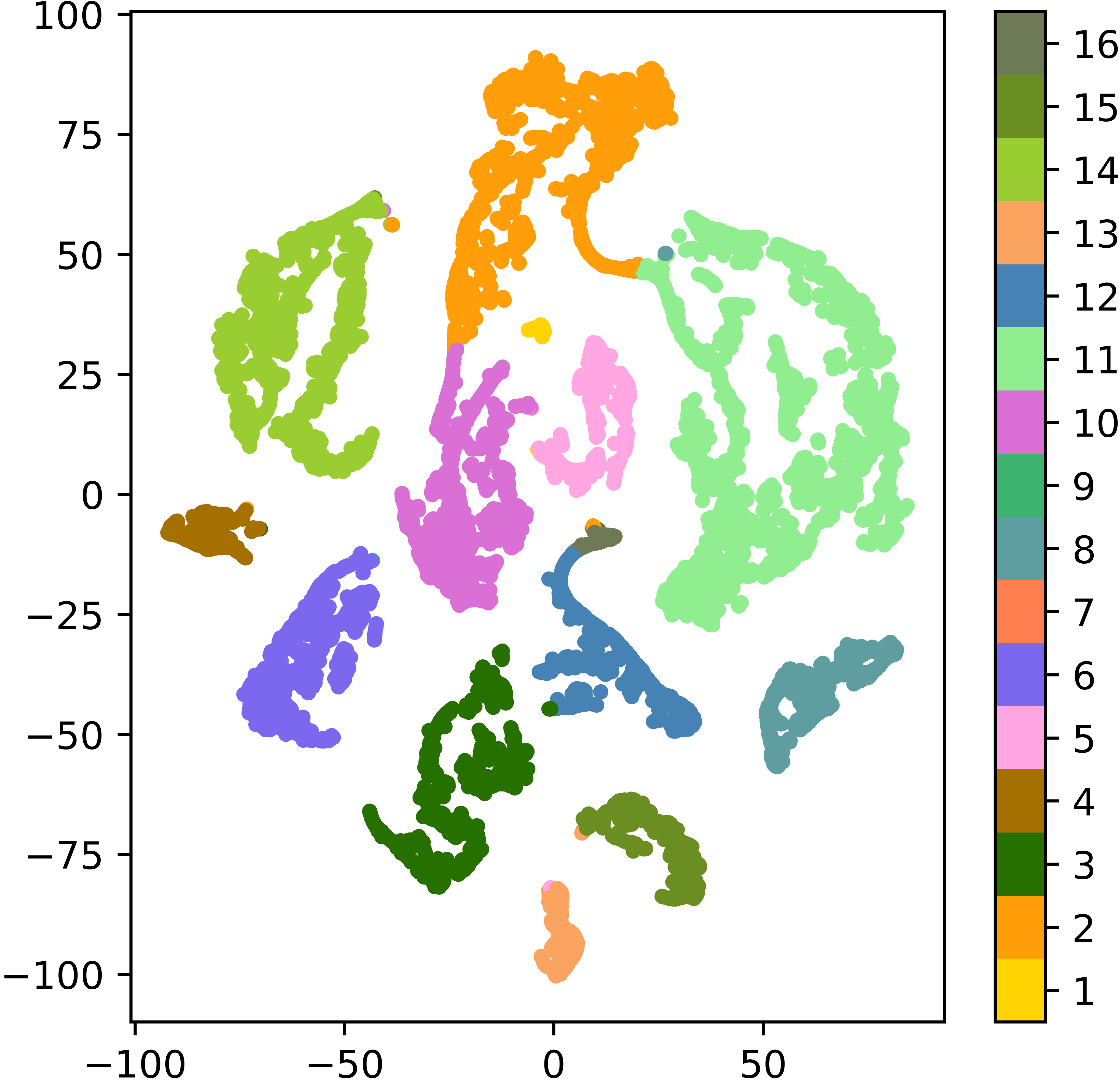}}
  {\includegraphics[width=0.15\textwidth]{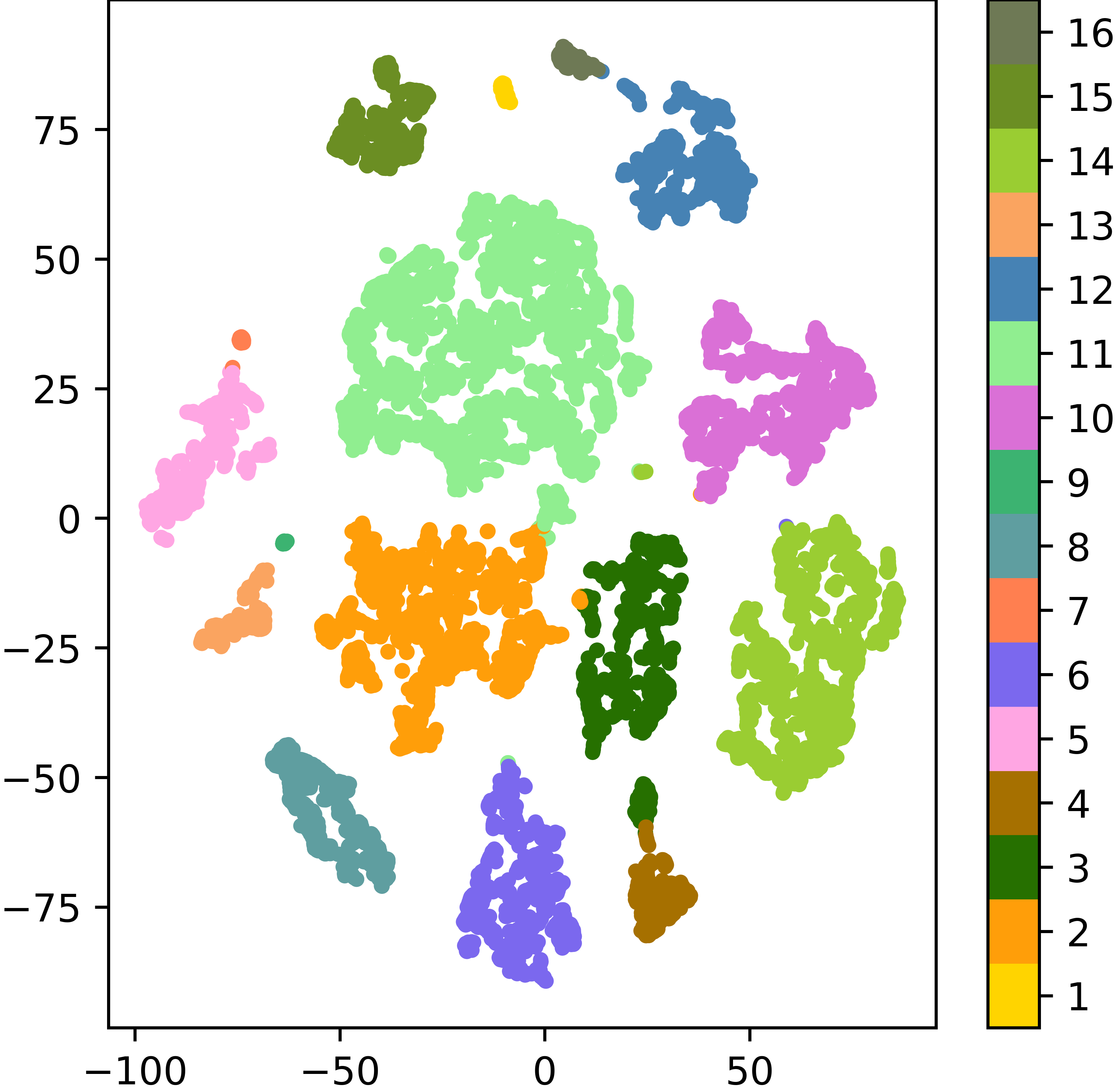}}
  }
  \vspace{3pt}
  \centerline{
  {\includegraphics[width=0.15\textwidth]{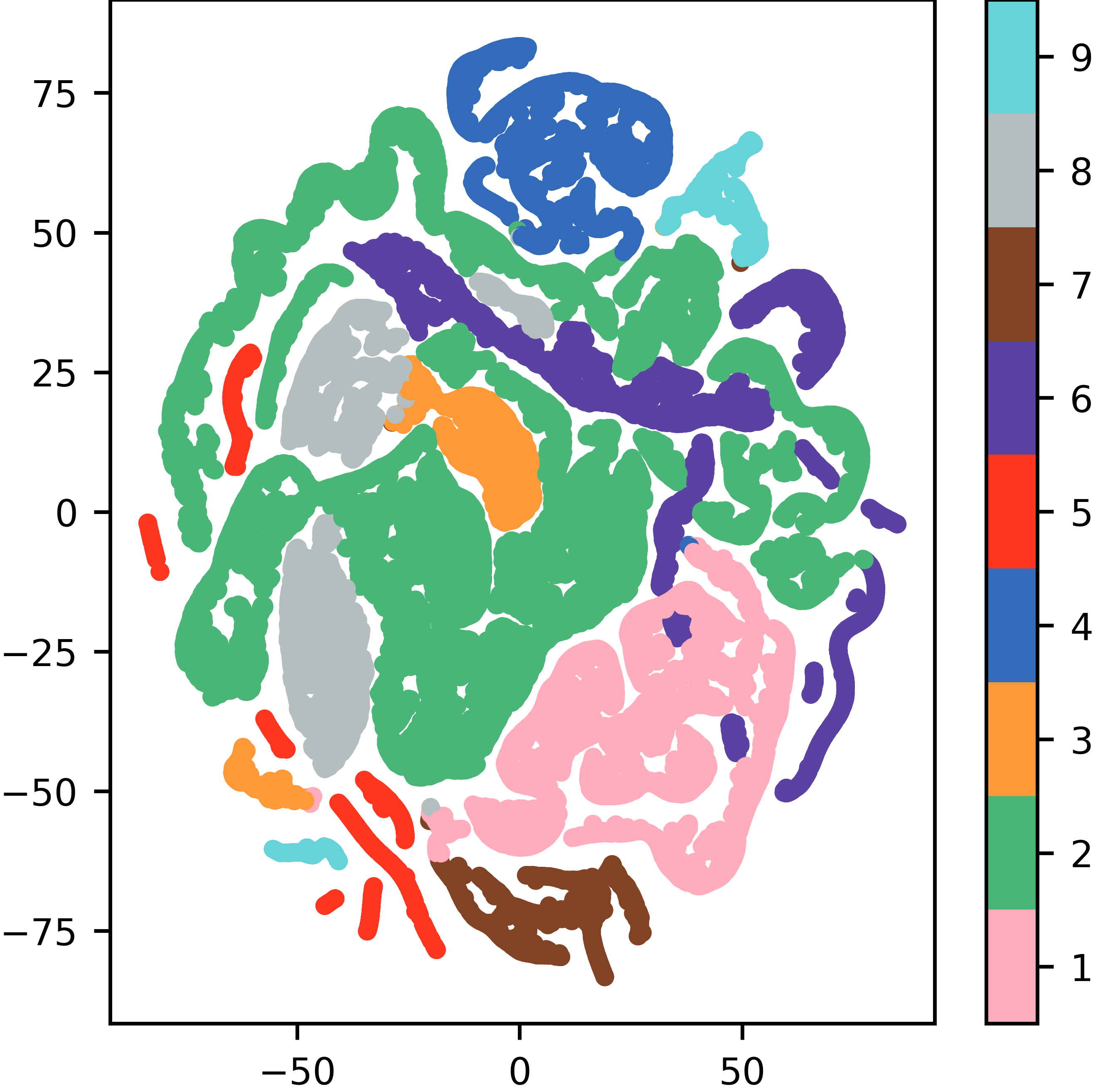}}
  {\includegraphics[width=0.15\textwidth]{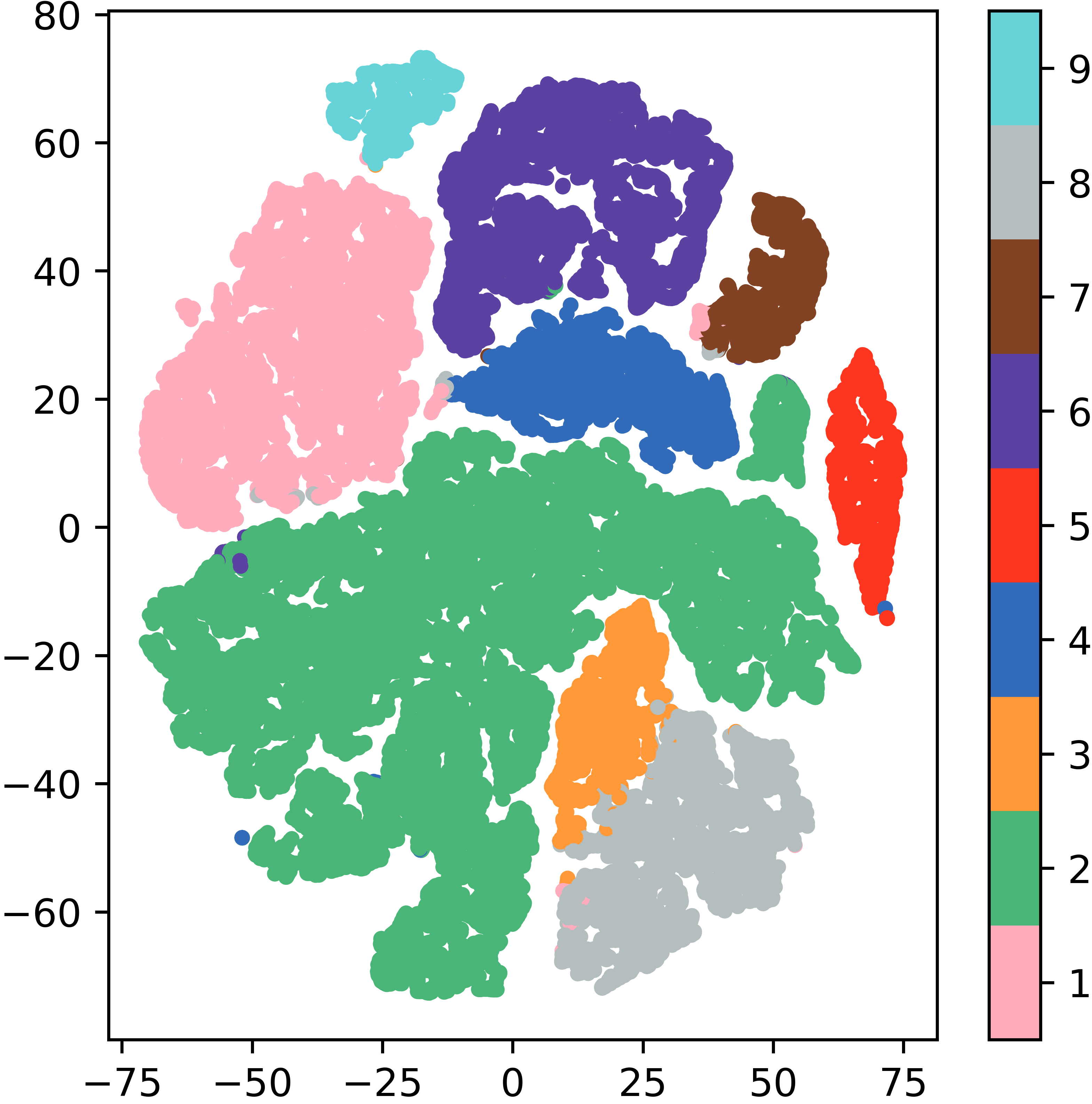}}
  {\includegraphics[width=0.15\textwidth]{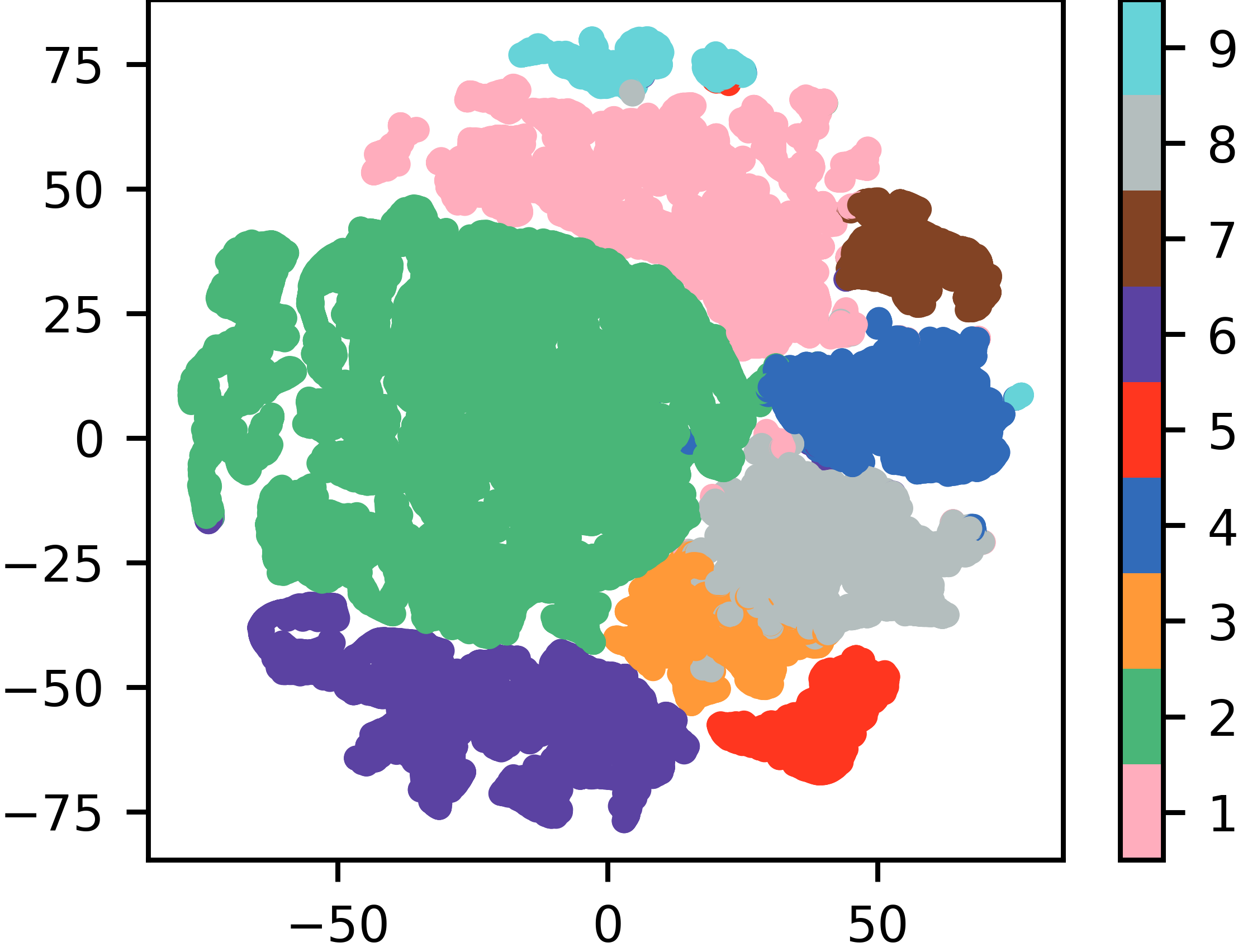}}
  {\includegraphics[width=0.15\textwidth]{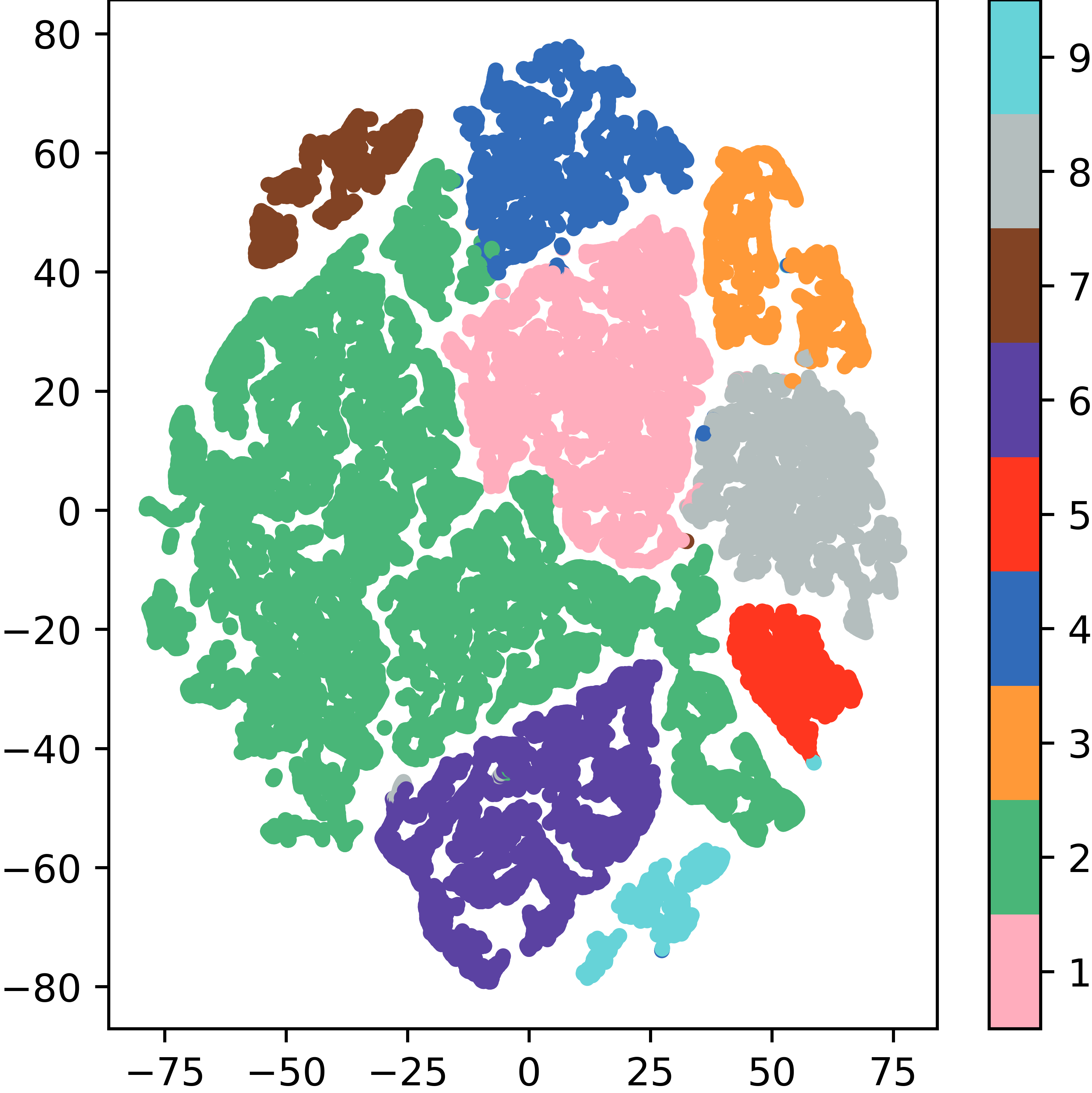}}
  {\includegraphics[width=0.15\textwidth]{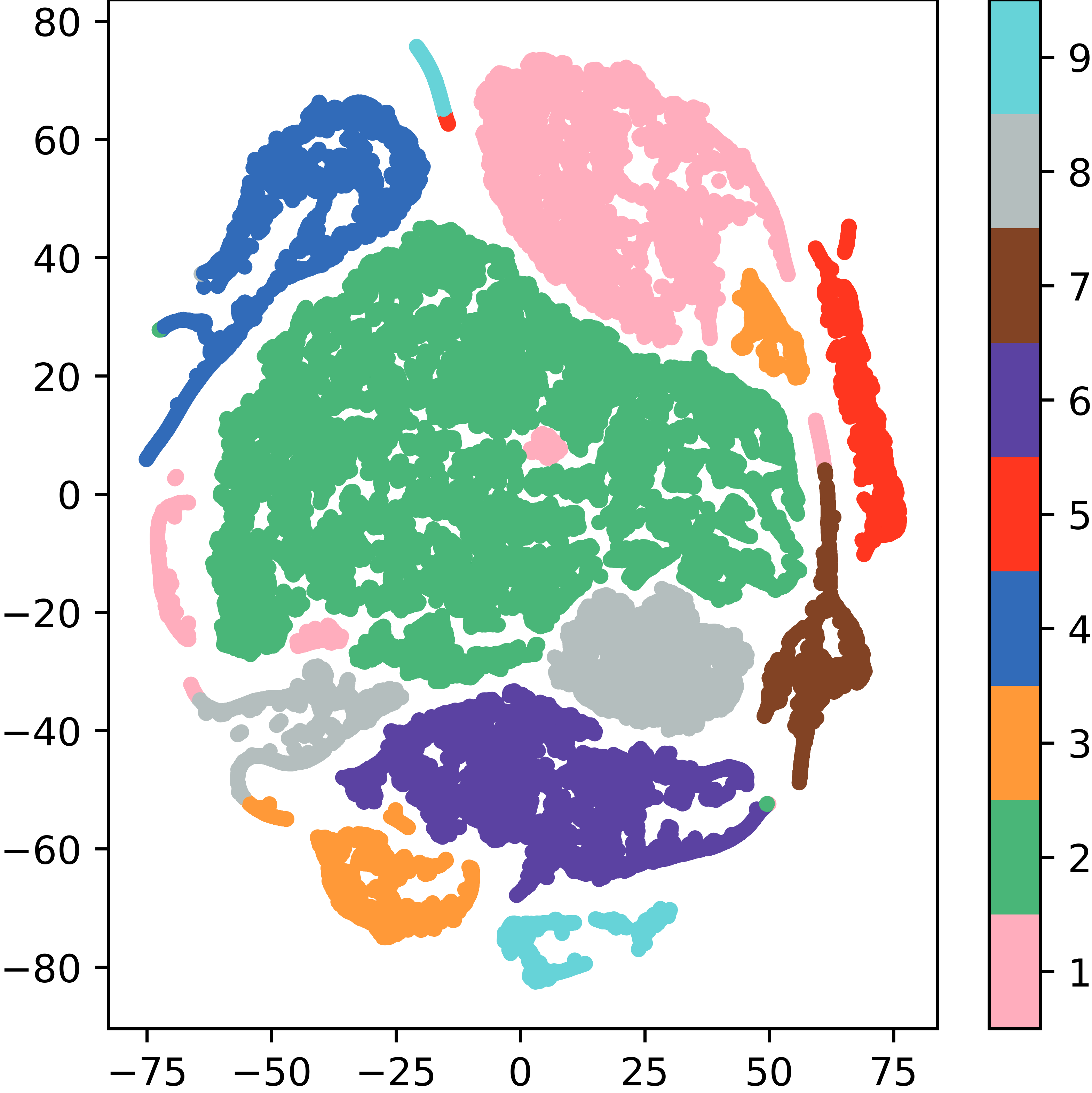}}
  {\includegraphics[width=0.15\textwidth]{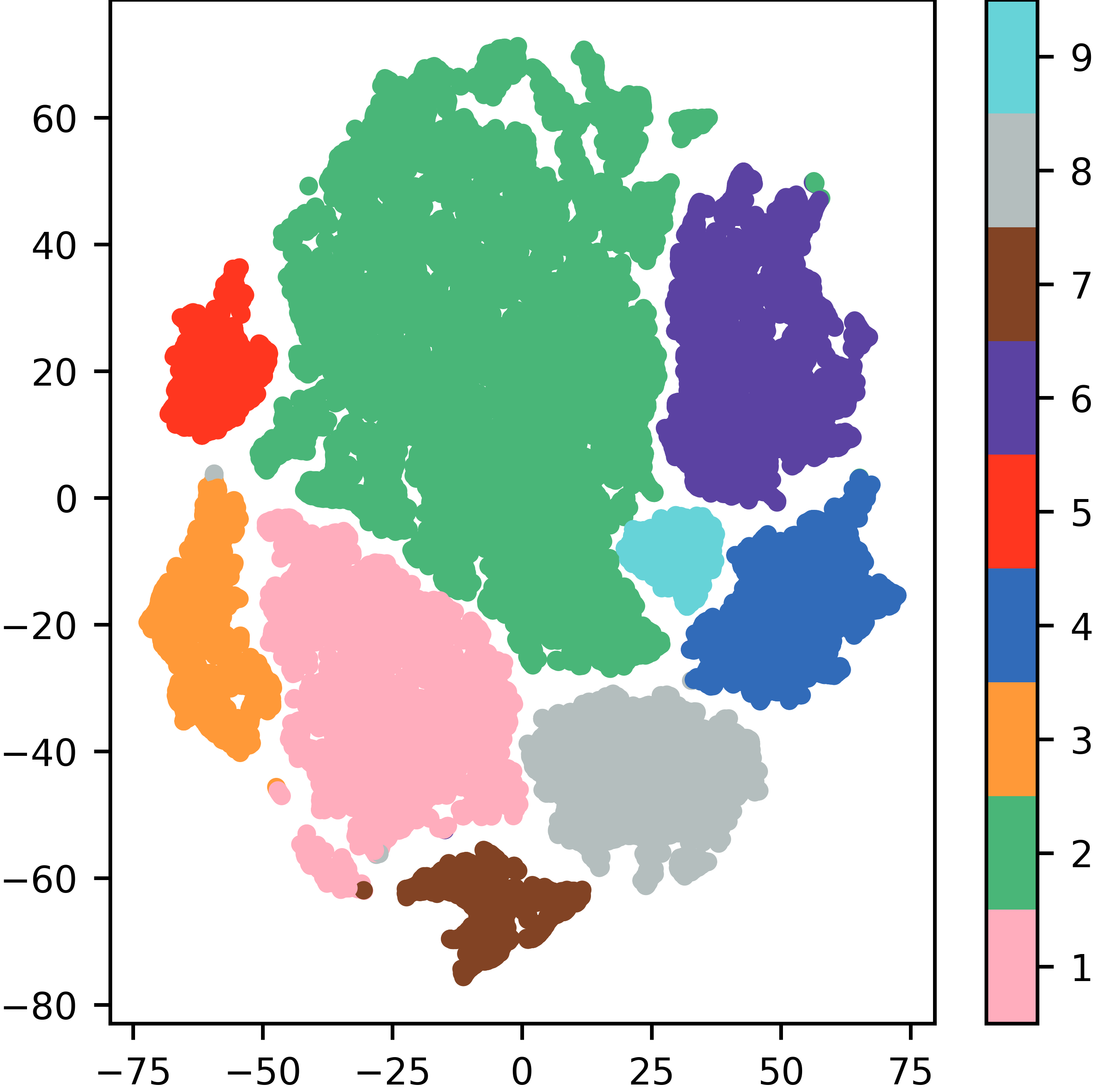}}
  }
    \vspace{3pt}
  \centerline{
  {\includegraphics[width=0.15\textwidth]{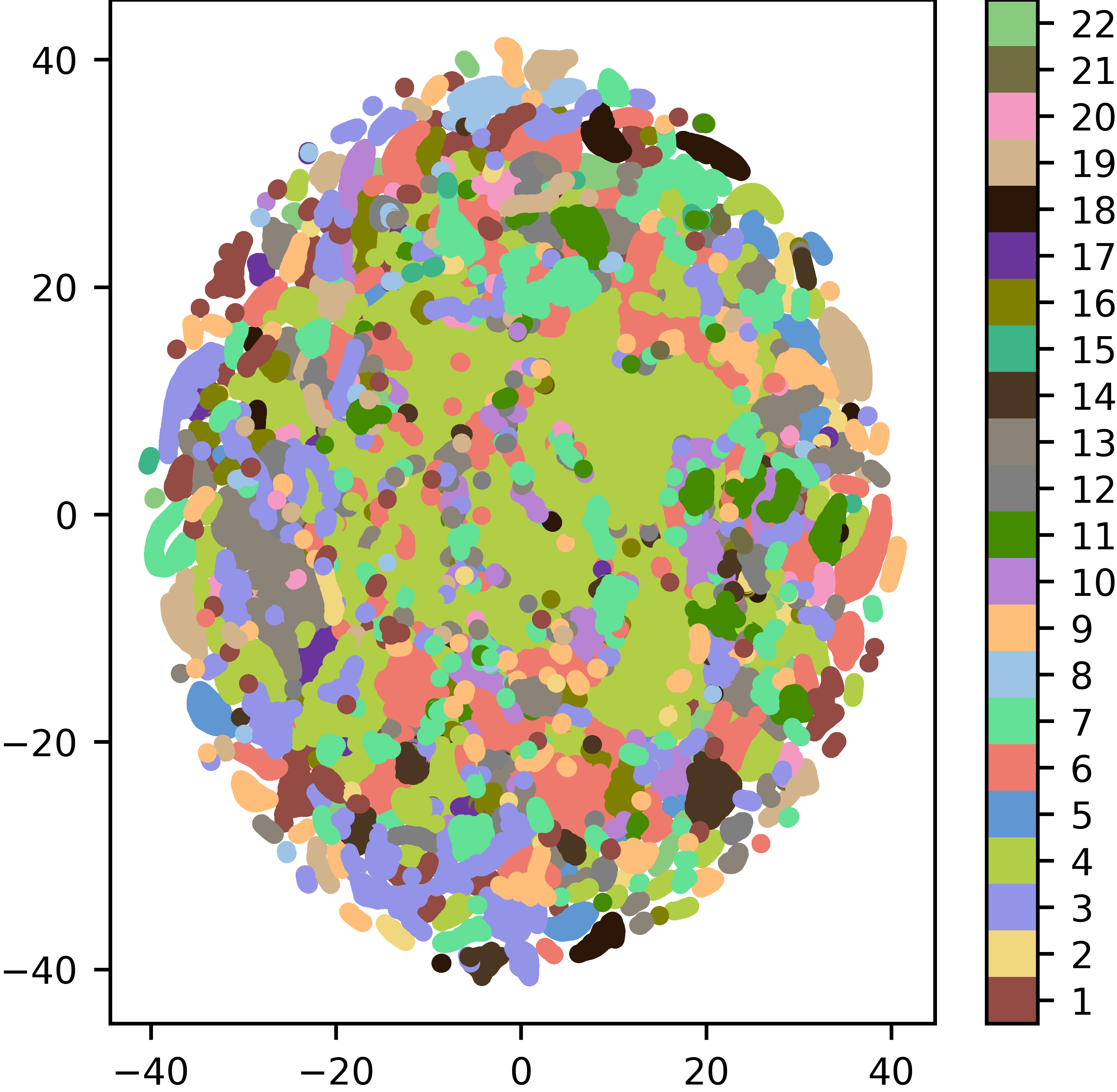}}
  {\includegraphics[width=0.15\textwidth]{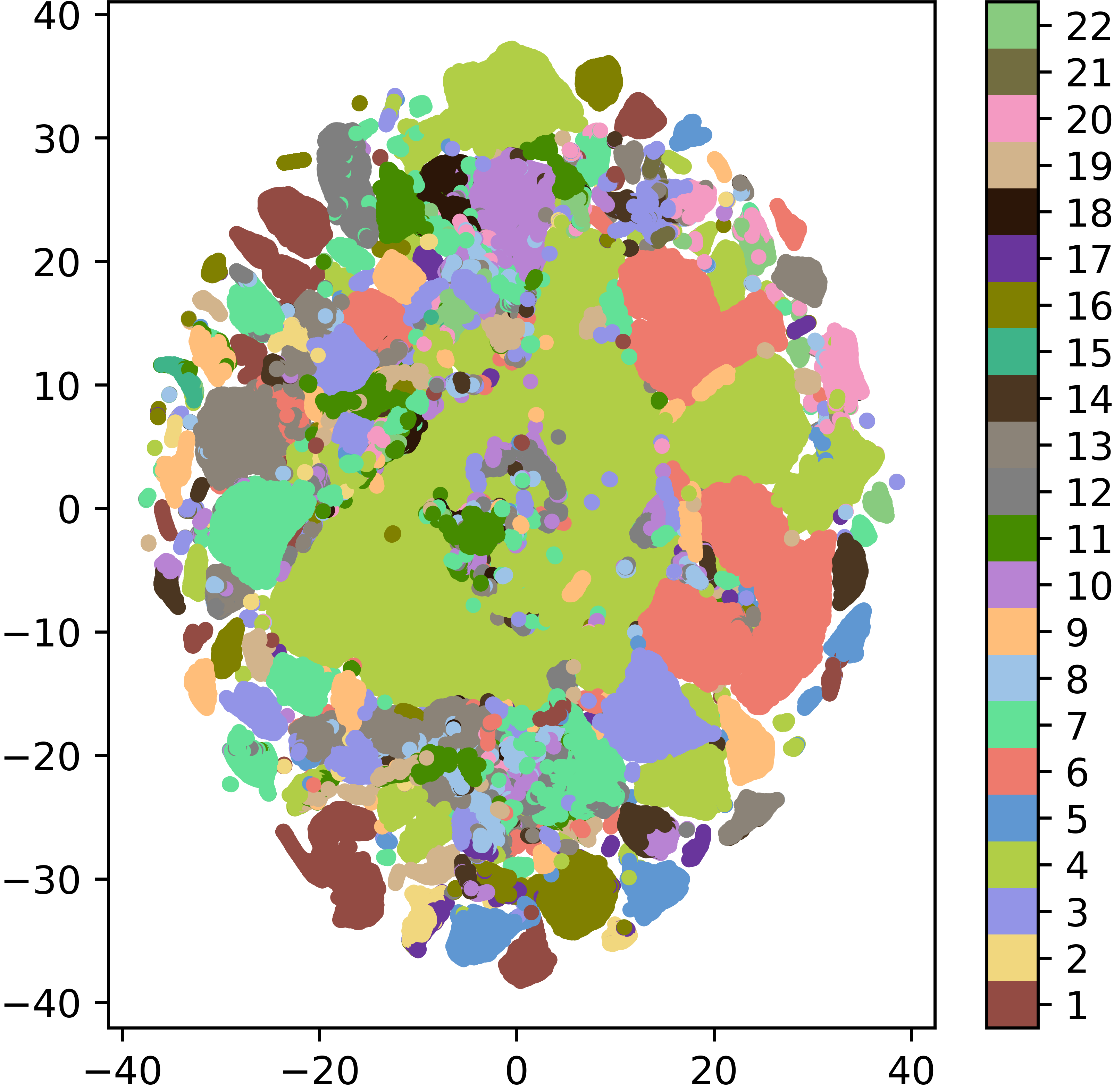}}
  {\includegraphics[width=0.15\textwidth]{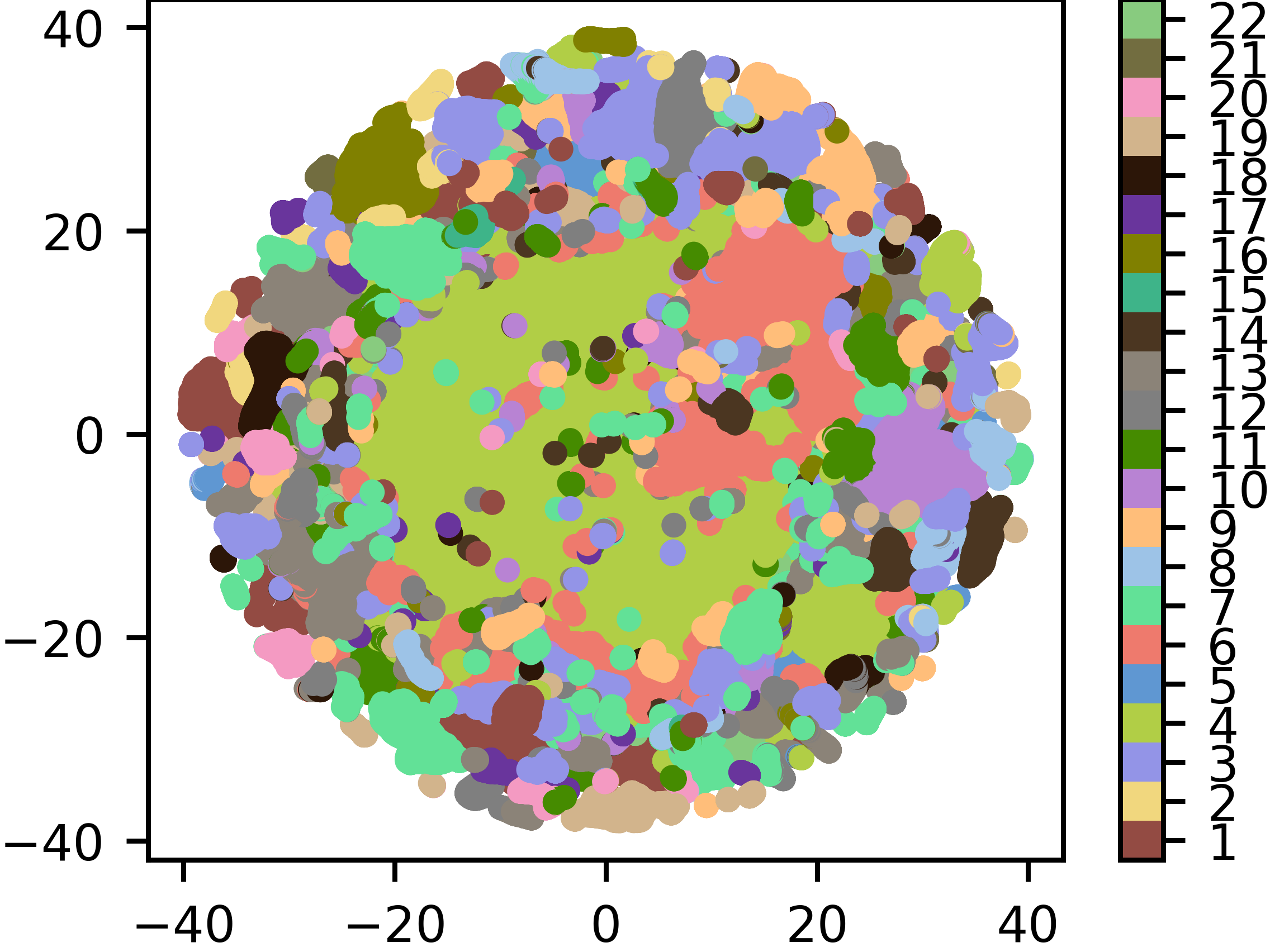}}
  {\includegraphics[width=0.15\textwidth]{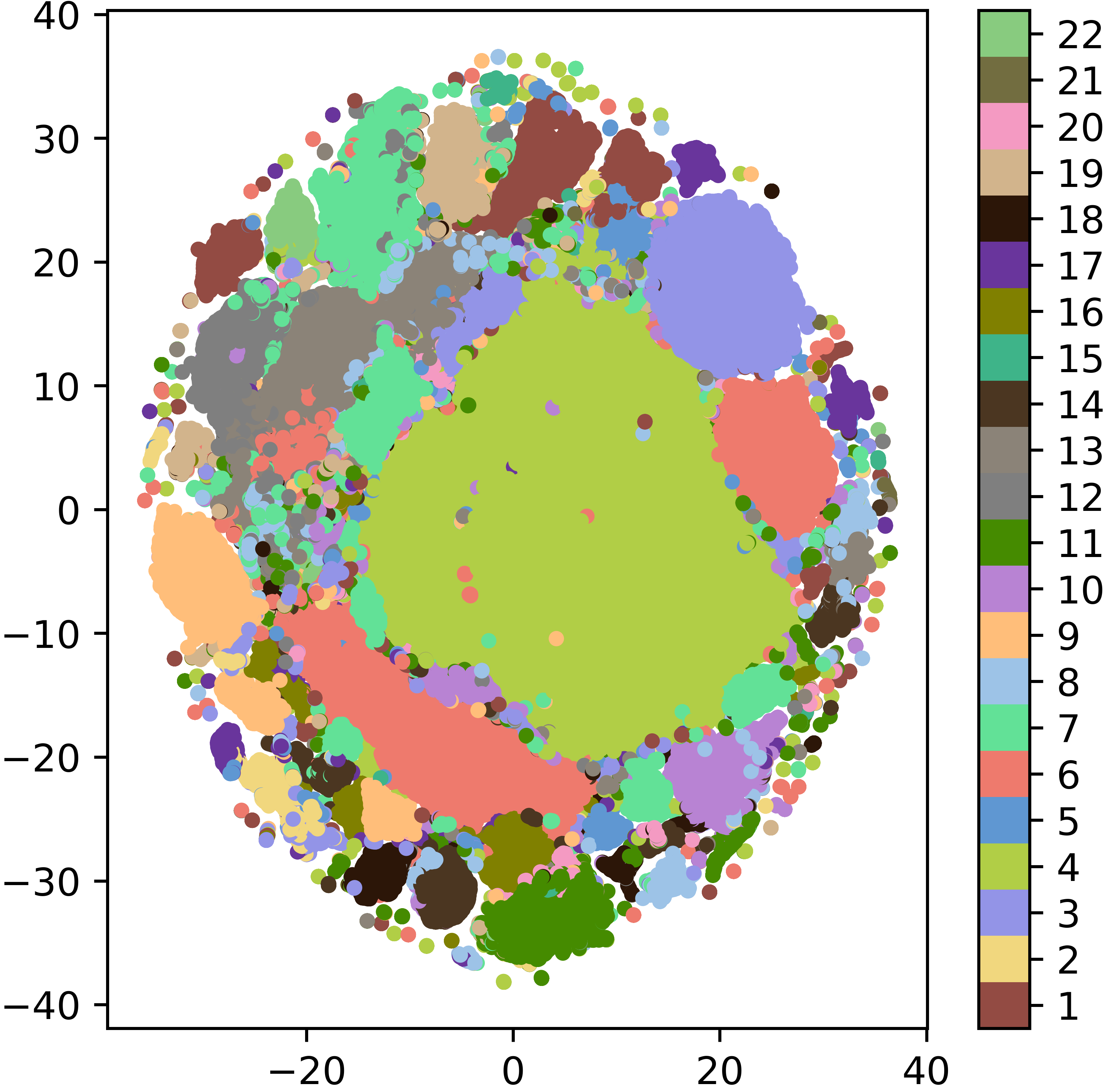}}
  {\includegraphics[width=0.15\textwidth]{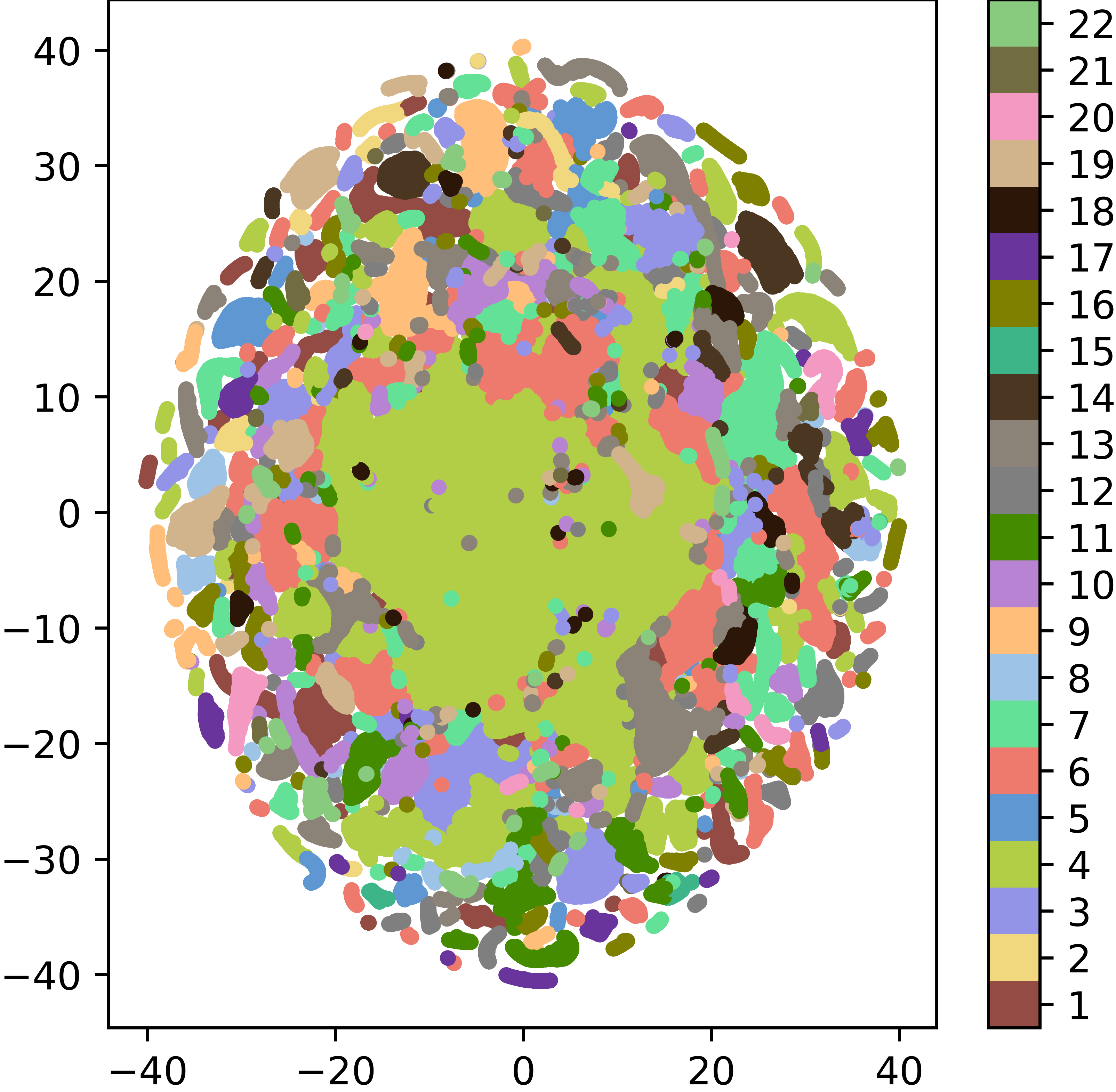}}
  {\includegraphics[width=0.15\textwidth]{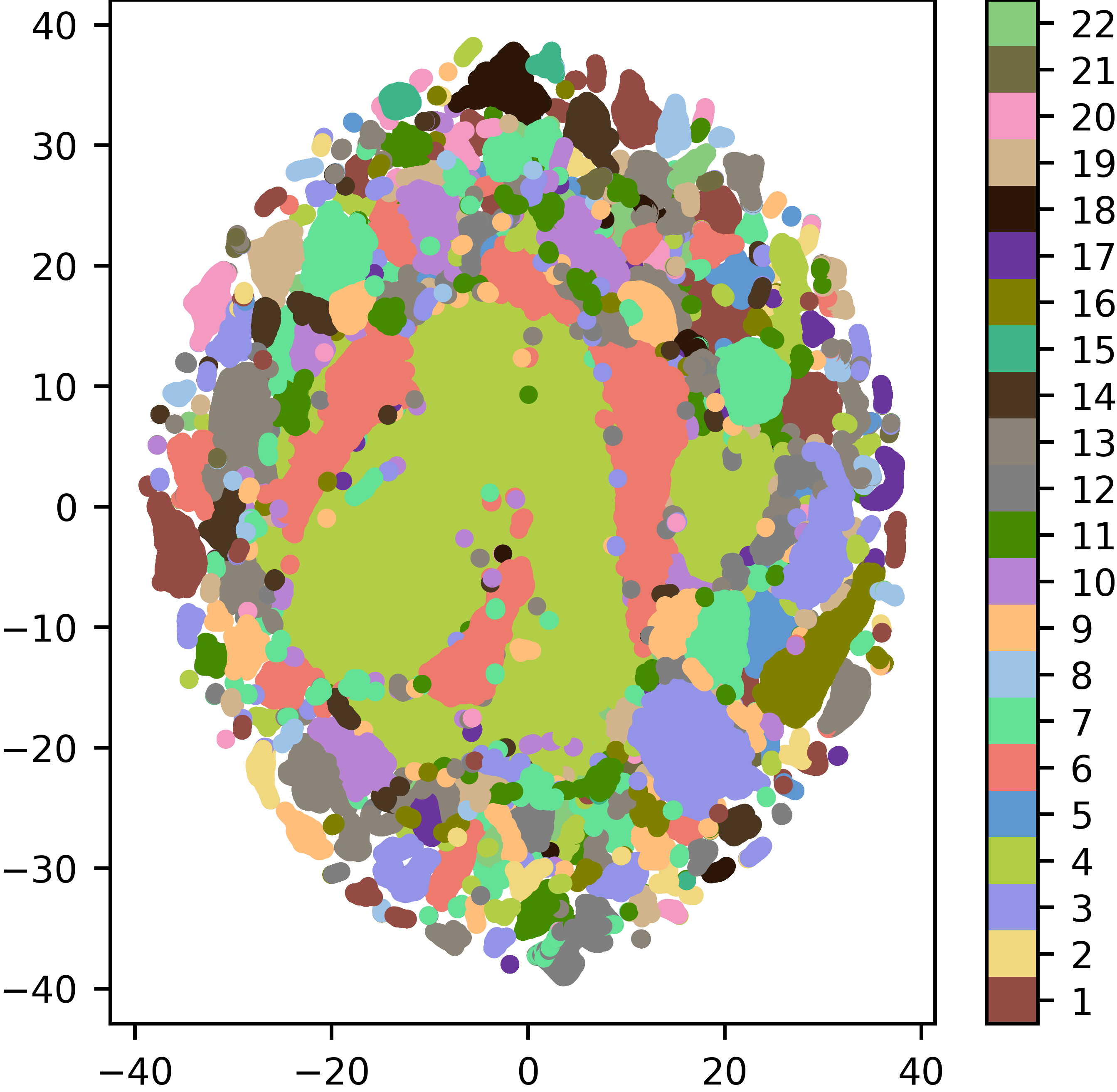}}
  }
  \vspace{3pt}
  \centerline{
  \subfigure[]{\includegraphics[width=0.15\textwidth]{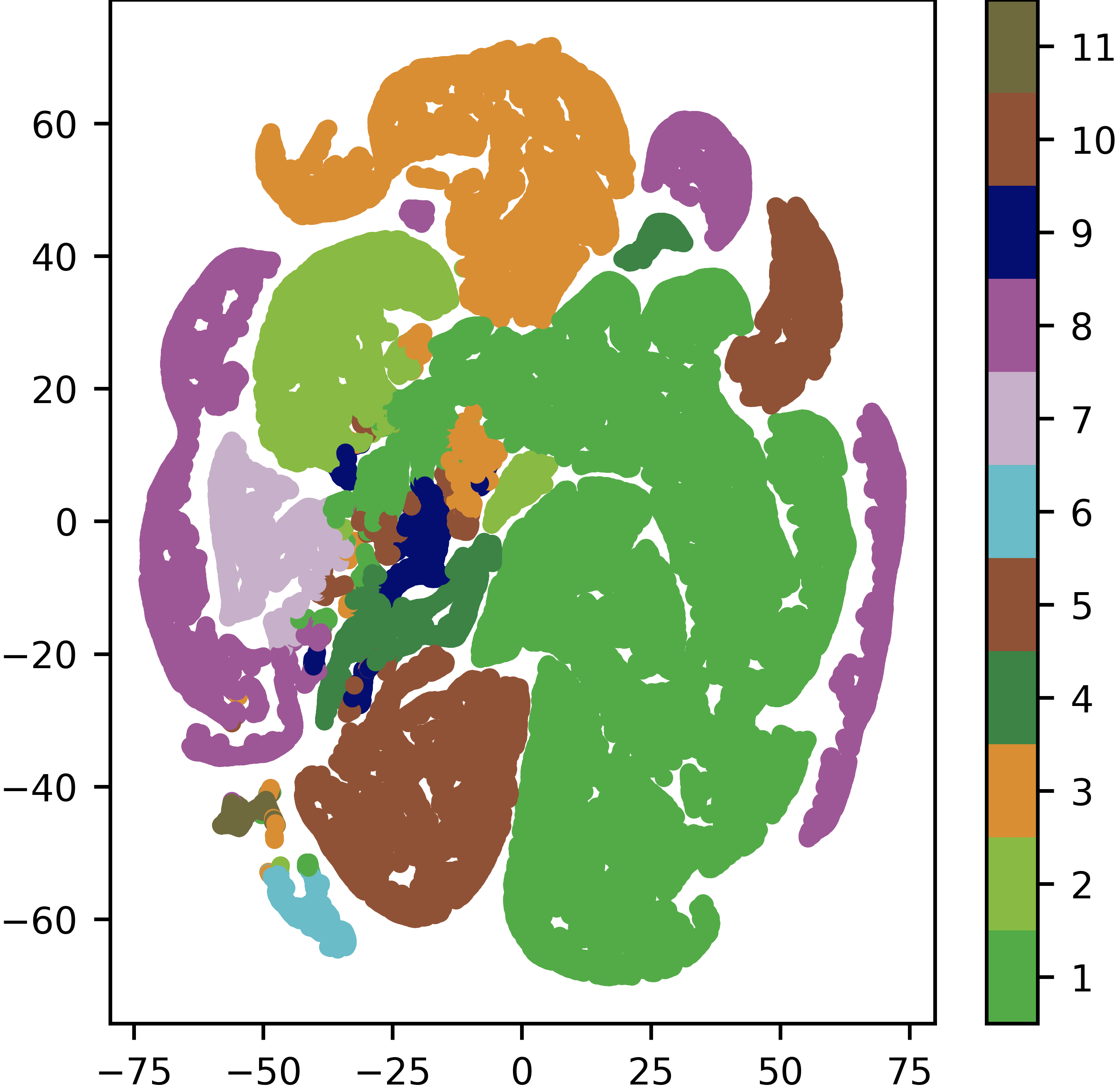}}
  \subfigure[]{\includegraphics[width=0.15\textwidth]{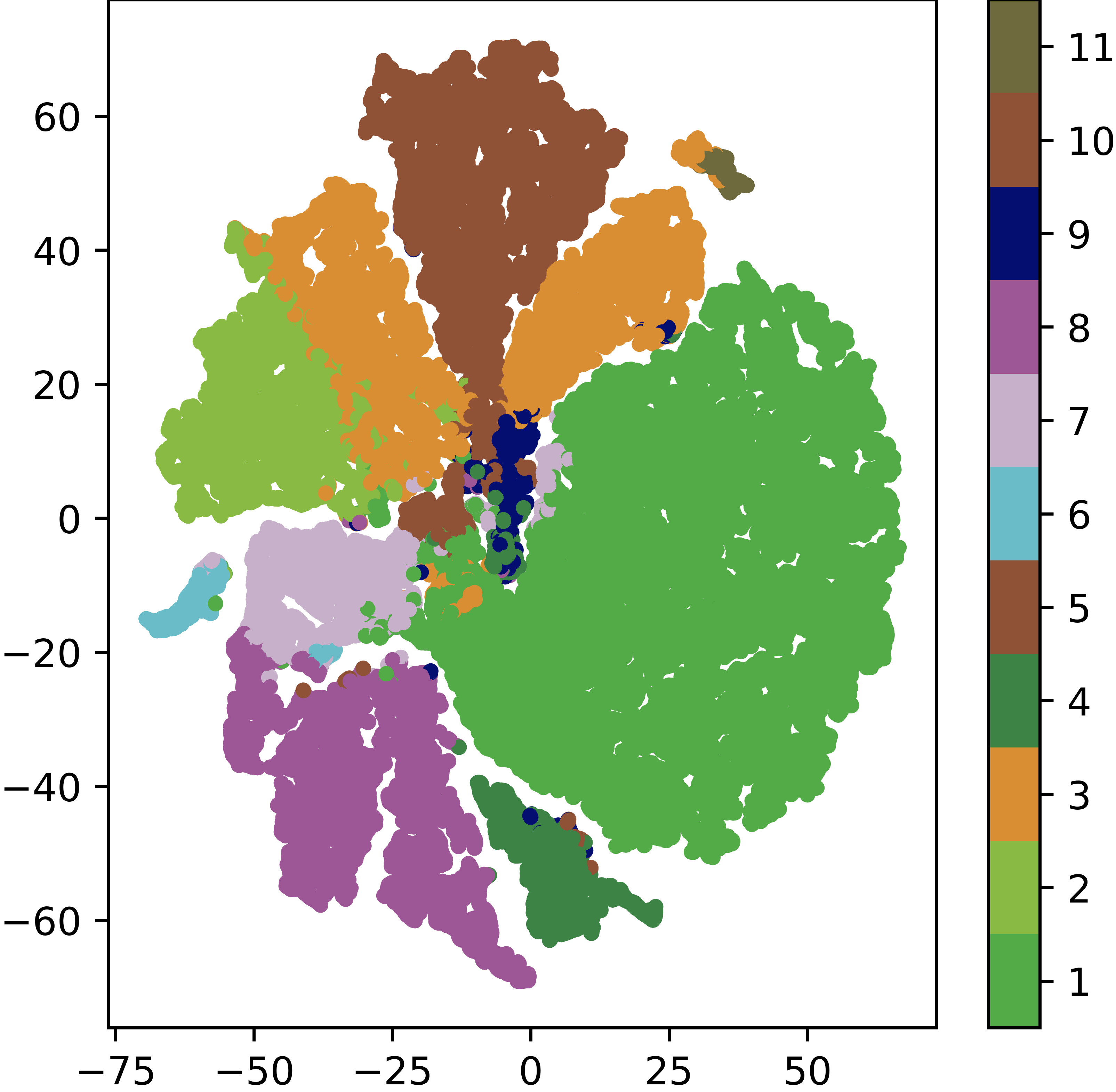}}
  \subfigure[]{\includegraphics[width=0.15\textwidth]{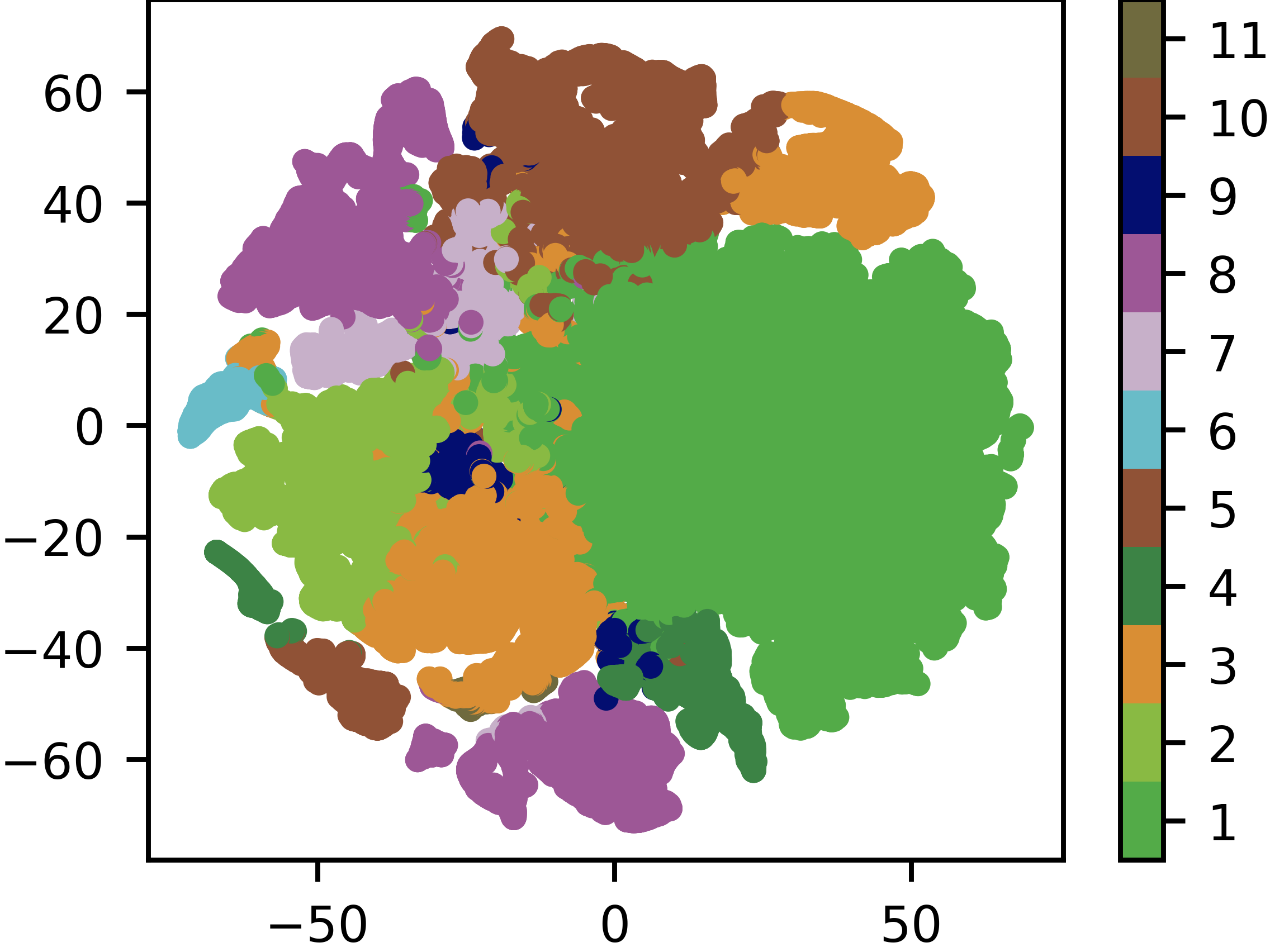}}
  \subfigure[]{\includegraphics[width=0.15\textwidth]{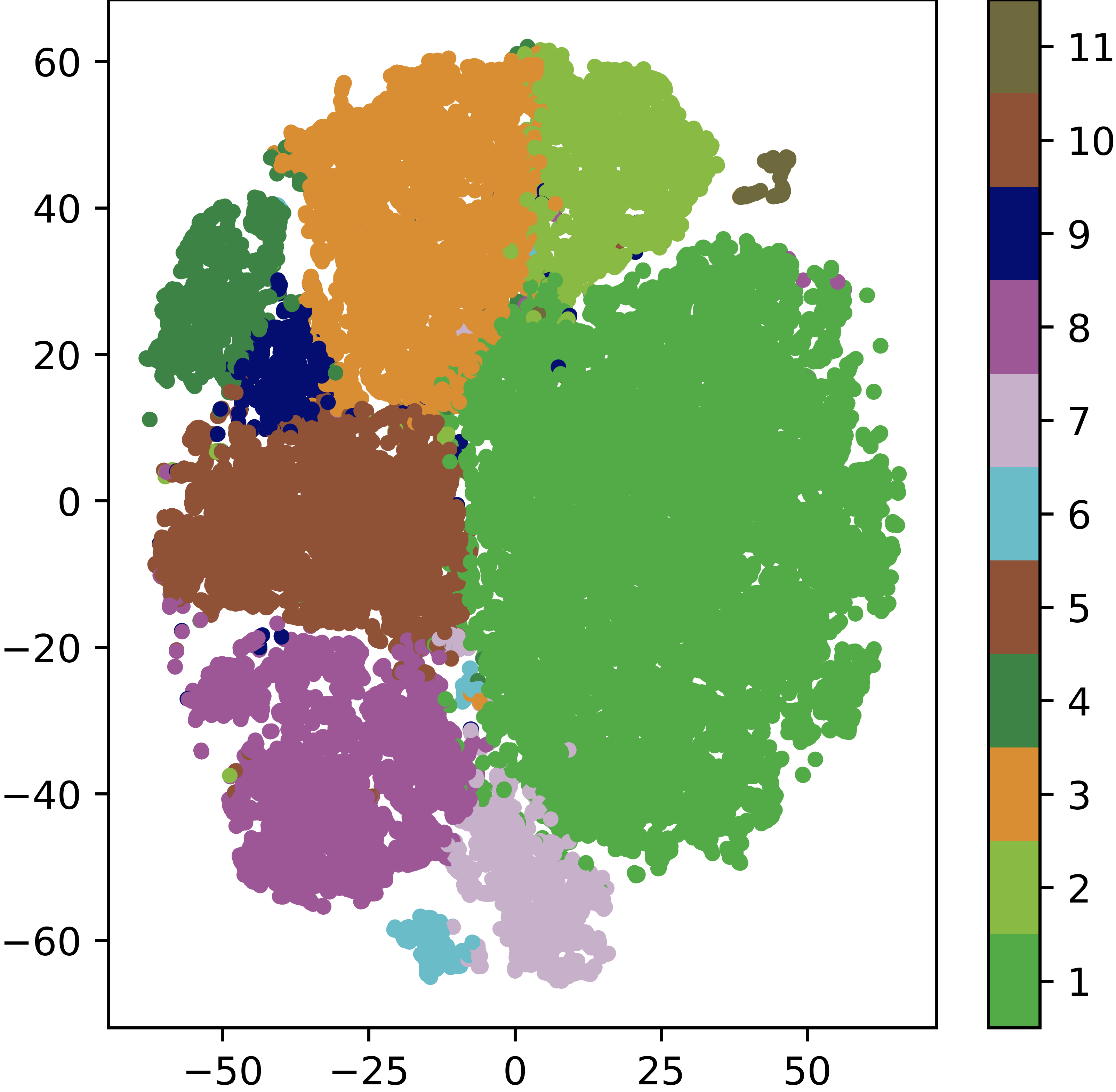}}
  \subfigure[]{\includegraphics[width=0.15\textwidth]{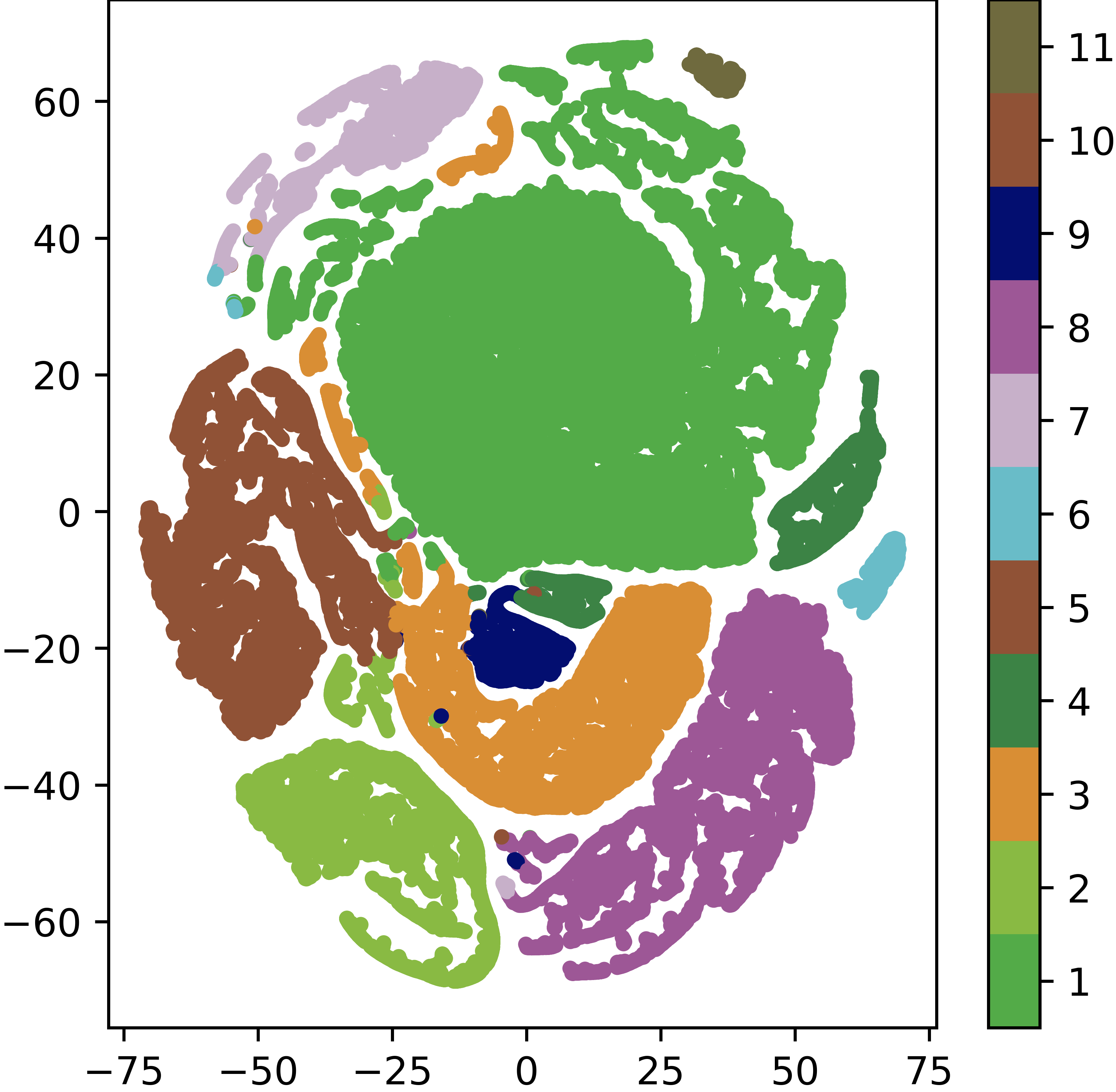}}
  \subfigure[]{\includegraphics[width=0.15\textwidth]{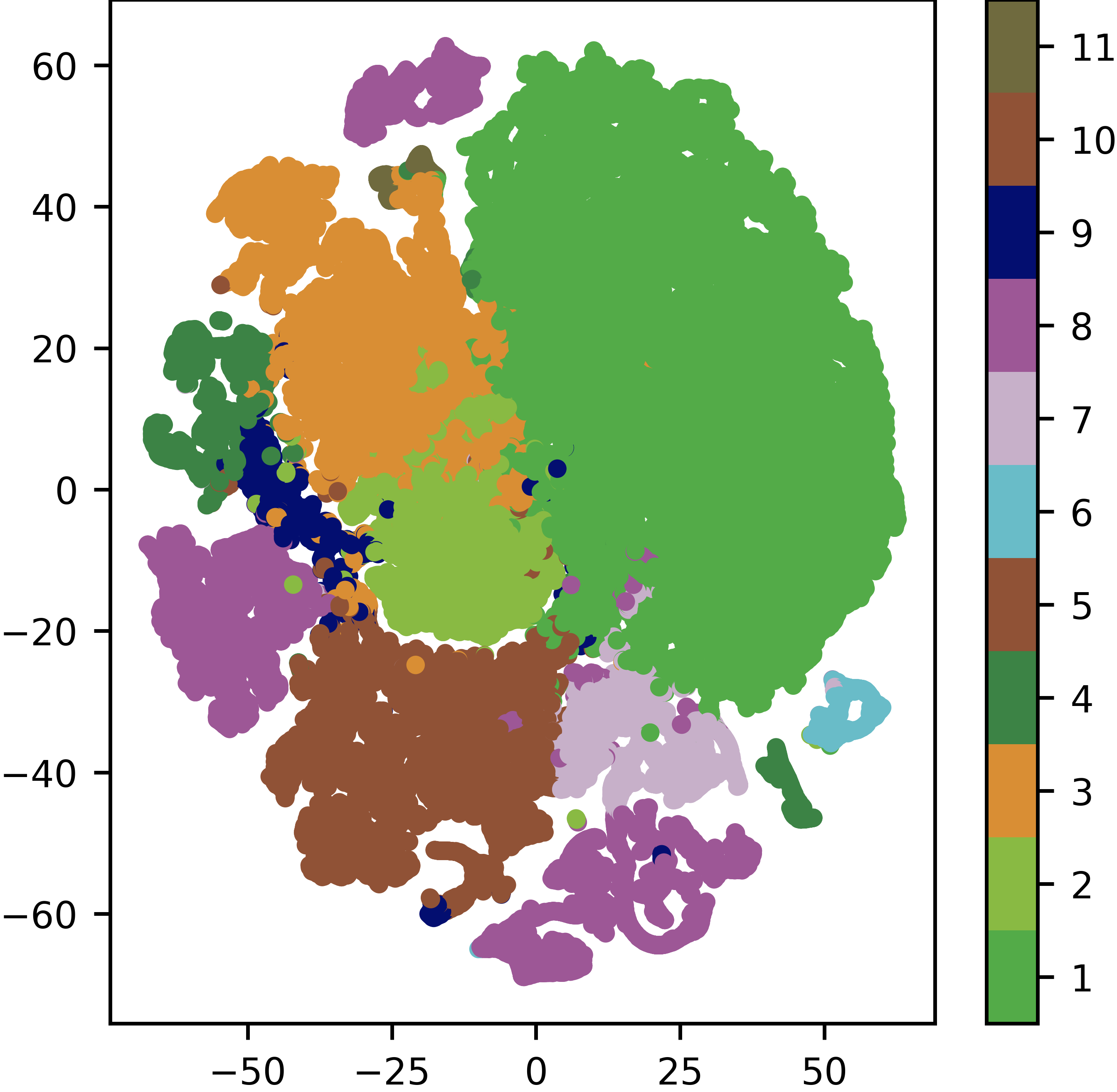}}
  }
  \caption{2-D graphical visualization of the features extracted by different methods and proposed DiffCRN through t-SNE on the four datasets (Indian Pines, Pavia University, WHU-Hi-HongHu, MUUFL, from top to bottom). (a) MTGAN. (b) SSRN. (c) SS-ConvNeXt. (d) SSFTT. (e) SC-SS-MTr. (f) DiffCRN. Results on Exp1.}
\label{TSNE_map_EXP1}
\end{figure*}

Futrhermore, 2-D graphical visulization map depicting the features extracted by different methods and proposed DiffCRN. \autoref{TSNE_map_EXP1} (a)-(f) for MTGAN, SSRN, SS-ConvNeXt, SSFTT, SC-SS-MTr and DiffCRN on four datasets respectively. Using the t-SNE, we can clearly see that test data of similar categories gather together, and intraclass variance is minimized on our proposed method DiffCRN.



\subsection{Performance Over Limited Training Sample Size}

To further shows the classification performance of proposed DiffCRN, we conduct Exp2 on four datasets. Twenty samples are selected for each class as listed in \autoref{tabel_sample_num}. The results are shown in \autoref{exp_20_perclass}. As a matter of course, even with a limited number of training samples, the proposed DiffCRN exhibits significantly better classification performance than the other methods, especially significant enhancement in terms of MIoU. Notably, the classification performance of the Indian Pines Dataset is largely improved compared with the previous SOTA pre-trained method based on masked image modeling, e.g., SC-SS-MTr, in terms of OA (92.03\% versus 90.04\%), AA (96.25\% versus 94.41\%), and $k$ (91.26\% versus 88.69\%), FWIoU (86.23\% versus 83.04\%), MIoU (85.99\% versus 78.66\%).

\begin{table}[htbp]
\centering
\Huge
\caption{Quantitative performance of different classification methods in terms of OA, AA, $k$, FWIoU, MIoU on the four datasets. The best results are in bold and colored shadow, Suboptimal are in underlined. Results on Exp2.}
\resizebox{0.49\textwidth}{!}{
\begin{tabular}{c|ccccc}
\hline
\multirow{2}{*}{\diagbox{Metrics}{Dataset}}
 &
  \multicolumn{5}{c}{Indian Pines} \\ \cline{2-6} 
 &
  \multicolumn{1}{c}{MTGAN \cite{Hangrl2020_TGRS}} &
  \multicolumn{1}{c}{SSRN \cite{SSRN}} &
  \multicolumn{1}{c}{SSFTT \cite{SSFTT}} &
  \multicolumn{1}{c}{SC-SS-MTr \cite{Huang2023_TGRS}} &
  \multicolumn{1}{c}{\textbf{DiffCRN}} \\ \hline
OA (\%) $\uparrow$&
   88.60±2.60&
   89.59±1.23&
   83.99±3.03&
   \underline{90.04±1.96}&
   \cellcolor[RGB]{251, 228, 213}\textbf{92.03±1.12}\\
AA (\%) $\uparrow$&
   92.90±1.19&
   \underline{94.78±2.03}&
   84.53±3.85&
   94.41±0.91&
   \cellcolor[RGB]{251, 228, 213}\textbf{96.25±0.66}\\
$k$*100 (\%) $\uparrow$&
   87.46±2.78&
   88.57±1.33&
   82.35±3.26&
   \underline{88.69±2.20}&
   \cellcolor[RGB]{251, 228, 213}\textbf{91.26±1.23}\\
FWIoU (\%) $\uparrow$&
   80.92±3.68&
   82.33±1.75&
   74.10±3.82&
   \underline{83.04±3.25}&
   \cellcolor[RGB]{251, 228, 213}\textbf{86.23±1.76}\\
MIoU (\%) $\uparrow$&
   77.39±3.37&
   \underline{81.39±2.24}&
   74.41±2.48&
   78.66±3.64&
   \cellcolor[RGB]{251, 228, 213}\textbf{85.99±2.91}\\ \hline
\multirow{2}{*}{\diagbox{Metrics}{Dataset}} &
  \multicolumn{5}{c}{Pavia University} \\ \cline{2-6} 
 &
  \multicolumn{1}{c}{MTGAN} &
  \multicolumn{1}{c}{SSRN} &
  \multicolumn{1}{c}{SSFTT} &
  \multicolumn{1}{c}{SC-SS-MTr} &
  \multicolumn{1}{c}{\textbf{DiffCRN}} \\ \hline
OA (\%) $\uparrow$&
   \cellcolor[RGB]{251, 228, 213}\textbf{96.66±0.94}&
   91.88±1.24&
   87.81±2.16&
   95.86±2.33&
   \underline{95.97±0.82}\\
AA (\%) $\uparrow$&
   96.02±1.26&
   93.60±3.41&
   92.25±2.70&
   \underline{96.72±1.33}&
   \cellcolor[RGB]{251, 228, 213}\textbf{97.06±0.3}6\\
$k$*100 (\%) $\uparrow$&
   \cellcolor[RGB]{251, 228, 213}\textbf{95.60±1.24}&
   89.42±1.64&
   84.44±2.68&
   94.59±3.01&
   \underline{94.73±1.04}\\
FWIoU (\%) $\uparrow$&
   \cellcolor[RGB]{251, 228, 213}\textbf{93.82±1.60}&
   85.74±2.12&
   79.02±3.15&
   \underline{92.66±3.99}&
   92.53±1.40\\
MIoU (\%) $\uparrow$&
   91.15±2.41&
   85.42±1.30&
   79.04±3.64&
   \underline{91.19±4.11}&
   \cellcolor[RGB]{251, 228, 213}\textbf{92.33±0.81}\\ \hline
\multirow{2}{*}{\diagbox{Metrics}{Dataset}} &
  \multicolumn{5}{c}{WHU-Hi-HongHu} \\ \cline{2-6} 
 &
  \multicolumn{1}{c}{MTGAN} &
  \multicolumn{1}{c}{SSRN} &
  \multicolumn{1}{c}{SSFTT} &
  \multicolumn{1}{c}{SC-SS-MTr} &
  \multicolumn{1}{c}{\textbf{DiffCRN}} \\ \hline
OA (\%) $\uparrow$&
   \underline{92.11±1.87}&
   86.48±1.61&
   75.94±3.93&
   88.14±2.03&
   \cellcolor[RGB]{251, 228, 213}\textbf{93.19±0.56}\\
AA (\%) $\uparrow$&
   \underline{91.23±1.79}&
   85.88±2.34&
   76.02±1.69&
   90.02±1.23&
   \cellcolor[RGB]{251, 228, 213}\textbf{92.67±0.55}\\
$k$*100 (\%) $\uparrow$&
   \underline{90.10±1.10}&
   83.28±1.92&
   70.95±4.23&
   87.36±1.40&
   \cellcolor[RGB]{251, 228, 213}\textbf{91.47±0.67}\\
FWIoU (\%) $\uparrow$ &
   \underline{86.94±1.32}&
   76.54±2.25&
   66.97±4.19&
   83.90±3.53&
   \cellcolor[RGB]{251, 228, 213}\textbf{88.64±0.76}\\
MIoU (\%) $\uparrow$ &
   \underline{75.57±3.10}&
   66.97±2.52&
   50.13±2.76&
   70.36±3.47&
   \cellcolor[RGB]{251, 228, 213}\textbf{78.57±1.02}\\ \hline
\multirow{2}{*}{\diagbox{Metrics}{Dataset}} &
  \multicolumn{5}{c}{MUUFL} \\ \cline{2-6} 
 &
  \multicolumn{1}{c}{MTGAN} &
  \multicolumn{1}{c}{SSRN} &
  \multicolumn{1}{c}{SSFTT} &
  \multicolumn{1}{c}{SC-SS-MTr} &
  \multicolumn{1}{c}{\textbf{DiffCRN}} \\ \hline
OA (\%) $\uparrow$ &
   81.55±1.52&
   \underline{81.76±2.97}&
   74.51±1.51&
   81.14±2.63&
   \cellcolor[RGB]{251, 228, 213}\textbf{83.75±1.53}\\
AA (\%) $\uparrow$ &
   76.14±3.84&
   \underline{82.38±5.20}&
   75.82±4.48&
   81.80±1.92&
   \cellcolor[RGB]{251, 228, 213}\textbf{84.12±1.28}\\
$k$*100 (\%) $\uparrow$ &
   76.05±1.93&
   76.80±3.45&
   67.82±1.76&
   76.03±3.29&
   \cellcolor[RGB]{251, 228, 213}\textbf{79.21±1.79}\\
FWIoU (\%) $\uparrow$ &
   71.68±2.13&
   \underline{72.71±3.70}&
   63.80±1.86&
   72.23±3.14&
   \cellcolor[RGB]{251, 228, 213}\textbf{75.46±1.91}\\
MIoU (\%) $\uparrow$ &
   54.55±3.24&
   \underline{59.04±2.76}&
   47.04±1.51&
   54.40±3.98&
   \cellcolor[RGB]{251, 228, 213}\textbf{59.19±1.67}\\ \hline
\end{tabular}
}
\label{exp_20_perclass}
\end{table}

\subsection{Model Analysis}

\subsubsection{\textbf{Ablation study}}
In this section, we analyze the effect of the key components and objective function in our method.
\subsubsubsection{\textbf{with or without spatial-spectral diffusion model ?}}
To demonstrate the effectiveness of DiffCRN, we designed the following three comparative experiments:

\textit{Case 1:} We only feed the raw all-band hyperspectral instance to the classification backbone without spectral-spatial diffusion contrastive representation learning, AWAM and CTSSFM. Noted as \textit{Baseline} in \autoref{raw_diffusion_feature_table}.

\textit{Case 2:} We feed the raw all-band hyperspectral instance to the proposed classifier. What needs illustration is that the clean all-band hyperspectral instance are processed by the AWAM and CTSSFM. Noted as \textit{Raw all-band} in \autoref{raw_diffusion_feature_table}.

\textit{Case 3:} We feed the features extracted by the pre-trained DiffCRN to the proposed classifier. Noted as \textit{Diffusion features} in \autoref{raw_diffusion_feature_table}.

As shown in \autoref{raw_diffusion_feature_table}, the classification results on four datasets are presented. The results demonstrate that using diffusion features as input significantly outperforms the use of raw features and baseline methods. For example, the improvement in terms of OA, with an increase of approximately 10.20\% on the Indian Pines dataset, 9.02\% on Pavia University dataset, 8.97\% on WHU-Hi-HongHu and 4.45\% on MUUFL. Furthermore, the more obvious improvement is in MIoU, with an increase of approximately 17.51\%, 17.68\%, 28.56\% and 13.76\% on these datasets, which indicates the extracted features have stronger discrimination resulting in lower misclassification ability and the effectiveness of using spectral-spatial diffusion contrastive representation learning.

\begin{table}[htbp]
\Huge
\centering
\caption{Classification performance with different input on four datasets. Best results are in bold and colored shadow.}
\resizebox{0.45\textwidth}{!}{
\begin{tabular}{c|ccccc}
\hline
\multirow{2}{*}{\diagbox{Case}{Metrics}} & \multicolumn{5}{c}{Indian Pines}     \\ \cline{2-6} 
                  & OA   & AA   & Kappa  & FWIoU  & MIoU \\ \hline
\textit{Baseline} & 89.13±2.06     & 86.16±2.68     & 88.68±2.21       & 82.31±3.14       & 81.09±2.78     \\                  
\textit{Raw all-band} & 96.07±0.99     & 94.54±2.43     & 95.85±1.11       & 93.08±1.77       & 92.39±2.65     \\
\textit{Diffusion feature}       & \cellcolor[RGB]{251, 228, 213}\textbf{99.33±0.16}     & \cellcolor[RGB]{251, 228, 213}\textbf{99.26±0.37}     & \cellcolor[RGB]{251, 228, 213}\textbf{99.31±0.16}       & \cellcolor[RGB]{251, 228, 213}\textbf{98.81±0.28}       & \cellcolor[RGB]{251, 228, 213}\textbf{98.60±0.55}     \\ \hline
\multirow{2}{*}{\diagbox{Case}{Metrics}} & \multicolumn{5}{c}{Pavia University} \\ \cline{2-6} 
                  & OA   & AA   & Kappa  & FWIoU  & MIoU \\ \hline
\textit{Baseline} & 90.68±1.58      & 91.15±1.78     & 88.01±2.07       & 83.70±2.49       & 81.32±3.03     \\  
\textit{Raw all-band} &  94.76±0.58    & 95.36±1.10     & 93.21±0.74       & 90.71±0.99       & 86.86±1.33     \\
\textit{Diffusion feature}       & \cellcolor[RGB]{251, 228, 213}\textbf{99.70±0.10}     & \cellcolor[RGB]{251, 228, 213}\textbf{99.41±0.21}     & \cellcolor[RGB]{251, 228, 213}\textbf{99.61±0.13}       & \cellcolor[RGB]{251, 228, 213}\textbf{99.42±0.19}       & \cellcolor[RGB]{251, 228, 213}\textbf{99.00±0.33}     \\ \hline
\multirow{2}{*}{\diagbox{Case}{Metrics}} & \multicolumn{5}{c}{WHU-Hi-HongHu}    \\ \cline{2-6} 
                  & OA   & AA   & Kappa  & FWIoU  & MIoU \\ \hline
\textit{Baseline} & 88.71±0.65     & 72.20±2.12     & 85.78±0.83       & 81.05±0.94      & 61.63±2.02     \\  
\textit{Raw all-band} & 95.56±0.57     & 88.53±1.52     & 94.40±0.72       & 91.95±0.91       & 82.42±1.91     \\
\textit{Diffusion feature}       &  \cellcolor[RGB]{251, 228, 213}\textbf{97.68±0.16}    & \cellcolor[RGB]{251, 228, 213}\textbf{94.39±0.39}     & \cellcolor[RGB]{251, 228, 213}\textbf{97.08±0.20}       & \cellcolor[RGB]{251, 228, 213}\textbf{95.60±0.28}       & \cellcolor[RGB]{251, 228, 213}\textbf{90.19±0.68}     \\ \hline
\multirow{2}{*}{\diagbox{Case}{Metrics}} & \multicolumn{5}{c}{MUUFL}            \\ \cline{2-6} 
                  & OA   & AA   & Kappa  & FWIoU  & MIoU \\ \hline
\textit{Baseline} & 88.99±0.75     & 73.02±2.39     & 85.85±0.95       & 81.76±1.07       & 62.50±2.71     \\  
\textit{Raw all-band} &     90.38±0.63 & 79.61±2.47     & 88.19±0.83       &  84.50±0.94      & 69.53±2.33     \\
\textit{Diffusion feature}       & \cellcolor[RGB]{251, 228, 213}\textbf{93.44±0.28}     & \cellcolor[RGB]{251, 228, 213}\textbf{82.77±1.08}     & \cellcolor[RGB]{251, 228, 213}\textbf{91.56±0.36}       & \cellcolor[RGB]{251, 228, 213}\textbf{88.44±0.44}       & \cellcolor[RGB]{251, 228, 213}\textbf{76.26±1.06}     \\ \hline
\end{tabular}
}
\label{raw_diffusion_feature_table}
\end{table}

\subsubsubsection{\textbf{with or without AWAM and CTSSFM ?}}

In DiffCRN, we use AWAM and CTSSFM to fuse information adaptively from identical and cross diffusion time step $t$, achieving more detailed classification map and obtaining higher-accuracy category prediction.

To verify the effectiveness of AWAM and CTSSFM, an ablation experiment is designed to ascertain the effectiveness. In detail, four comparisons are listed below: (1) only linear classifier; (2) AWAM with linear classifier; (3) CTSSFM with linear classifier; (4) AWAM and CTSSFM with linear classifier. \autoref{AWAM_CTSSFM_ablation} shows the relevant experimental results on the Indian Pines dataset. Experimental results shows that w/o AWAM\&CTSSFM performaces worse, possibly due to the loss of multi-timestep representations fusion. What’s more, the use of AWAM or CTSSFM can significantly improve the model performance. Furthermore, the combined utilization of AWAM and CTSSFM can further improve. More specifically, adding the AWAM and CTSSFM improves the OA, AA, $k$, FWIou and MIoU scores on the
Indian Pines by 2.39\%, 3.37\%, 2.48\%, 4.18\% and 5.23\%.

\begin{table}[htbp]
\centering
\Huge
\caption{Ablation study for the AWAM and CTSSAM on the Indian Pines dataset. The Best result are in bold and colored shadow.}
\resizebox{0.49\textwidth}{!}{
\begin{threeparttable}
\begin{tabular}{c|ccccc}
\hline
Metric          & OA & AA & Kappa & FWIoU & MIoU \\ \hline
w/o AWAM\&CTSSFM     & 96.94±0.36   & 95.89±0.77   & 96.83±0.38      & 94.63±0.60      & 93.37±1.71     \\
w/o CTSSAM      & 97.47±0.38   & 96.37±1.07   & 97.40±0.39      & 95.58±0.62      & 93.86±1.64      \\
w/o AWAM        & 99.15±0.25   & 99.0±1.23   & 99.13±0.26      & 98.50±0.45      & 98.22±1.13     \\
w/ AWAM\&CTSSAM & \cellcolor[RGB]{251, 228, 213}\textbf{99.33±0.16}    & \cellcolor[RGB]{251, 228, 213}\textbf{99.26±0.37}    & \cellcolor[RGB]{251, 228, 213}\textbf{99.31±0.16}    & \cellcolor[RGB]{251, 228, 213}\textbf{98.81±0.28}    & \cellcolor[RGB]{251, 228, 213}\textbf{98.60±0.55}      \\ \hline
\end{tabular}
\begin{tablenotes}
        \footnotesize
        \Huge
        \item[] w/: with, w/o: without
\end{tablenotes}
\end{threeparttable}
}
\label{AWAM_CTSSFM_ablation}
\end{table}

\subsubsubsection{\textbf{Compound Loss function or Single objective function of spatial-spectral diffusion model ?}}

To thoroughly evaluate the performance of the proposed method with the compound loss, we compare five cases of objective function of spatial-spectral diffusion model and evaluate the classification performance on Indian Pine dataset. These will be referred to as the following cases.

\textit{Case 1:} The MSE used as the objective function of spatial-spectral diffusion model, which is also the original and unmodified loss in DDPM \cite{ho2020DDPM} shown in \autoref{eq10}. For the sake of simplicity, as follows, 
\(\mathbfcal{L} = \mathbfcal{L}_{MSE}\)

\textit{Case 2:} The modified MSE loss function, named LAE, is used as objective function,as shown in \autoref{eq18}.  For the sake of simplicity, as follows, 
\(\mathbfcal{L} = \mathbfcal{L}_{diff}\)

\textit{Case 3:} Considering contrastive learning loss, the loss function can be writen as, \(\mathbfcal{L} = \mathbfcal{L}_{diff} + \mathbfcal{L}_{con}\)

\textit{Case 4:} Considering the uncertainty of optimization objective, and using contrastive learning loss, the loss function can be writen as, \(\mathbfcal{L} = e^{-s_{diff}} \mathbfcal{L}_{diff} + \mathbfcal{L}_{con} + s_{diff}\)

\textit{Case 5:} Considering the uncertainty of optimization objective, reconstruction loss in \autoref{eq_rec}, and using contrastive learning loss, the loss function can be writen as, \(\mathbfcal{L} = e^{-s_{diff}} \mathbfcal{L}_{diff} + e^{-s_{rec}} \mathbfcal{L}_{rec} + \mathbfcal{L}_{con} + s_{diff} + s_{rec}\)

\begin{figure}[htbp]
    \centering
    \includegraphics[width=0.49\textwidth]{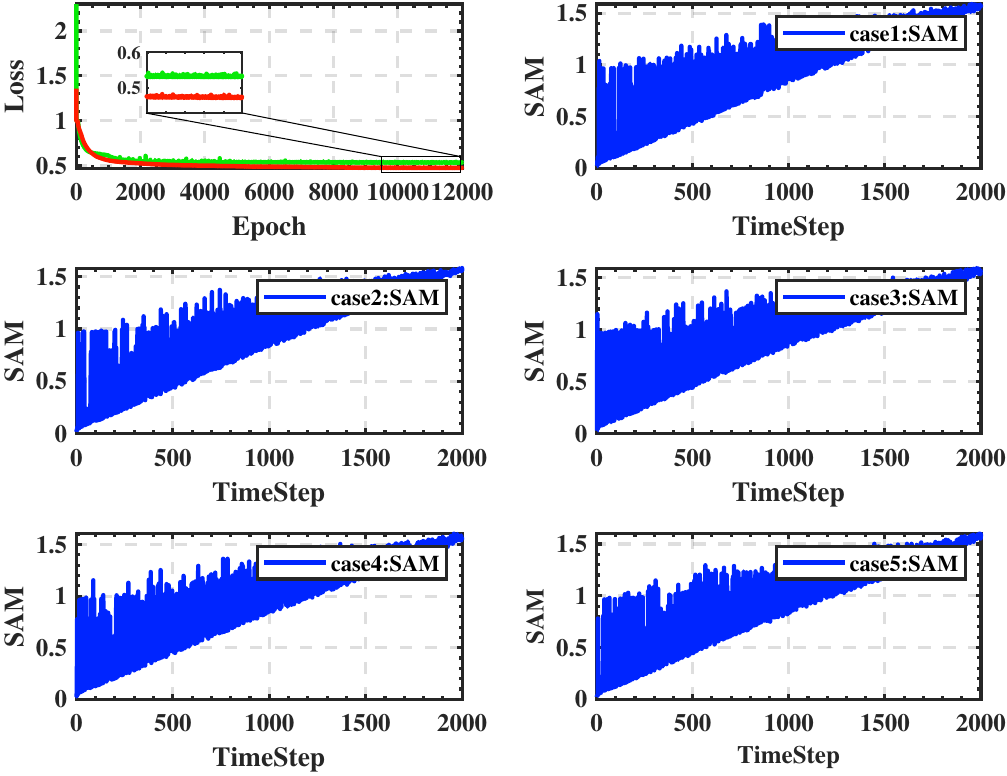}
    \caption{Top left corner: comparison between MSE loss and LAE loss on Indian Pines dataset, red line: LAE, blue line: MSE. The rest: learned spectral angle mapping (SAM) value under different cases on Indian Pines dataset. The smaller the SAM, the higher the similarity.}
    \label{loss_SAM_IN}
\end{figure}

The results are listed in \autoref{loss_function}, Comparing \textit{case2} with \textit{case1}, there has been a significant improvement in all metrics, e.g., 1\% improvement in OA. So, the modification of the raw objective function of diffusion model e.g., \autoref{eq18}, can better reconstruct the added gaussian noise and model the local-global spectral-spatial relationship of hyperspectral image by putting more emphasis on heavily penalizing the big gap between $\boldsymbol{\epsilon}$ and $\boldsymbol{\epsilon}_{\theta}(\mathbfcal{H}_{\boldsymbol{t}},t)$, as shown in top left corner of \autoref{loss_SAM_IN}. \textit{case2}, \textit{case3} and \textit{case4} have the similar results. In \textit{case 5}, by comprehensive considering of the uncertainty of optimization objective, reconstruction loss in \autoref{eq_rec}, and using contrastive learning loss, the metrics are still improving, especially in terms of AA, an increase of about 0.6\%. It means that more discriminative feature are obtained by optimizing the compound loss function. The diffusion time step $t$ are determined by the learned SAM as shown in \autoref{loss_SAM_IN}.

\begin{table}[htbp]
\Huge
\centering
\caption{Classification performance with different sets of objective function $t$ on the Indian Pines dataset.  The Best results are in bold and colored shadow.}
\resizebox{0.49\textwidth}{!}{
\begin{threeparttable}
\begin{tabular}{c|c|ccccc}
\hline
Objective function & Time Step            & OA         & AA         & Kappa      & FWIoU      & MIoU       \\ \hline
\textit{Case 1}    & {[}0, 1, 2, 4, 5{]}  & 98.13±0.22 & 97.69±0.69 & 98.08±0.23 & 96.72±0.38 & 95.57±0.63 \\
\textit{Case 2}    & {[}0, 1, 3, 2, 4{]}  & 99.25±0.16 & 98.65±0.54 & 99.23±0.16 & 98.68±0.28 & 97.91±0.40 \\
\textit{Case 3}    & {[}0, 1, 4, 7, 13{]} & 99.15±0.16 & 98.91±0.41 & 99.12±0.16 & 98.48±0.28 & 98.02±0.77 \\
\textit{Case 4}    & {[}1, 2, 5, 6, 10{]} & 99.20±0.22 & 98.82±0.82 & 99.18±0.22 & 98.58±0.38 & 97.94±0.87 \\
\textbf{\textit{Case 5}}    & {[}0, 1, 2, 3, 5{]}  & \cellcolor[RGB]{251, 228, 213}\textbf{93.33±0.16} & \cellcolor[RGB]{251, 228, 213}\textbf{99.25±0.40} & \cellcolor[RGB]{251, 228, 213}\textbf{99.31±0.16} & \cellcolor[RGB]{251, 228, 213}\textbf{98.81±0.28} & \cellcolor[RGB]{251, 228, 213}\textbf{98.60±0.55} \\ \hline
\end{tabular}
\begin{tablenotes}
        \footnotesize
        \Huge
        \item[1] Choosing time step $t$ according to the top 5 results of SAM value.
\end{tablenotes}
\end{threeparttable}
}
\label{loss_function}
\end{table}

\begin{table}[htbp]
\Huge
\centering
\caption{Classification performance with different sets of time step $t$ on the Indian Pines dataset with 10\% training samples. Best results are in bold and colored shadow..}
\resizebox{0.49\textwidth}{!}{
\begin{threeparttable}
\begin{tabular}{c|c|ccccc}
\hline
\multirow{2}{*}{Index} & \multirow{2}{*}{Time Step \tnote{1}}       & \multicolumn{5}{c}{Metric}     \\ \cline{3-7} 
                       &                                  & OA & AA & Kappa & FWIoU & MIoU \\ \hline
S1  &  {[}0, 1, 2, 3, 4{]}                          & 99.27±0.17     & 99.08±0.40    & 99.25±0.18    & 98.70±0.31    & 98.43±0.62   \\
S2  &  \makecell[c]{{[}0, 1, 2, 3, 4, 5, 6, 7, 8, 9{]}}                          & 99.18±0.26     & 99.03±0.5   & 99.16±0.26    & 98.54±0.45    & 98.29±0.64   \\
S3  &  \makecell[c]{{[}0, 1, 2, 3, 4, ... , 26, 27, 28, 29{]}}                          & 98.68±0.28     & 98.29±0.60   & 98.66±0.29    & 97.69±0.48    & 96.99±1.33   \\
S4  &  {[}50, 100, 200, 300, 500{]}                          & 98.80±0.24     & 98.91±0.51    & 98.77±0.25    & 97.87±0.42    & 97.85±0.75   \\
S5  &  \multicolumn{1}{c|}{{[}700, 900, 1100, 1300, 1500{]}} & 94.09±0.64    & 93.27±2.64    & 93.78±0.70    & 89.78±1.11    & 89.34±2.67   \\
S6  &  \multicolumn{1}{c|}{{[}50, 200, 500, 900, 1500{]}}    & 98.69±0.19    & 98.57±0.75    & 98.66±0.19    & 97.68±0.32    & 97.37±0.97   \\
S7  &  Top5 of SAM value                                     & \cellcolor[RGB]{251, 228, 213}\textbf{99.33±0.16}    & \cellcolor[RGB]{251, 228, 213}\textbf{99.26±0.37}    & \cellcolor[RGB]{251, 228, 213}\textbf{99.31±0.16}    & \cellcolor[RGB]{251, 228, 213}\textbf{98.81±0.28}    & \cellcolor[RGB]{251, 228, 213}\textbf{98.60±0.55}   \\
S8  &  Top10 of SAM value                                    & 99.23±0.11    & 98.99±0.39    & 99.21±0.11    & 98.62±0.19    & 98.26±0.48   \\
S9  &  Top30 of SAM value                                    & 99.18±0.11    & 98.83±0.74    & 99.16±0.11    & 98.55±0.19    & 98.22±0.80   \\
S10  &  Top100 of SAM value                                    & 98.45±0.67    & 98.36±1.88    & 98.42±1.16    & 97.28±1.16    & 97.21±0.74   \\ \hline
\end{tabular}
\begin{tablenotes}
        \footnotesize
        \Huge
        \item[1] {[}50, 100, 200, 300, 500{]}, {[}700, 900, 1100, 1300, 1500{]} and {{[}50, 200, 500, 900, 1500{]}}  means manually select time step $t$, while Top5, Top10, Top30 of SAM value mean adaptively select time step $t$ according to the top few value of SAM.
\end{tablenotes}

\end{threeparttable}
}
\label{ablation_timestep}
\end{table}

\begin{figure*}[htbp]
  \centerline{
    {\includegraphics[width=0.08\textwidth]{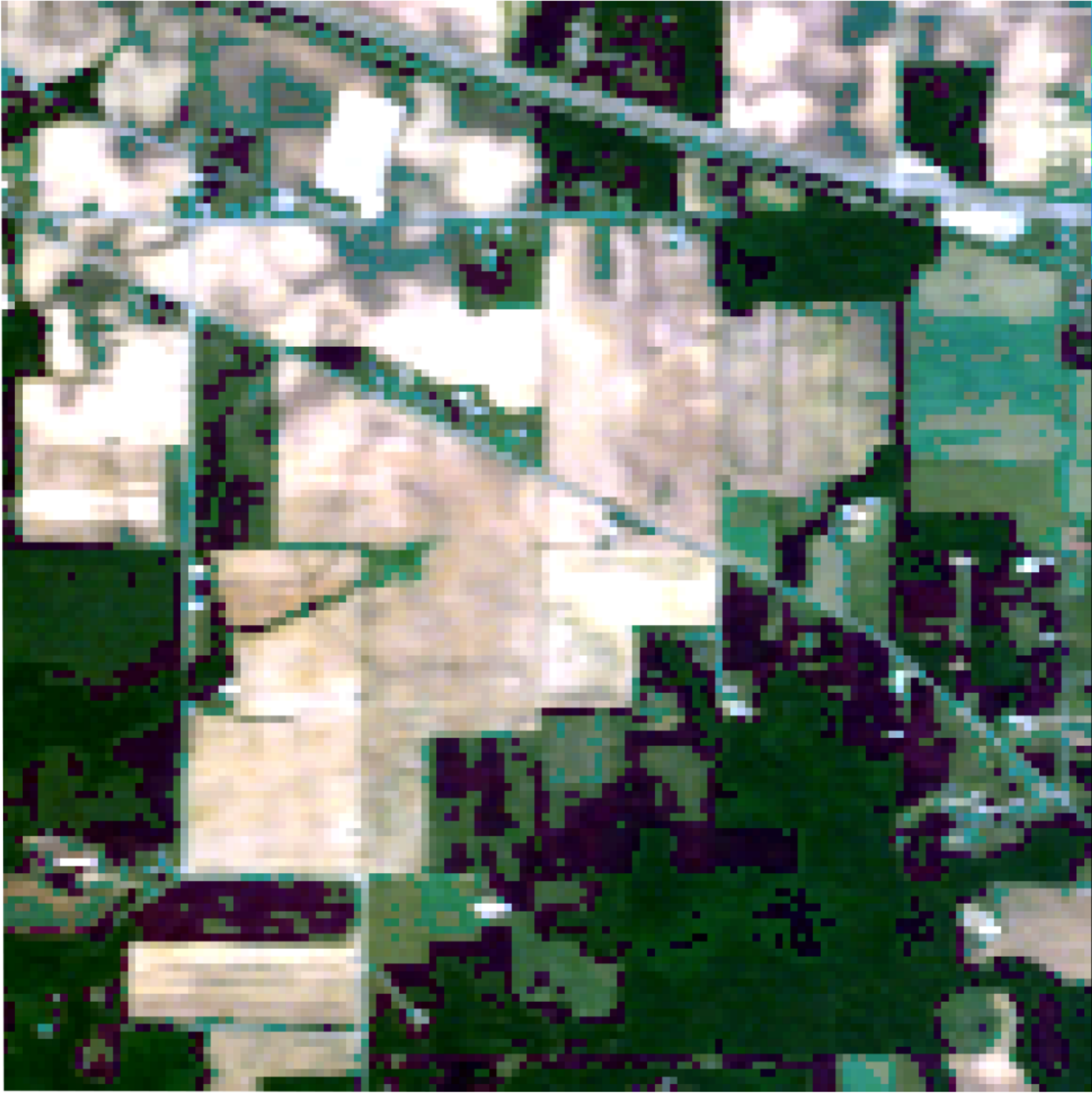}}
  {\includegraphics[width=0.08\textwidth]{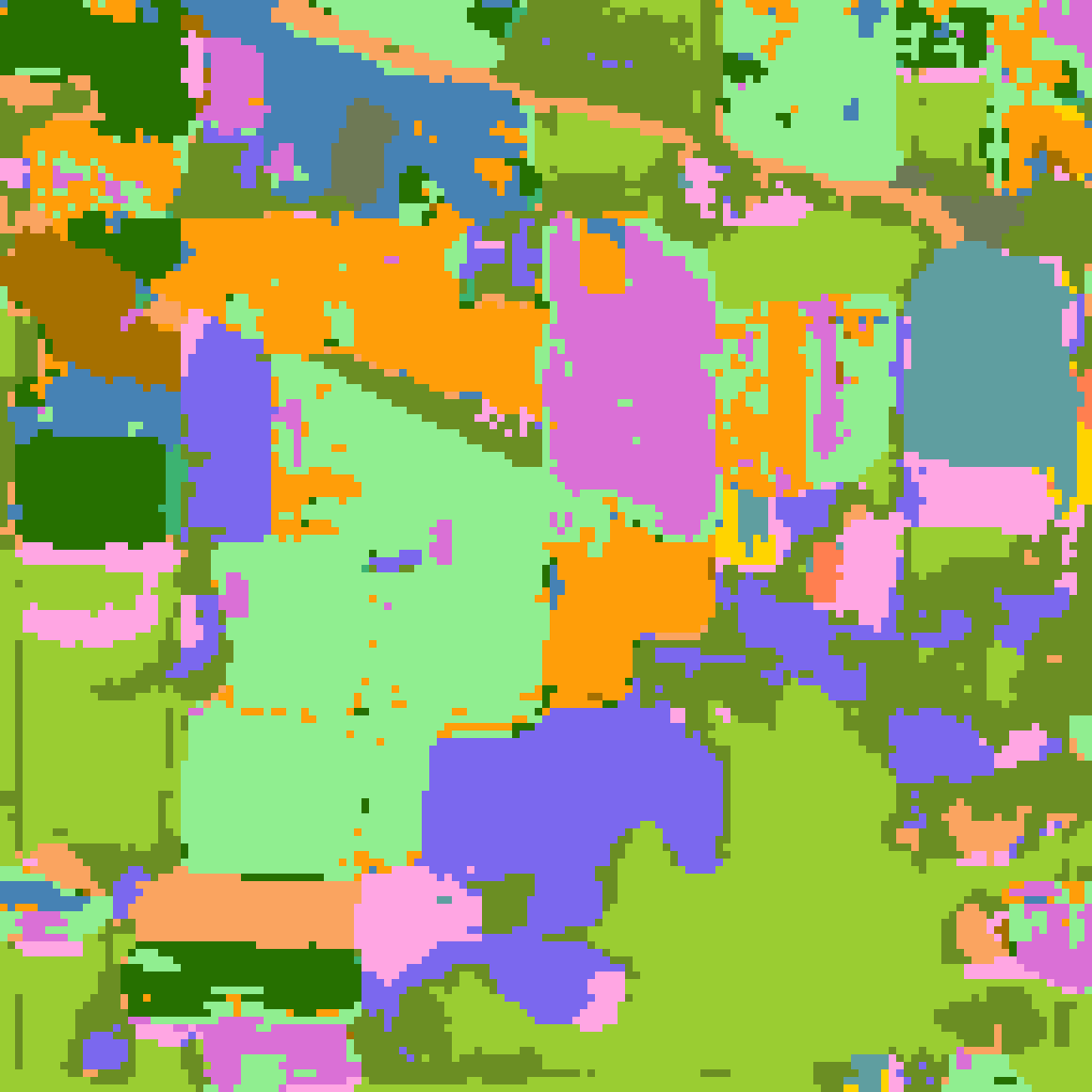}}
  {\includegraphics[width=0.08\textwidth]{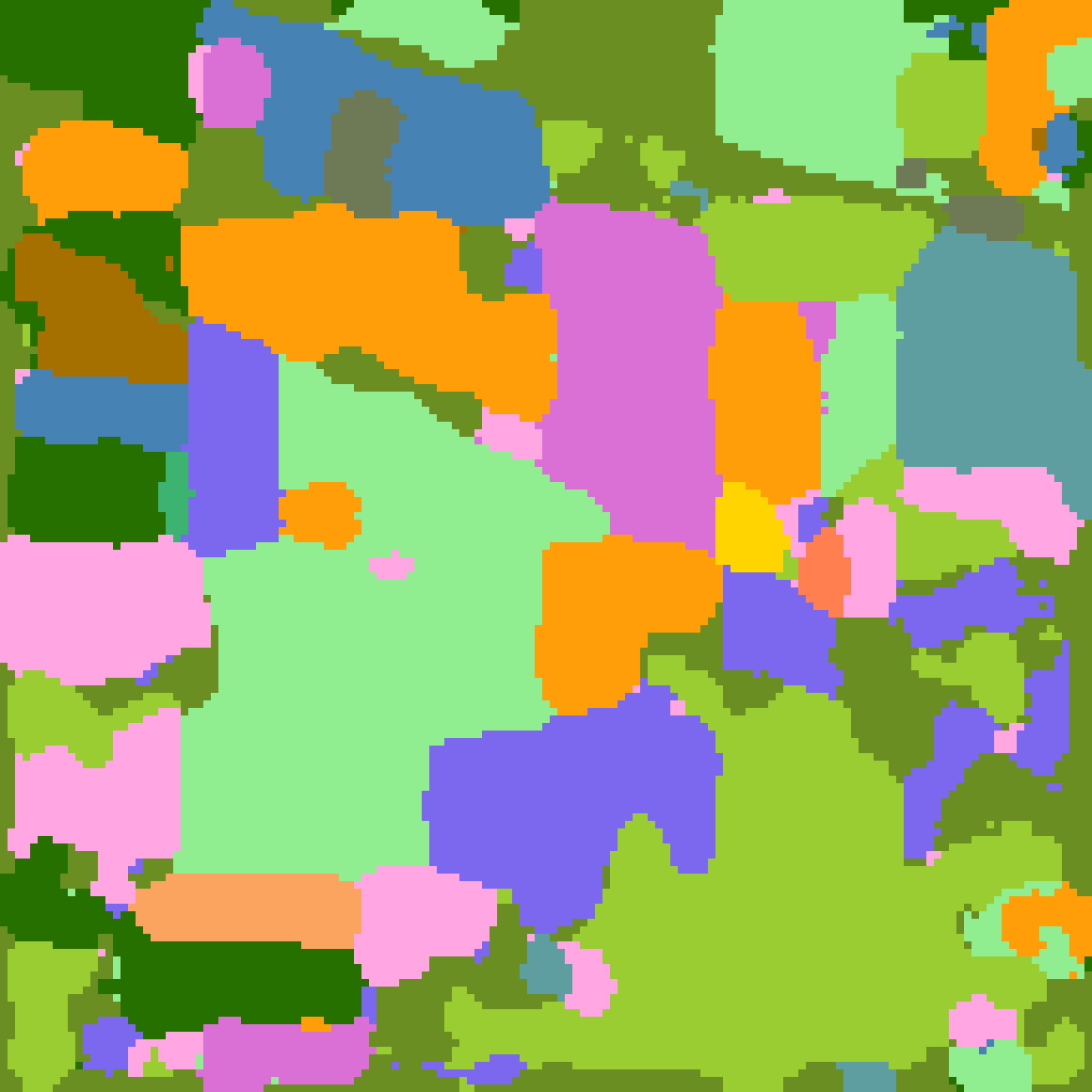}}
  {\includegraphics[width=0.08\textwidth]{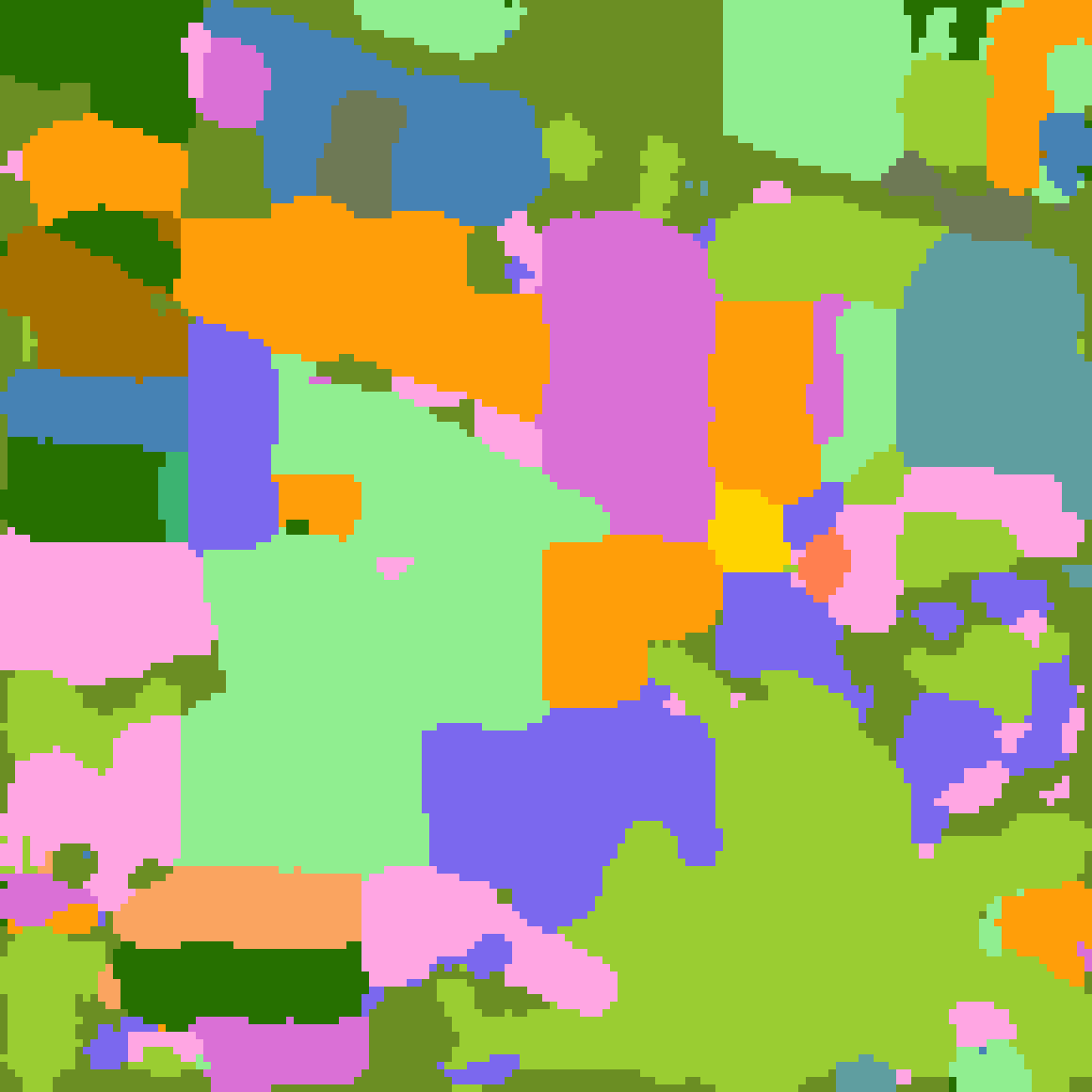}}
  {\includegraphics[width=0.08\textwidth]{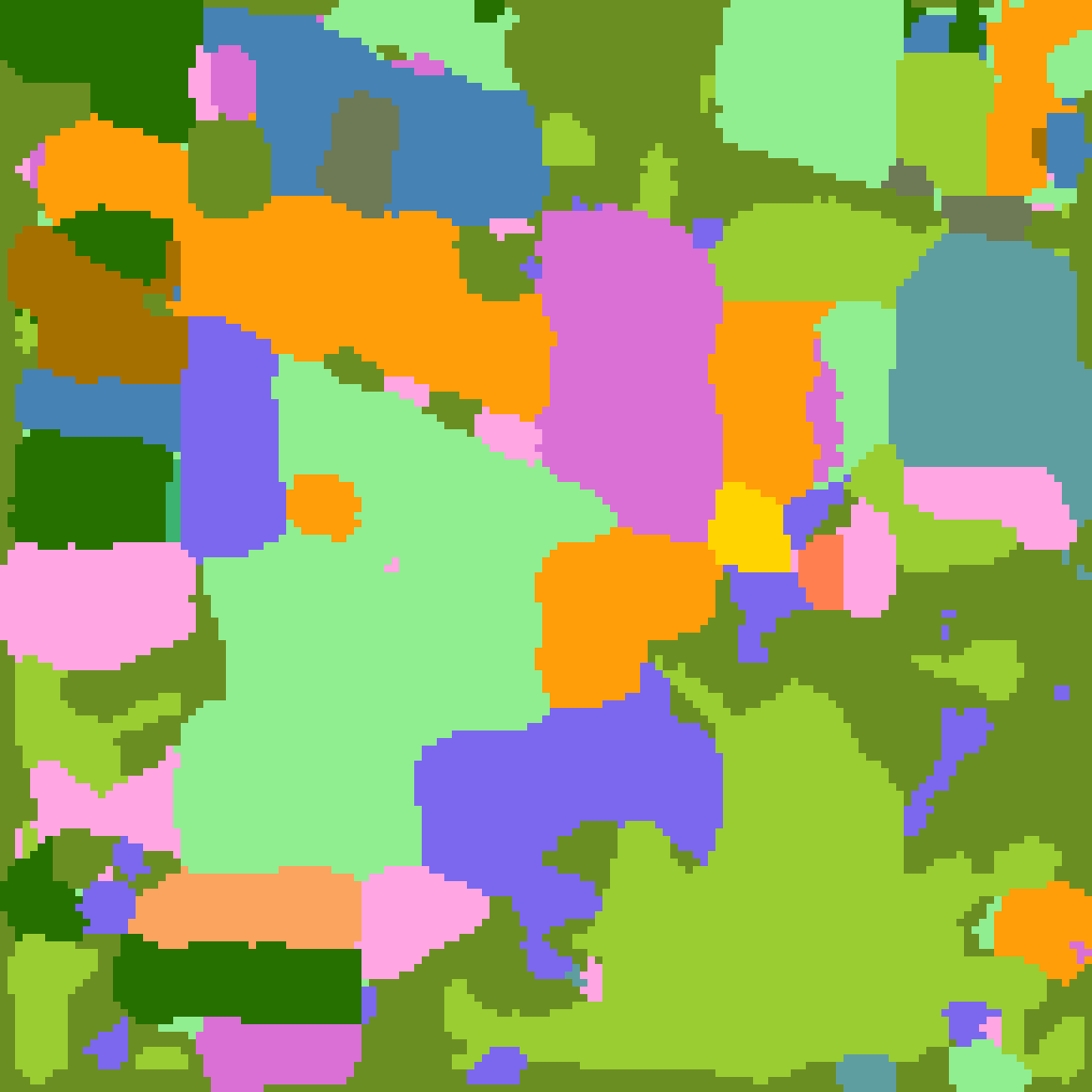}}
  {\includegraphics[width=0.08\textwidth]{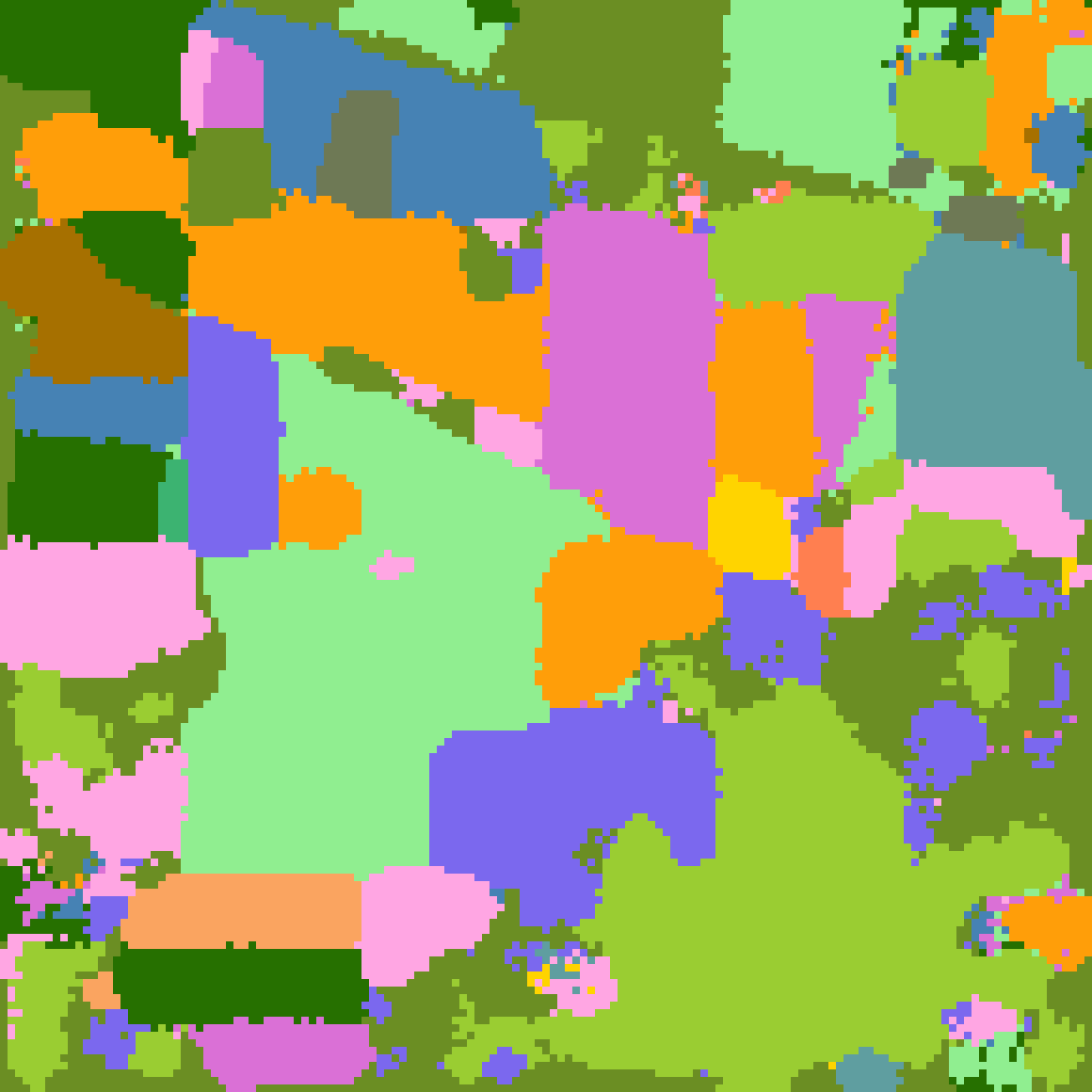}}
  {\includegraphics[width=0.08\textwidth]{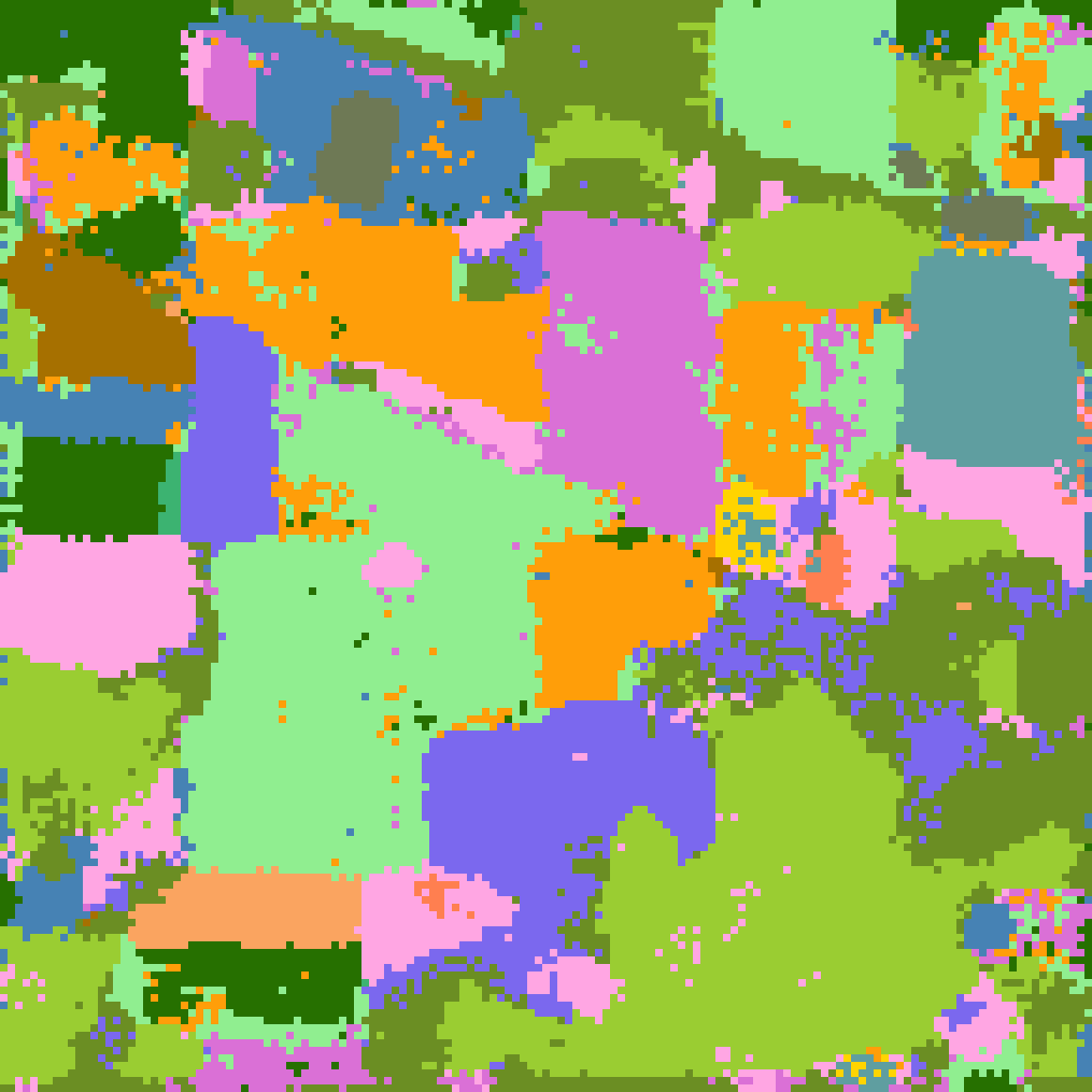}}
  {\includegraphics[width=0.08\textwidth]{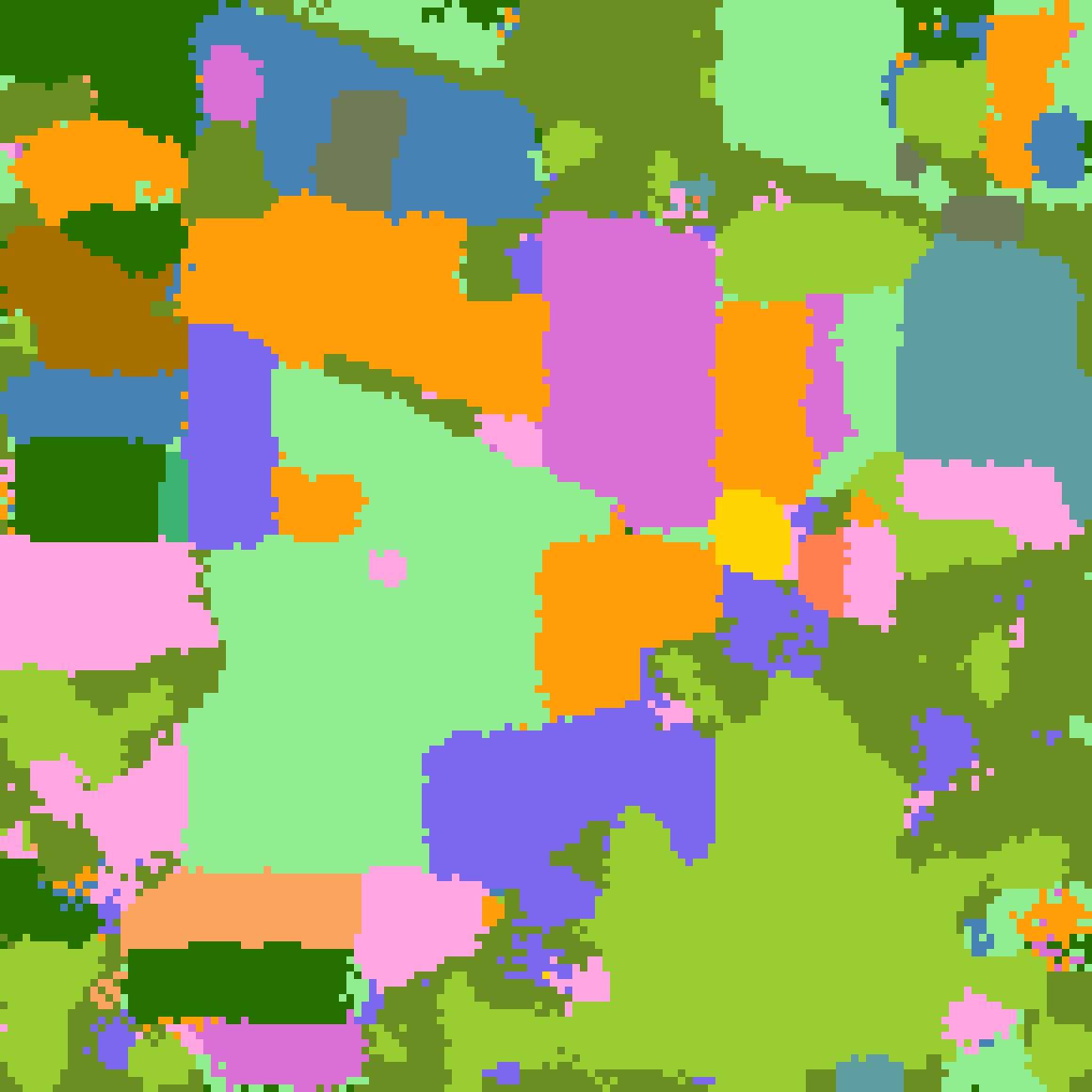}}
  {\includegraphics[width=0.08\textwidth]{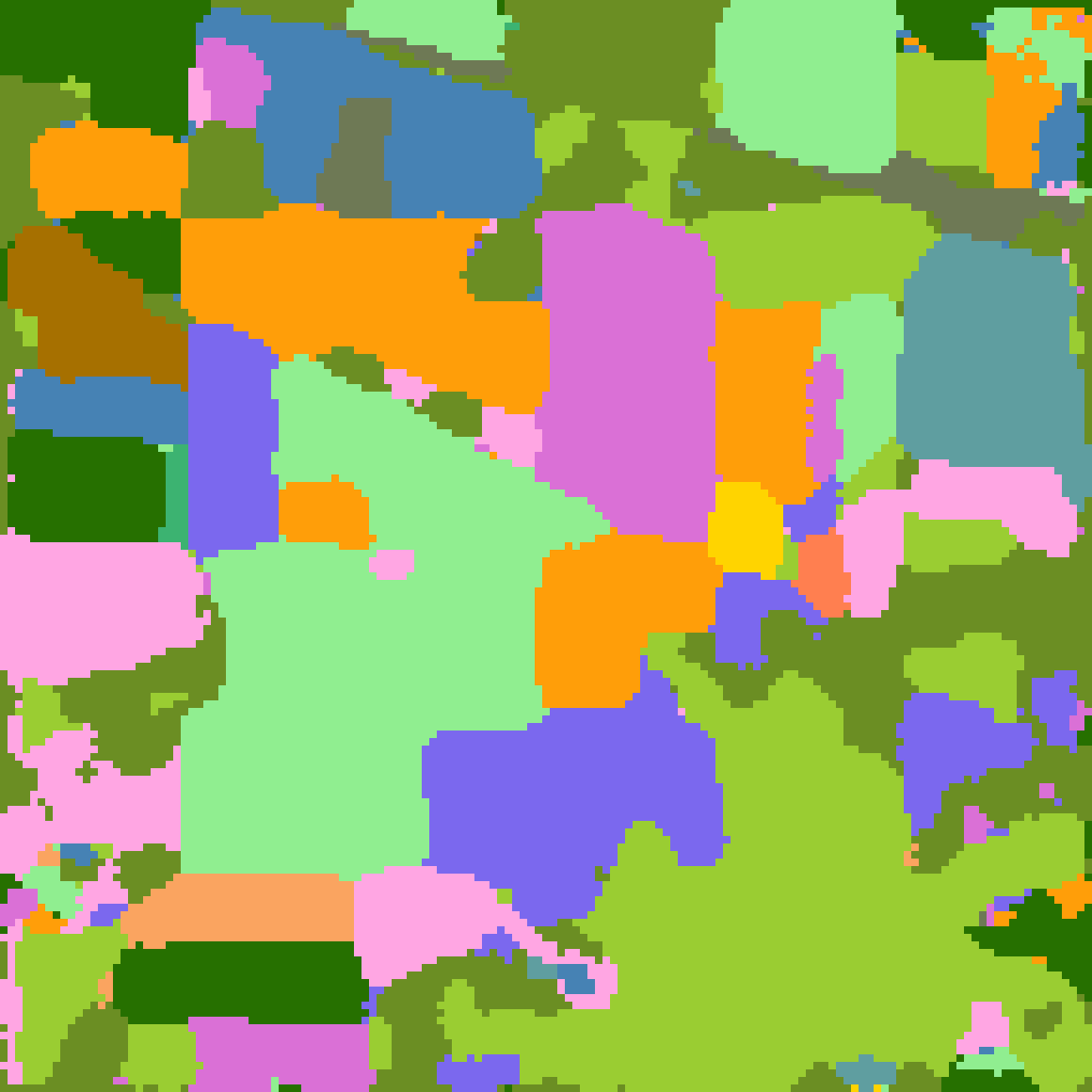}}
  {\includegraphics[width=0.08\textwidth]{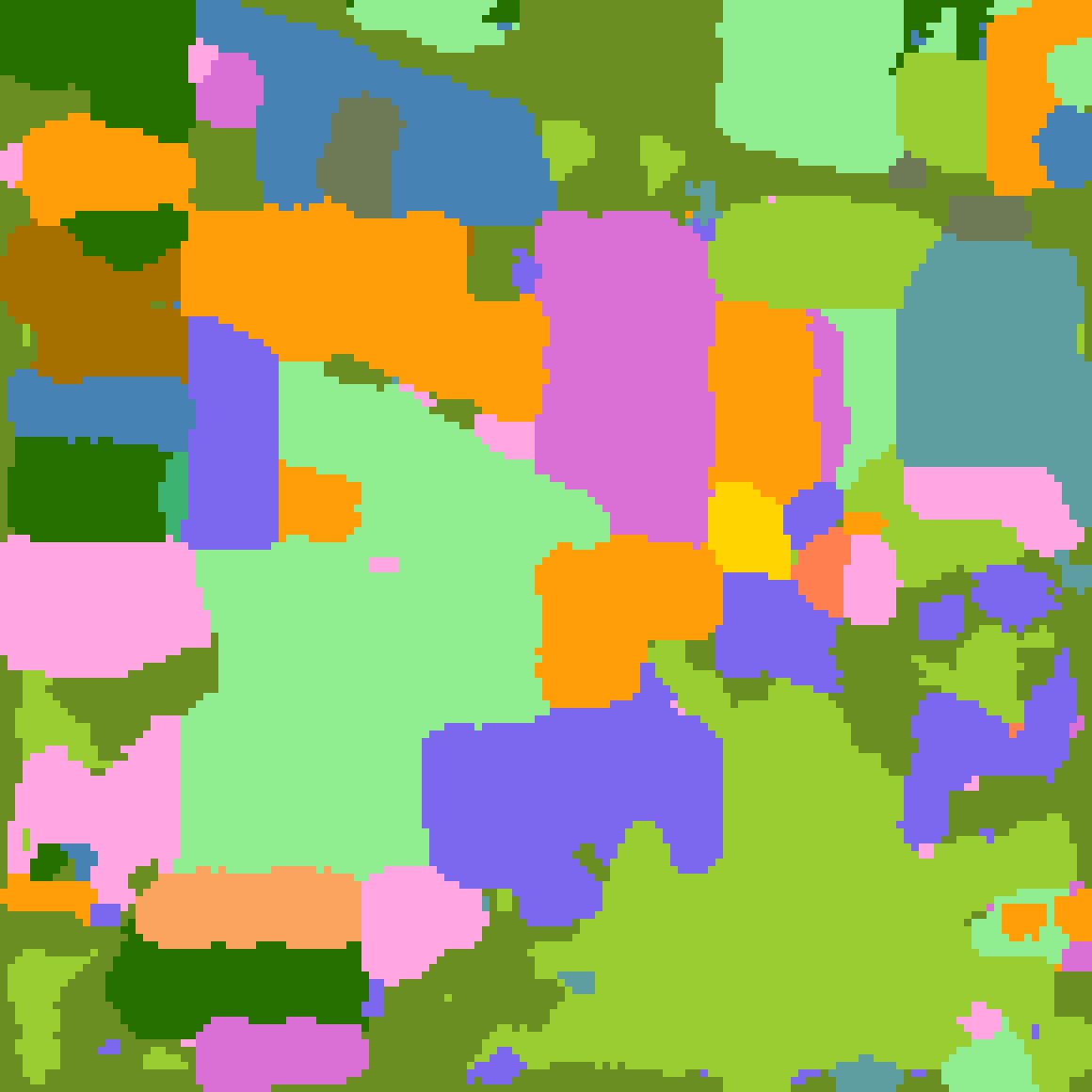}}
  {\includegraphics[width=0.08\textwidth]{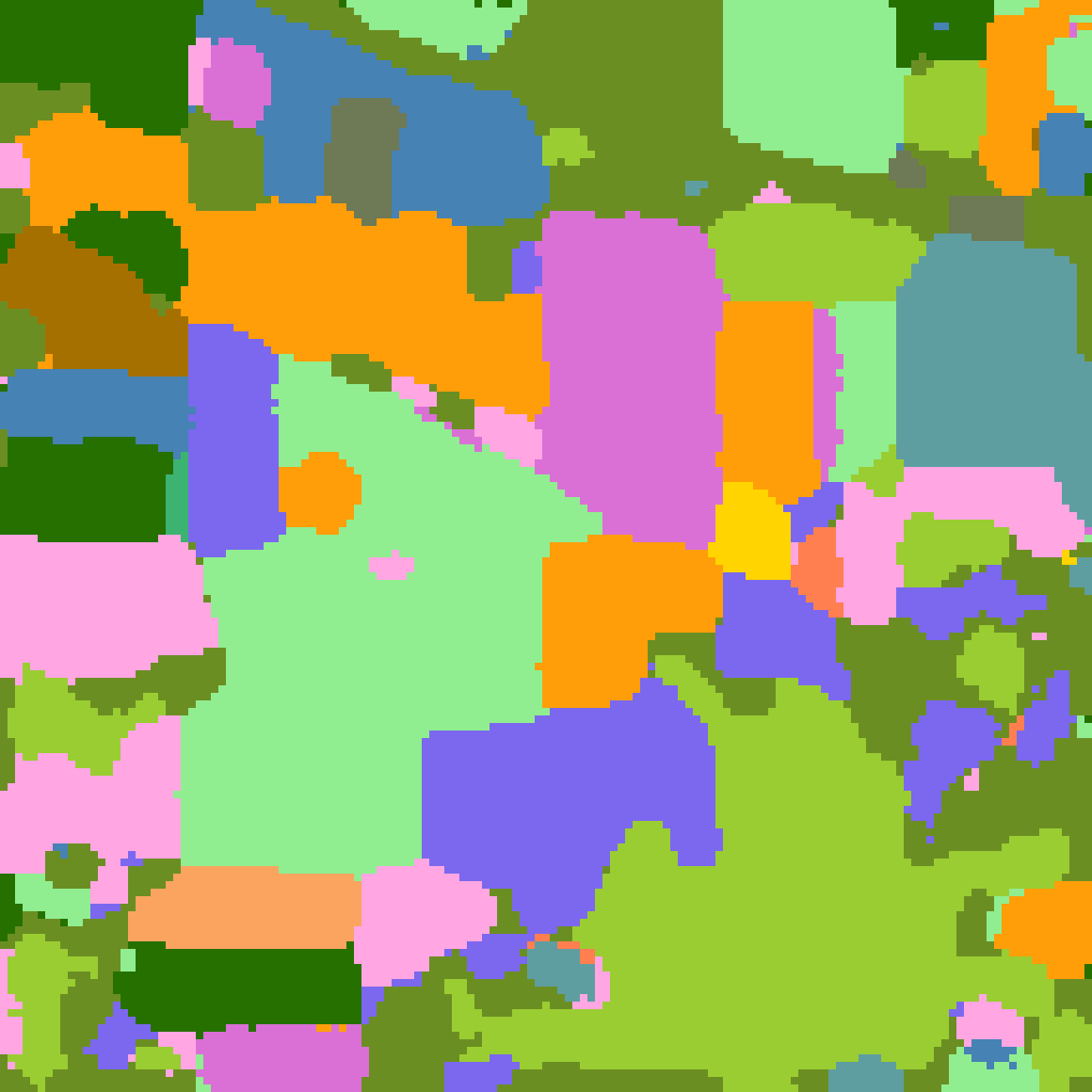}}
  {\includegraphics[width=0.08\textwidth]{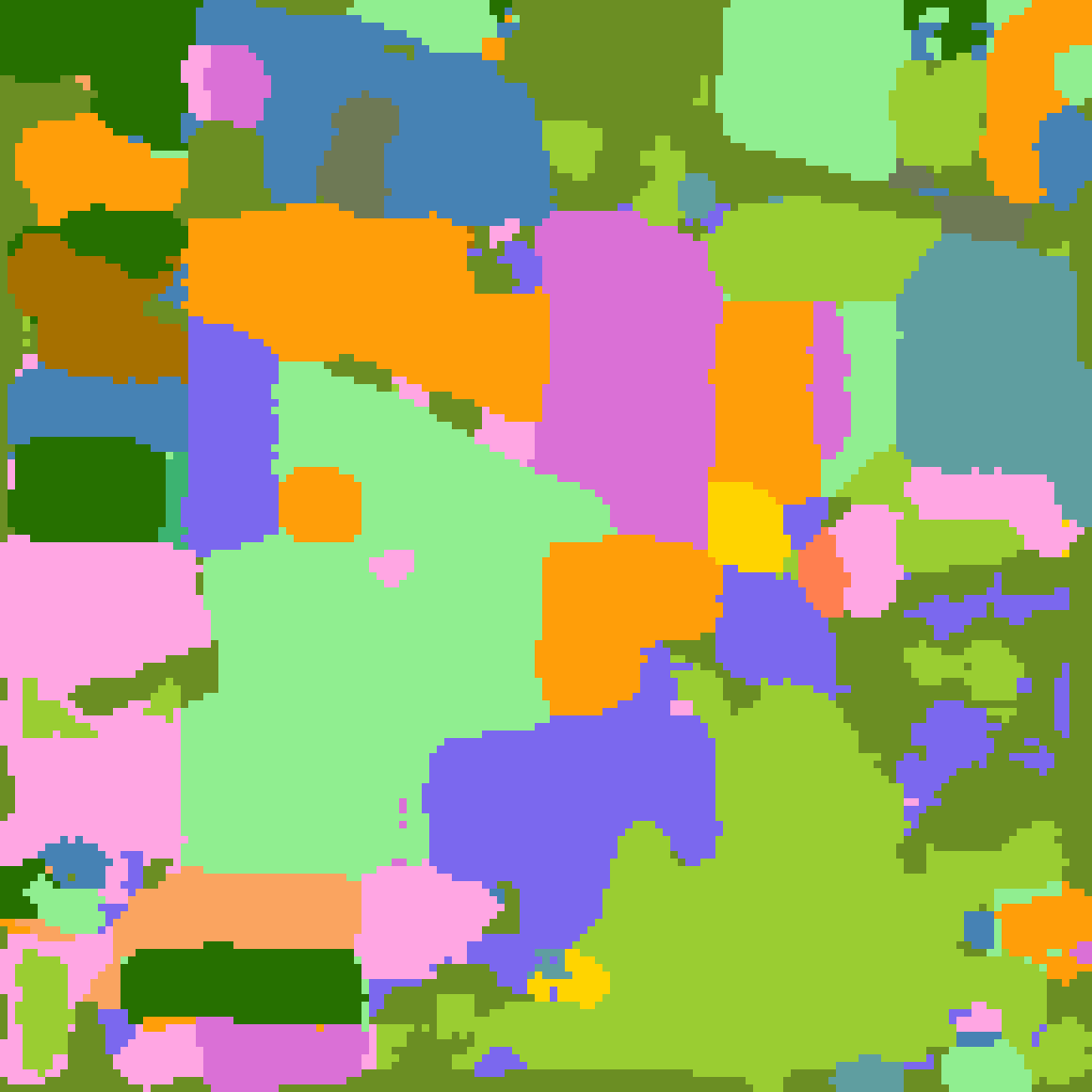}}}
  \vspace{3pt}
  
  \centerline{
  \subfigure[Raw]{\includegraphics[width=0.08\textwidth]{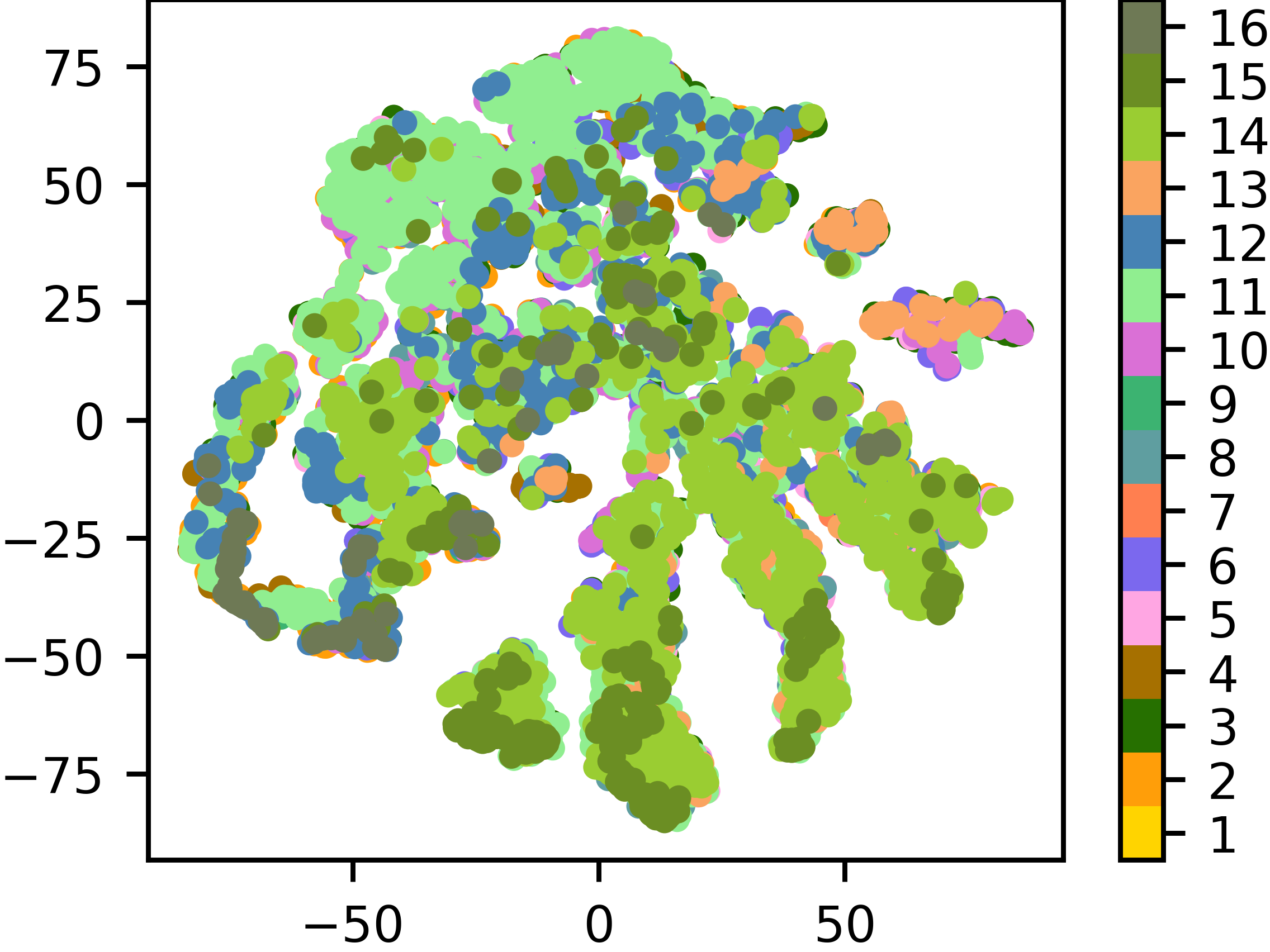}}
  \subfigure[Baseline]{\includegraphics[width=0.08\textwidth]{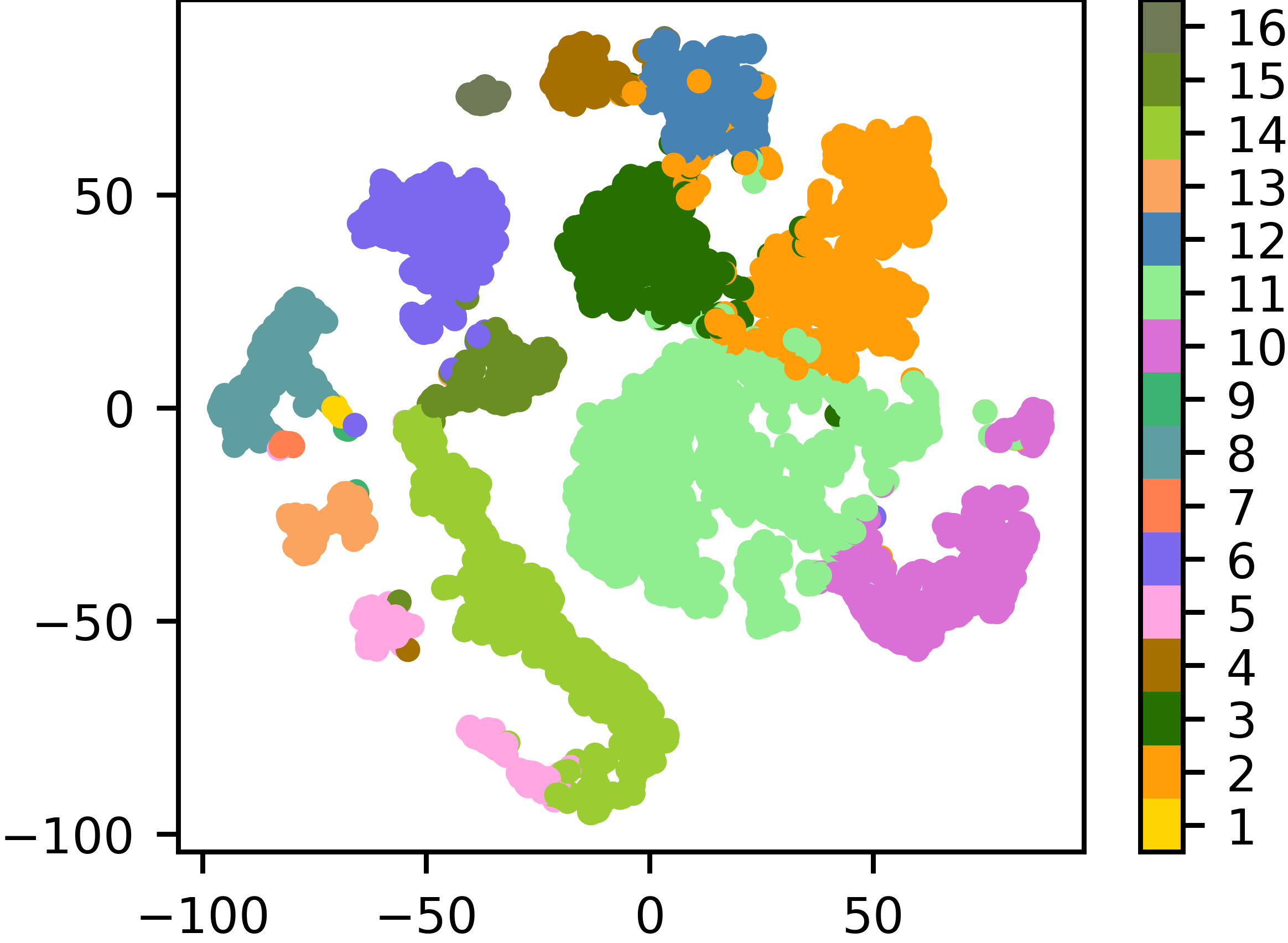}}
  \subfigure[S1]{\includegraphics[width=0.08\textwidth]{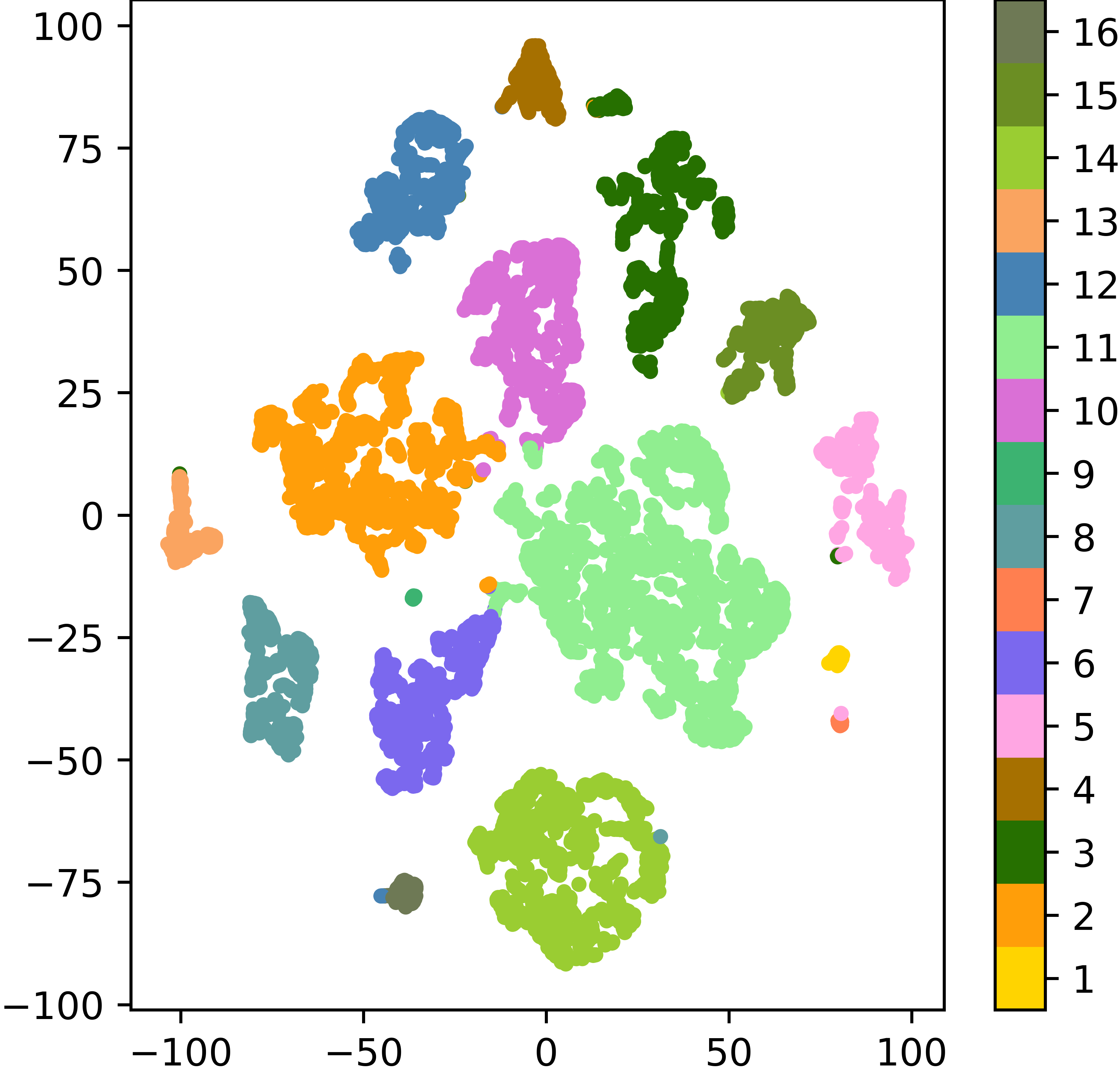}}
  \subfigure[S2]{\includegraphics[width=0.08\textwidth]{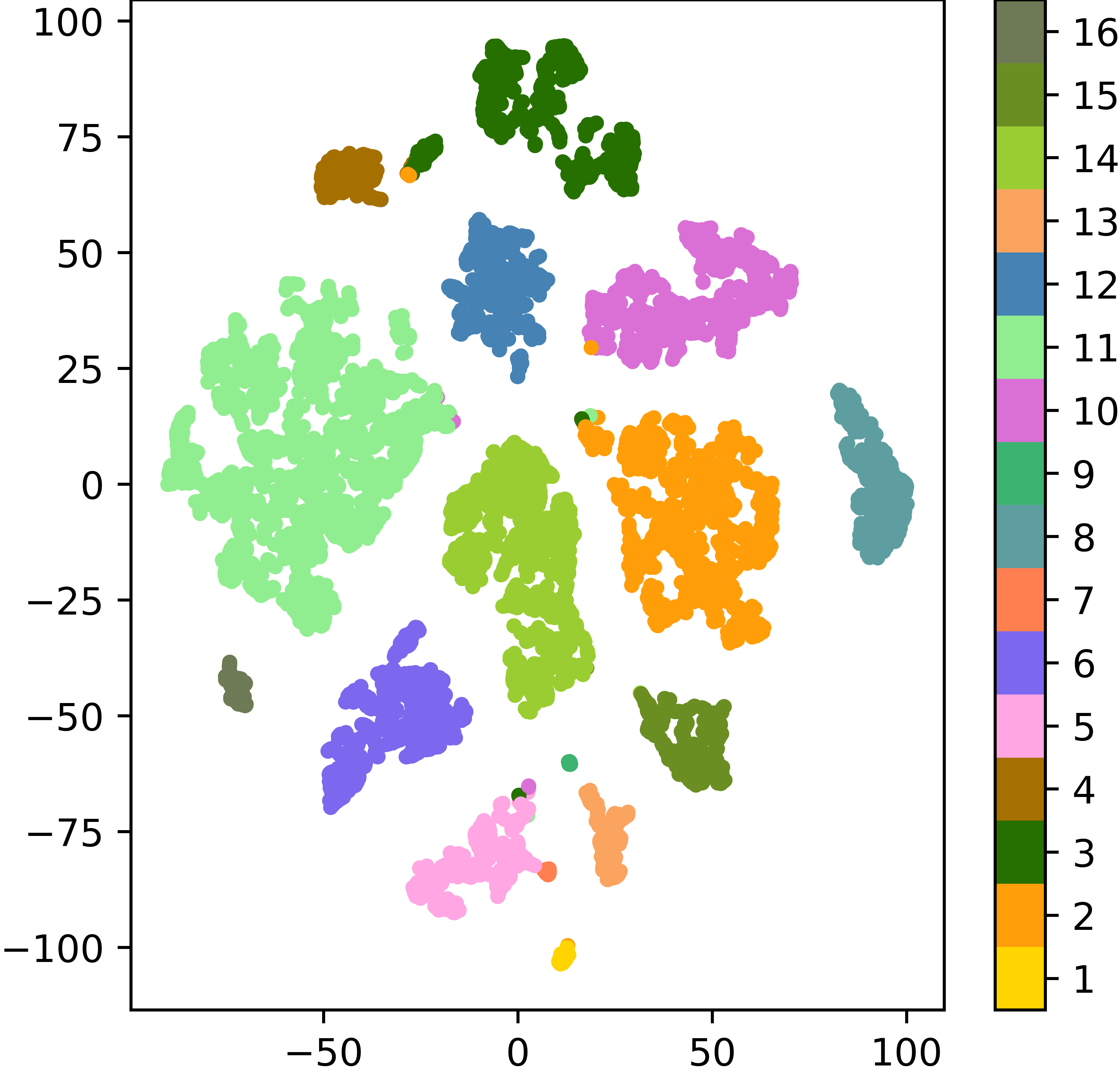}}
  \subfigure[S3]{\includegraphics[width=0.08\textwidth]{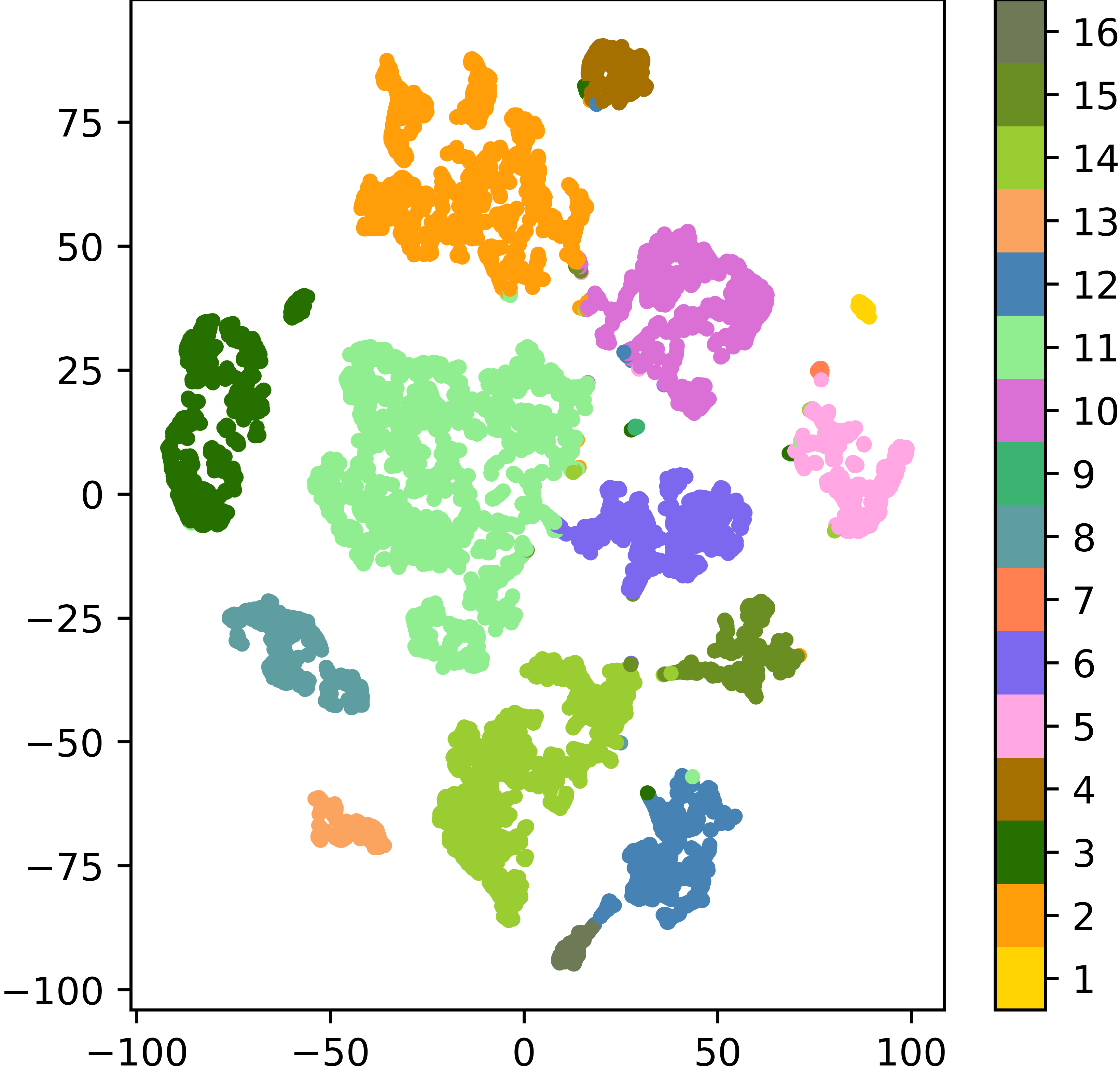}}
  \subfigure[S4]{\includegraphics[width=0.08\textwidth]{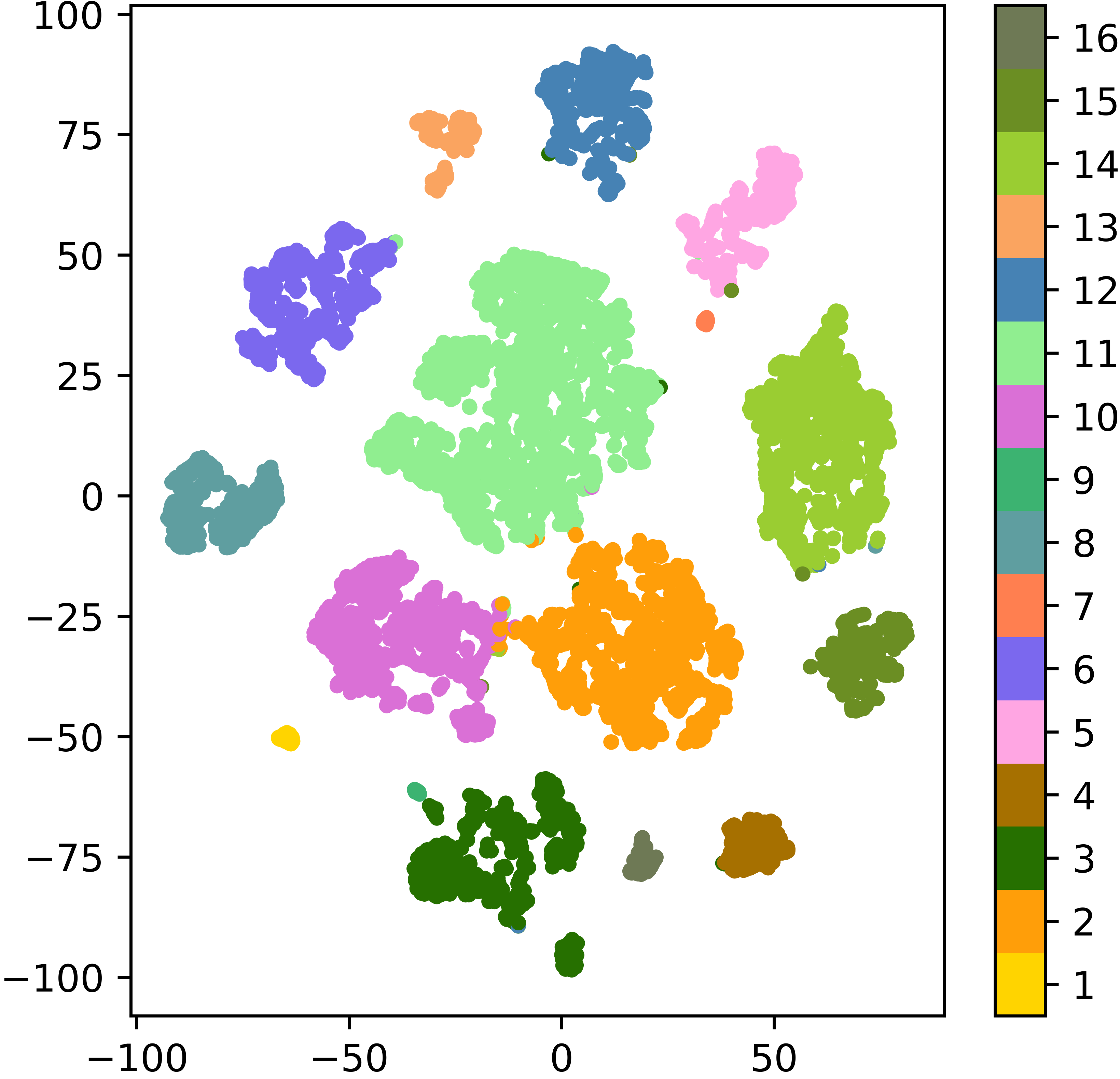}}
  \subfigure[S5]{\includegraphics[width=0.08\textwidth]{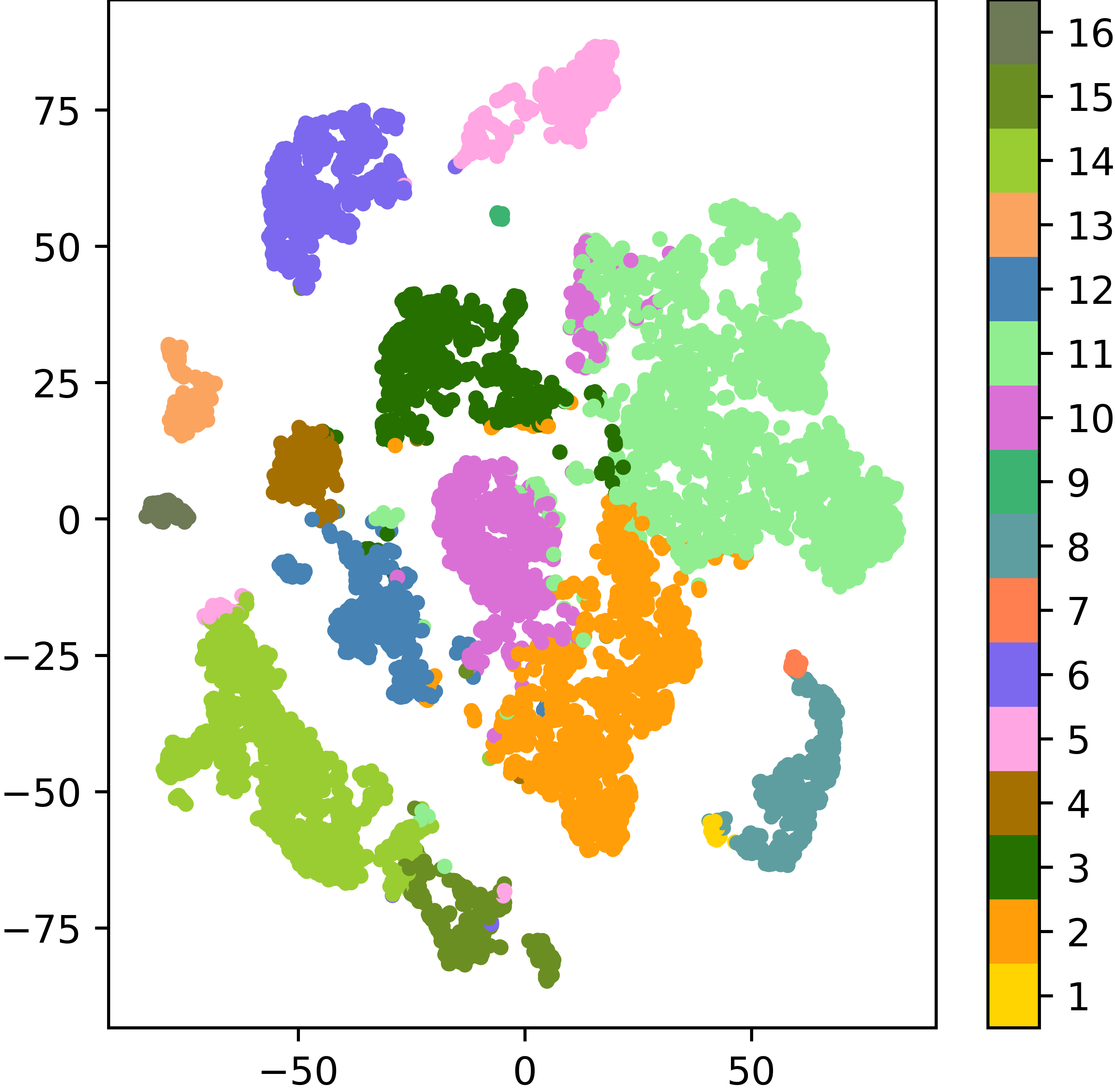}}
  \subfigure[S6]{\includegraphics[width=0.08\textwidth]{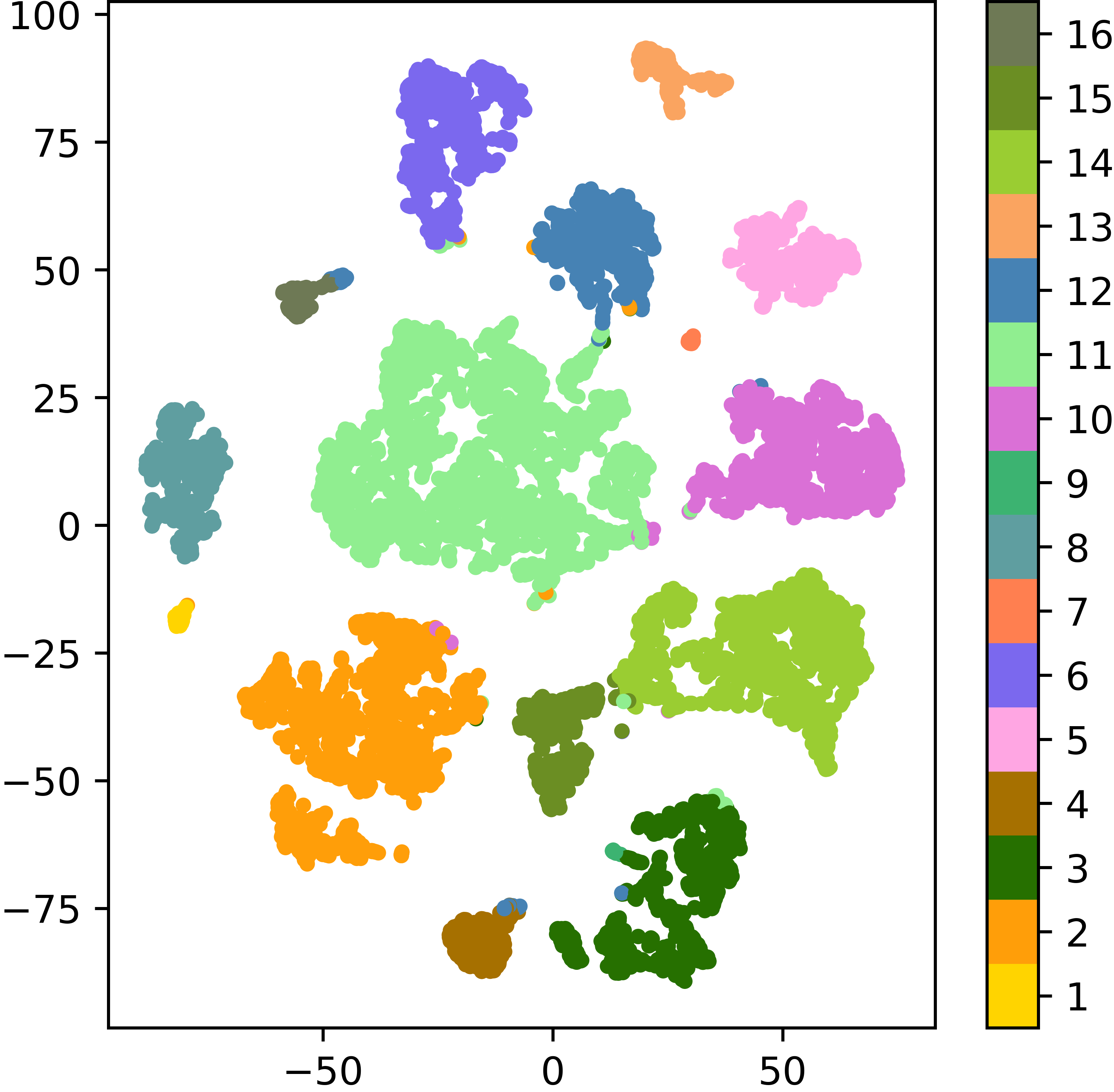}}
  \subfigure[S7]{\includegraphics[width=0.08\textwidth]{figures/exp/IN/IN_TSNE/png/HSIC_DiffCRNTSNE_IN.png}}
  \subfigure[S8]{\includegraphics[width=0.08\textwidth]{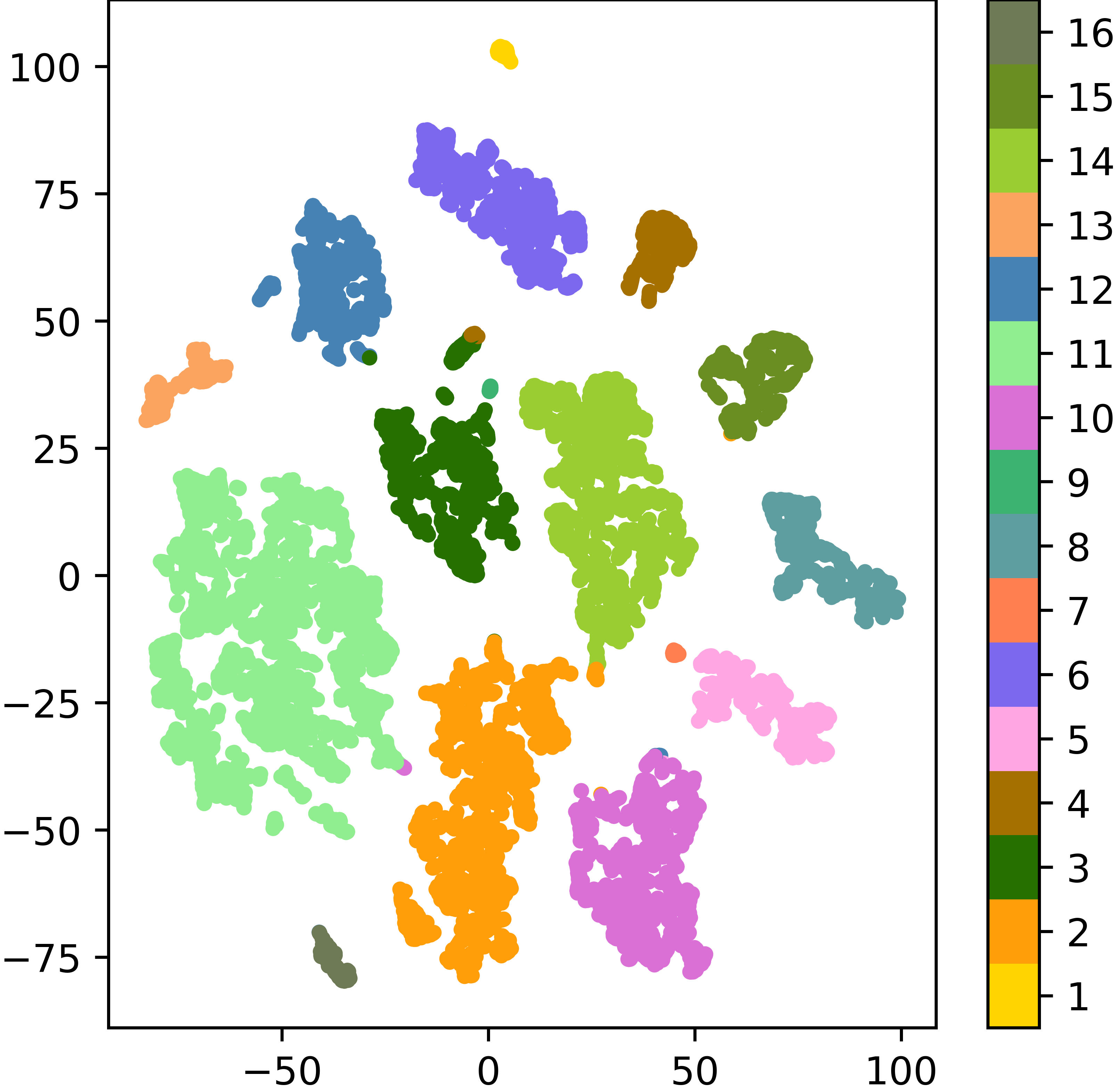}}
  \subfigure[S9]{\includegraphics[width=0.08\textwidth]{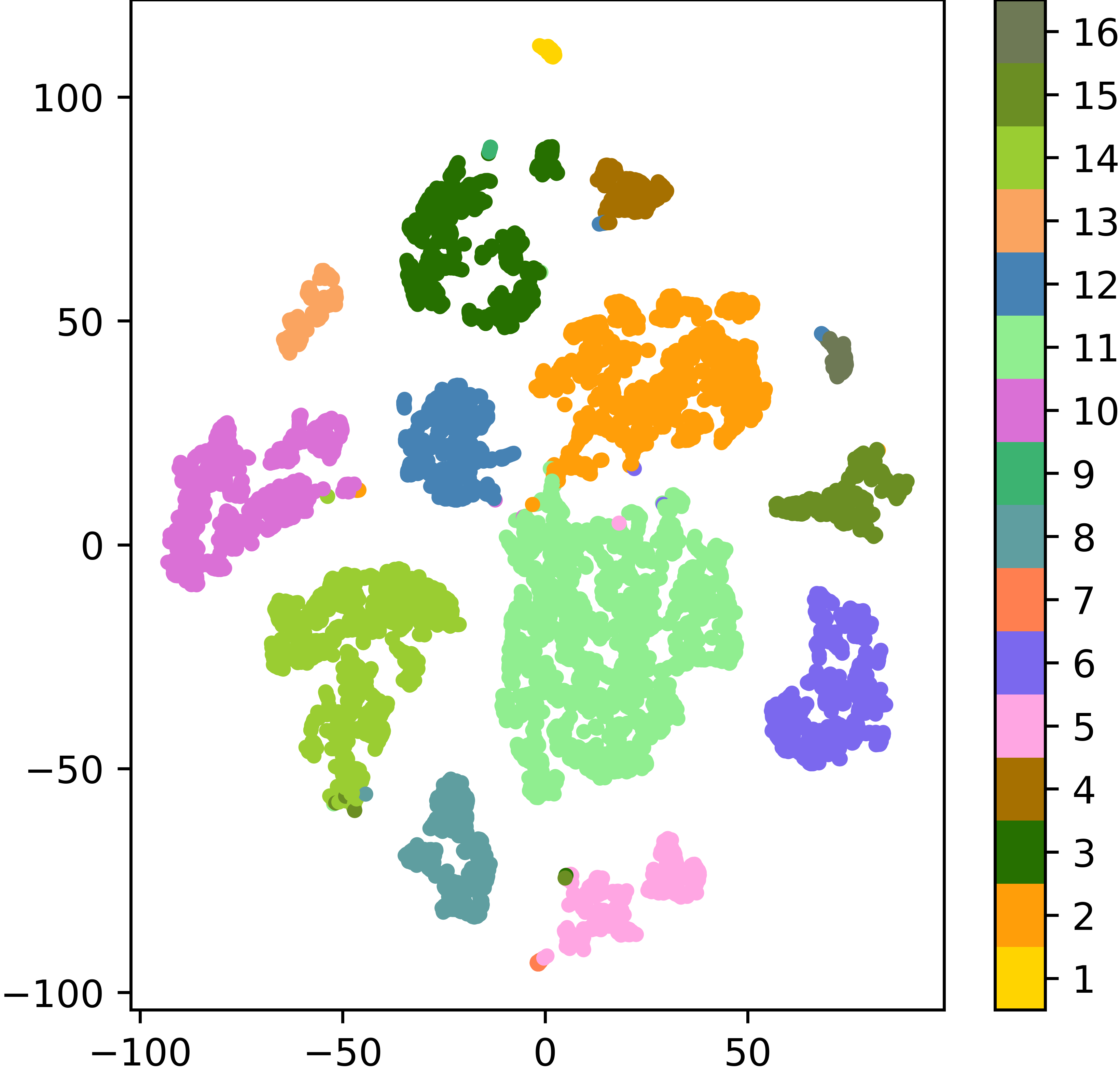}}
  \subfigure[S10]{\includegraphics[width=0.08\textwidth]{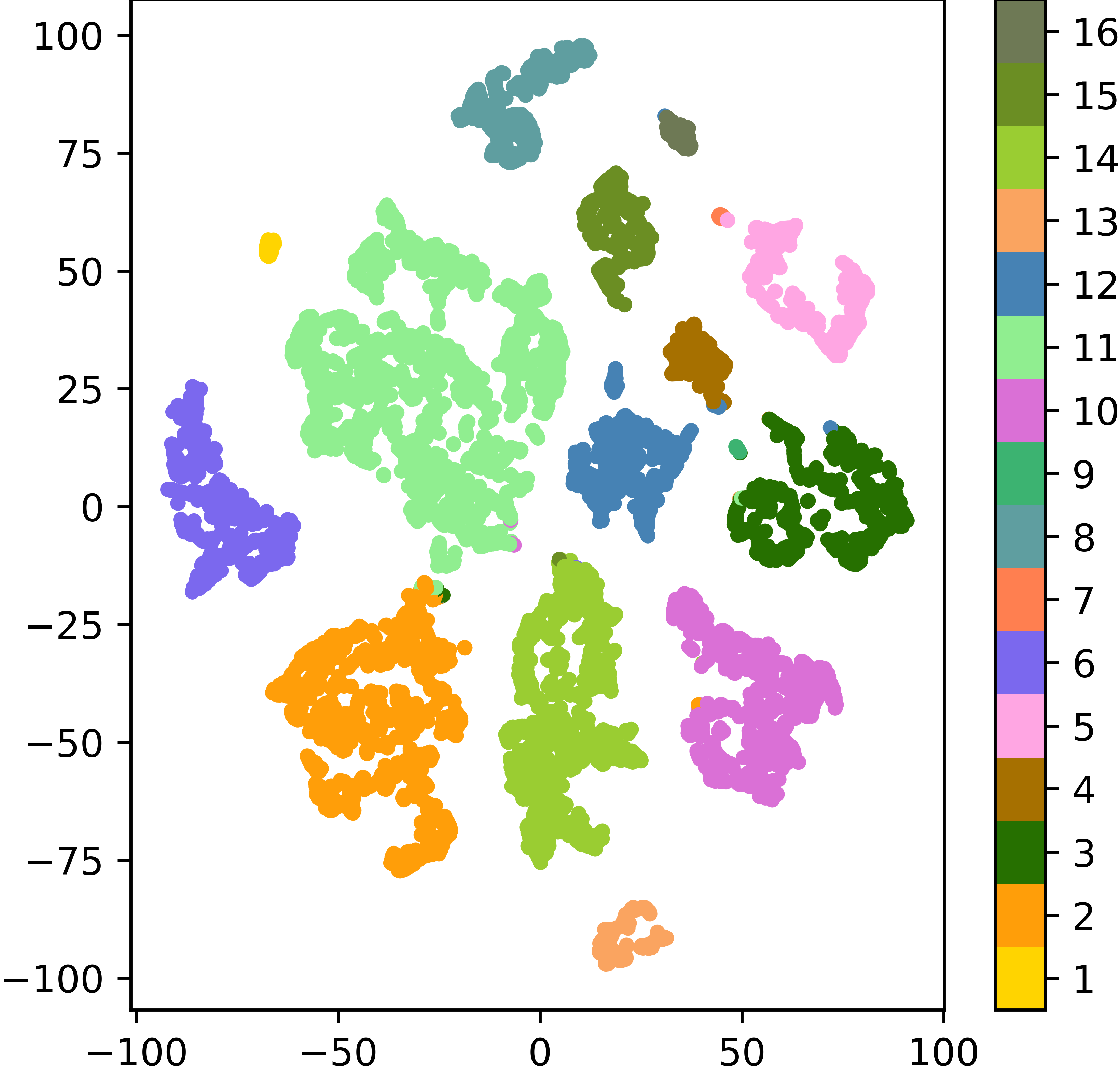}}
  }
  \caption{Classification map under different sets of time step $t$ as listed on \autoref{ablation_timestep} on Indian Pines dataset. Each column corresponds to \autoref{ablation_timestep} in order. The second and fourth row is the corresponding 2-D graphical visualization of the features extracted under different sets of time step through t-SNE. (a) distribution characteristics of raw labelled data. (b) OA=89.13\%. (c) OA=99.27\%. (d) OA=99.18\%. (e) OA=98.68\%. (f) OA=98.80\%. (g) OA=94.09\%. (h) OA=98.69\%. (i) OA=99.33\%. (j) OA=99.23\%. (k) OA=99.18\%. (l) OA=98.45\%. Zoom in for best view.}
\label{time_step_case}

\end{figure*}

\begin{figure*}[htbp]
  \centerline{\includegraphics[width=0.31\textwidth]{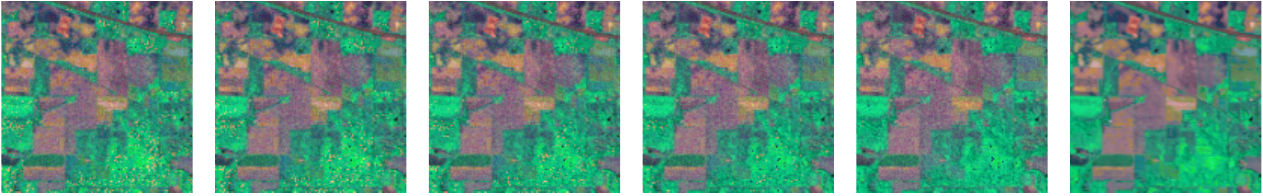}
  {\includegraphics[width=0.31\textwidth]{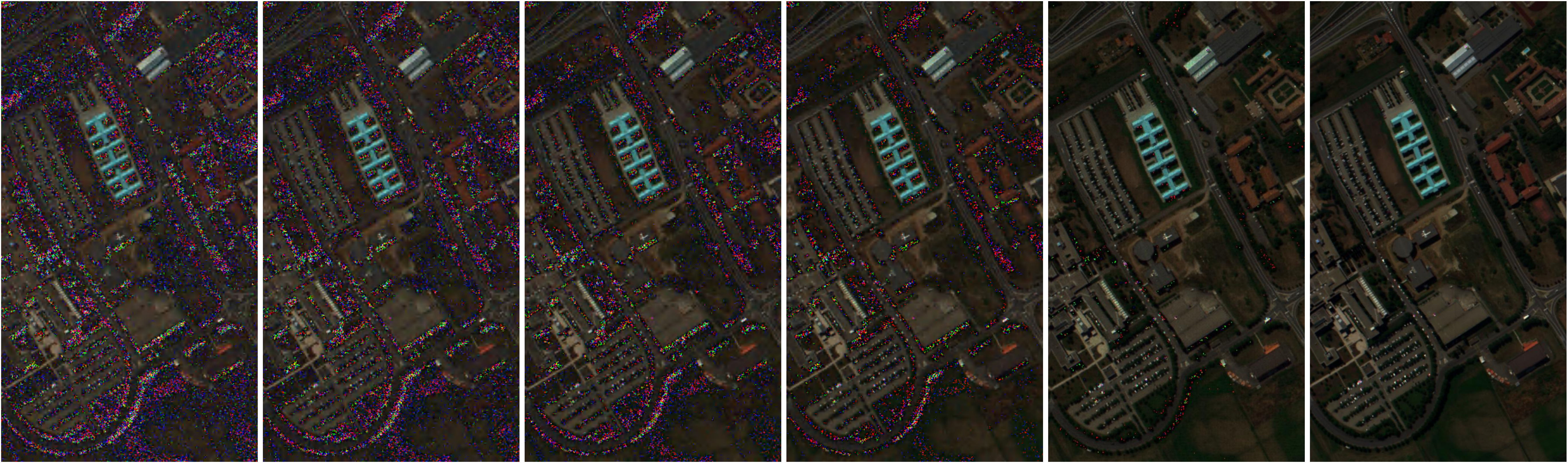}}
  {\includegraphics[width=0.31\textwidth]{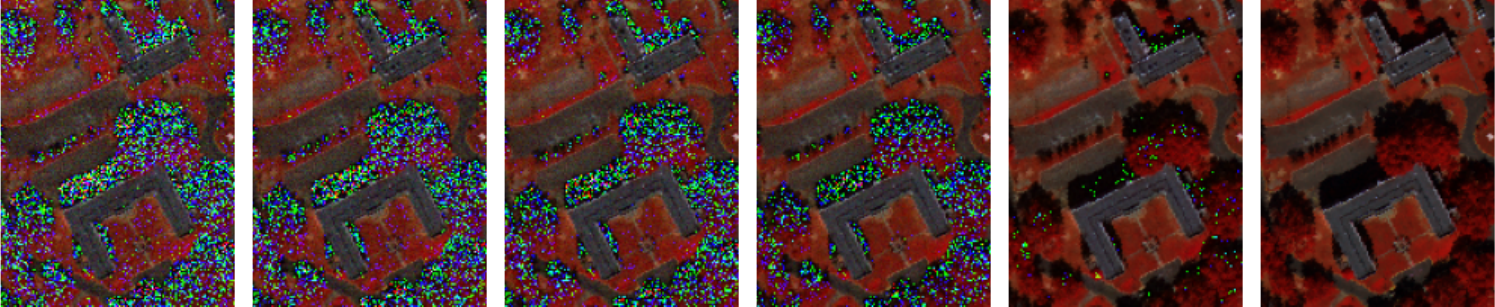}}}
  
  \centerline{\includegraphics[width=0.31\textwidth]{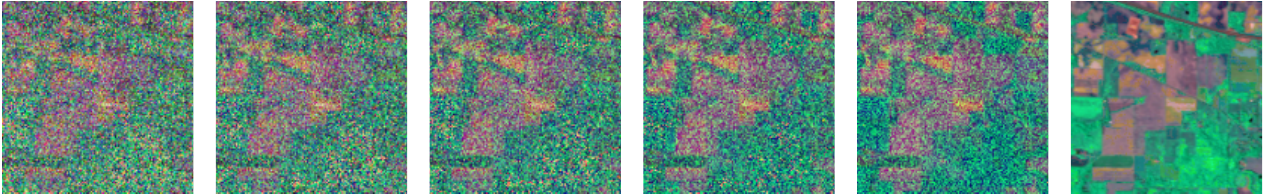}
  \includegraphics[width=0.31\textwidth]{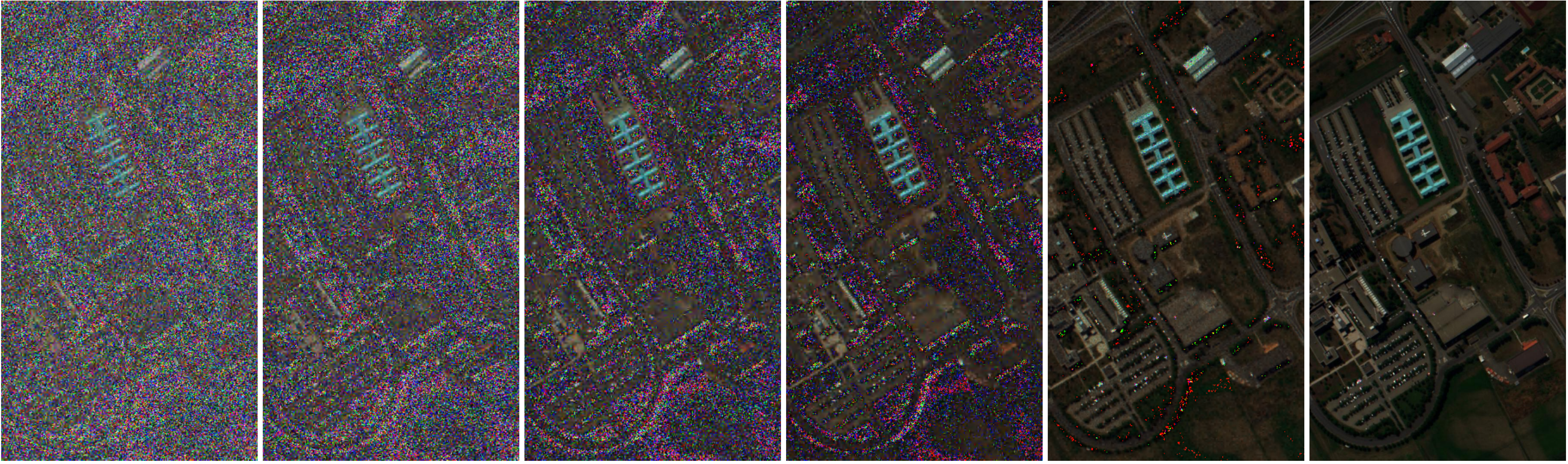}
  \includegraphics[width=0.31\textwidth]{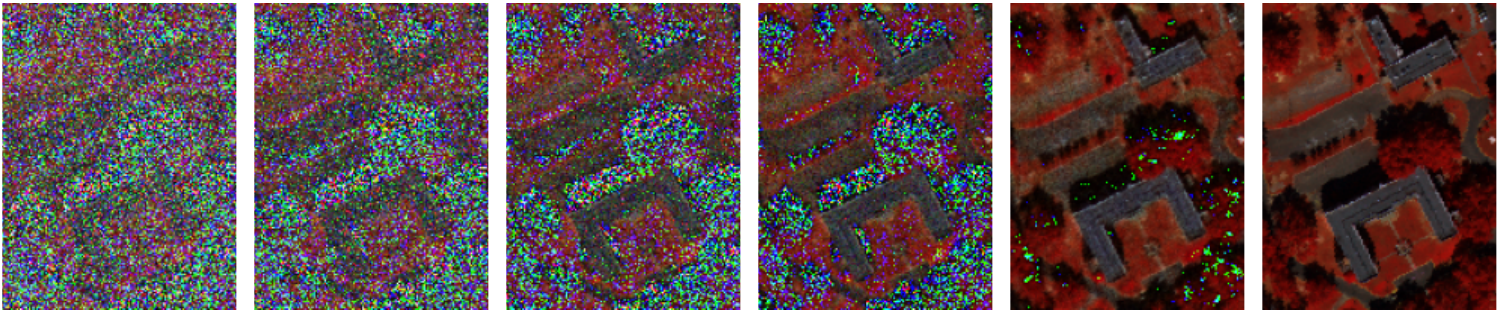}}
  
   \centerline{\includegraphics[width=0.31\textwidth]{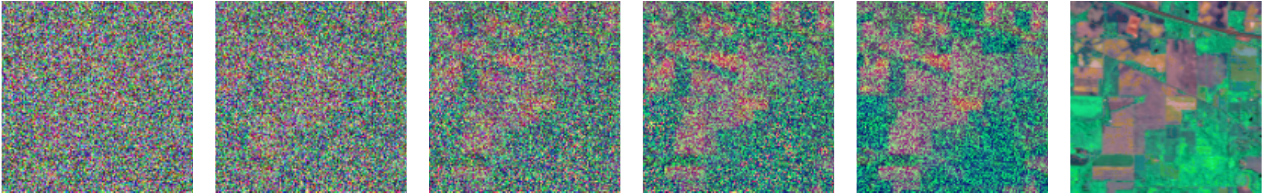}
  \includegraphics[width=0.31\textwidth]{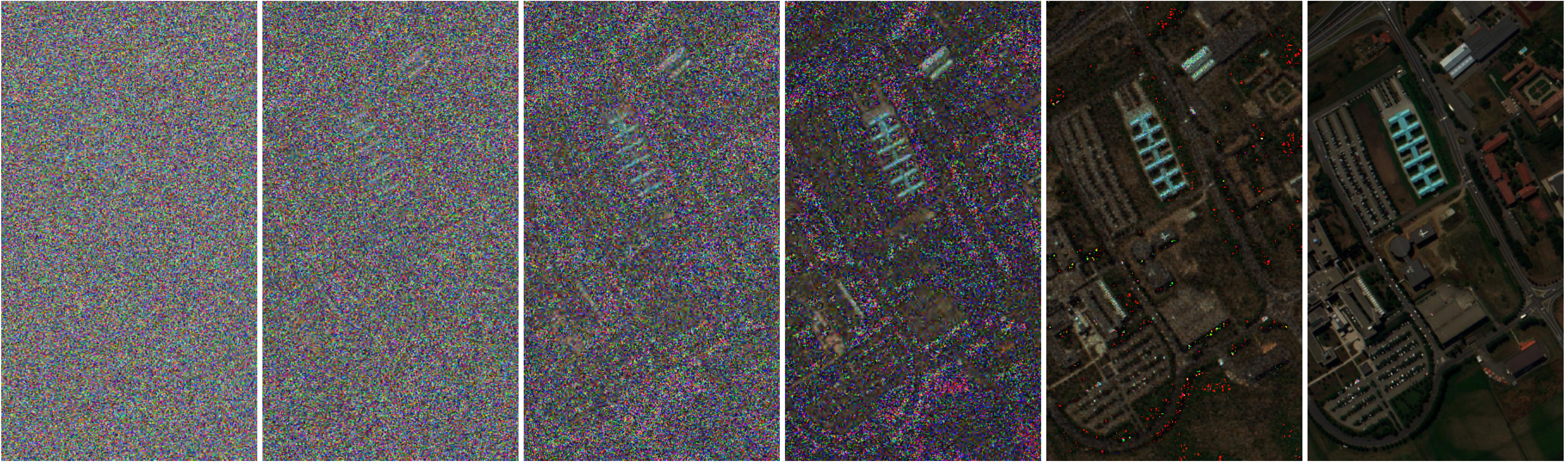}
  \includegraphics[width=0.31\textwidth]{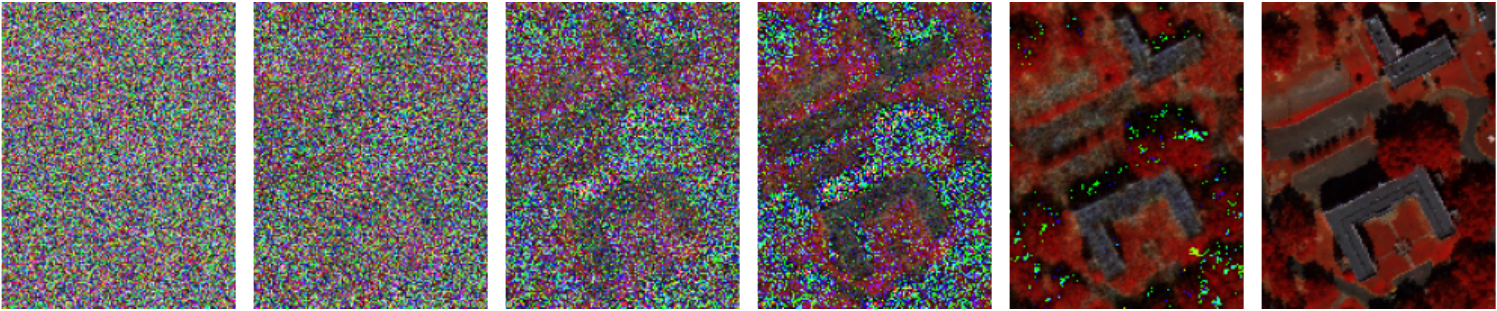}}

   \centerline{\includegraphics[width=0.31\textwidth]{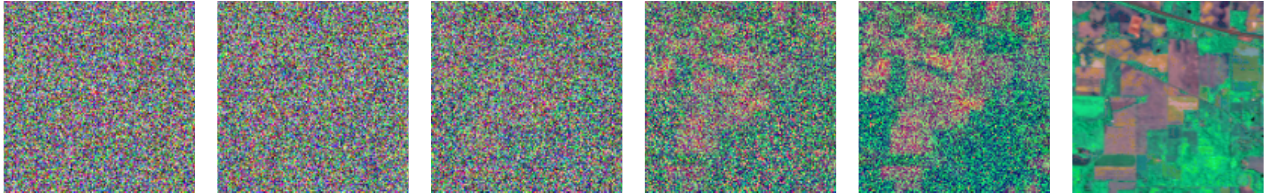}
  \includegraphics[width=0.31\textwidth]{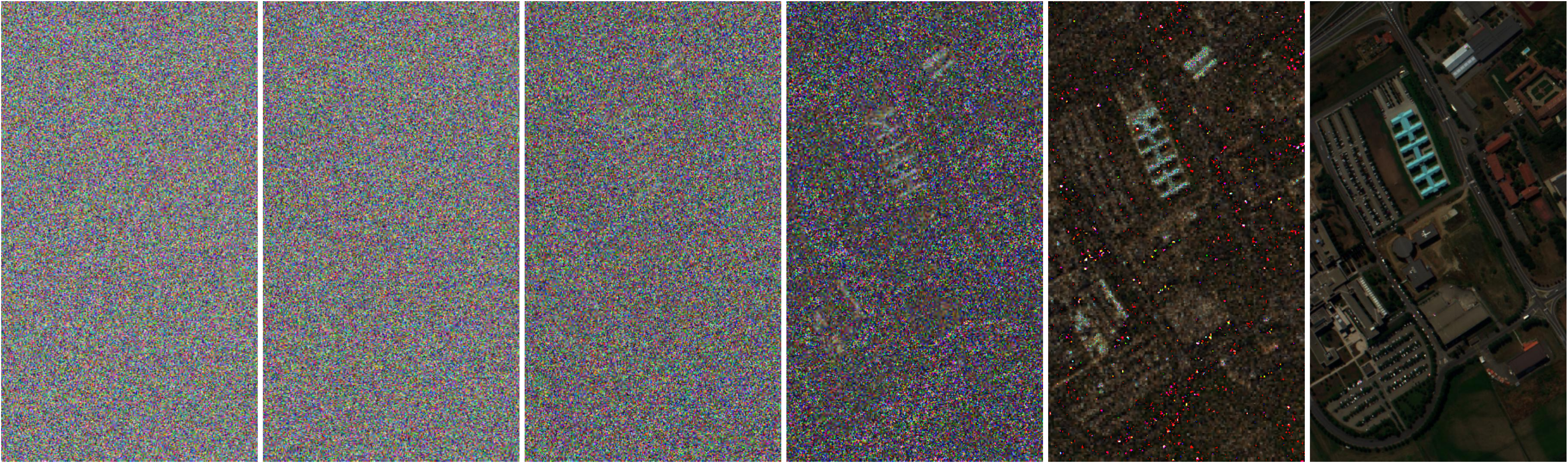}
  \includegraphics[width=0.31\textwidth]{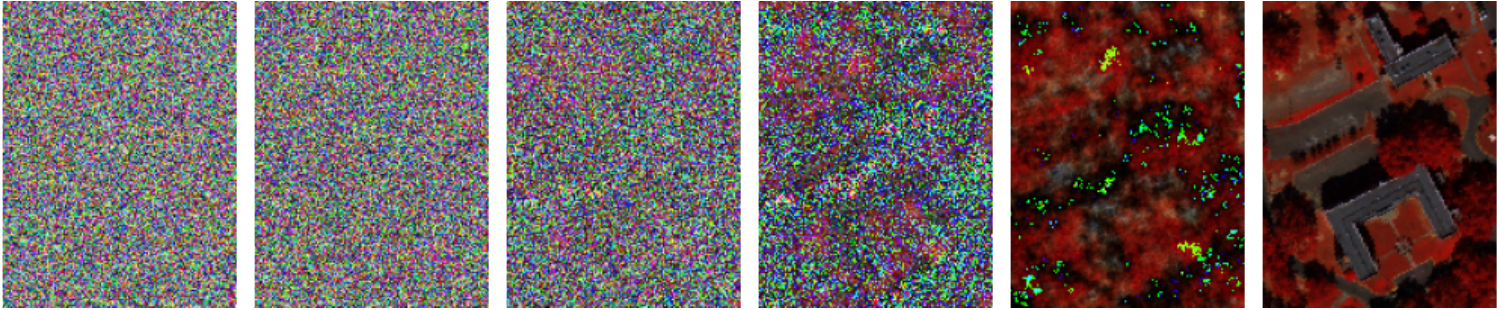}}

  \caption{False-color images of the reconstructed Indian Pines, subset of 
 Pavia University and MUUFL dataset ($t$=110, 510, 1010, 1610, from top to bottom) corresponding to different time steps by reverse spectral-spatial diffusion process. The first column is 
 \(\mathbfcal{H}_{\boldsymbol{t}}\). The penultimate column is the reconstructed \(\mathbfcal{H}_{\boldsymbol{0}}\). The second, third, fourth column are the intermediate results sampled by \autoref{sampling}. The last column is the original hyperspectral false-color image \(\mathbfcal{H}_{\boldsymbol{0}}\).}
  \label{reconstruction}
\end{figure*}

\subsubsubsection{\textbf{Select \(t\) adaptively or manually ?}}

We analyze the representations produced by the noise predictor $\boldsymbol{\epsilon}_{\theta}(\mathbfcal{H}_{\boldsymbol{t}},t)$ for five different sets of diffusion time step $t$. The first six rows of the \autoref{ablation_timestep} are manually selected time step set, while the rest are determined by the learned SAM. As shown in \autoref{ablation_timestep}, we can clearly see the performance difference under different sets of time step $t$. Specifically, it can be seen that the following conditions hold. 
    
\textit{(i)} In the case of manually selecting the time step $t$, the extracted features under larger $t$ with much more noise, hence underperformances time step with small $t$. For example, time step with {{[}700, 900, 1100, 1300, 1500{]}} , only 94.09\% accuracy in term of OA, which is nearly 5\% lower than that of time step with {[}50, 100, 200, 300, 500{]}. Time step with small (i.e., {[}50, 100, 200, 300, 500{]}) and medium $t$ (i.e., {{[}700, 900, 1100, 1300, 1500{]}}) have approximate performance.

\textit{(ii)} Time step $t$ determined by the learned SAM value performs better than  manually assigned $t$, further performs better than time step with {{[}0, 1, 2, 3, 4{]}}, {[}0, 1, 2, 3, 4, 5, 6, 7, 8, 9{]} and {[}0, 1, 2, 3, 4, ... , 26, 27, 28, 29{]}, which demonstrates the effectiveness of building the intermediate features according to SAM. Three different scenarios according to the top few value of SAM have the similar performance.
    
\textit{(iii)}  Furthermore, we visualize the features in 2-dimensional space using t-distributed stochastic neighbor embedding (TSNE) to show the diversity of features at different sets of time step $t$. Although the original distribution characteristics of raw data are complex i.e., \autoref{time_step_case} (a), we can observe that by using SAM to extract features, samples of similar categories gather together, and intraclass variance is minimized, i.e, \autoref{time_step_case} (i)-(l). On the contrary, it can be seen that the features learned by manually assigned $t$ are less discriminative, i.e., \autoref{time_step_case} (g), as well as baseline method, i.e., \autoref{time_step_case} (b), which also reflect in the corresponding classification map full of salt and pepper noise.

\subsubsection{\textbf{Diffusion Model Analysis}}


Since diffusion model can convert the gaussian noise distribution into target distribution \(\mathbfcal{H}_{\boldsymbol{0}}\). In this section, we analyze the restoration and reconstruction performance of DiffCRN, in other words, we used the pre-trained diffusion model to recover and reconstruct the hyperspectral imagery data. As shown in \autoref{reconstruction}, the first column is \(\mathbfcal{H}_{\boldsymbol{t}}\), 
which is almost completely destroyed as the diffusion steps increase, e.g., \(t=1010, 1610, 1790\). Even so, from a visual perspective, the diffusion model 
can basically recover the details of hyperspectral imagery.

\section{Conclusion} \label{section6}
In this study, we leverage the DDPM to model the local-global spectral-spatial relationships in an unsupervised manner. To narrow the distance between added gaussian noise and reconstructed ones, instead of widely used MSE loss function in DDPM, we use a logarithm loss function (LAE) to optimize diffusion model, and achieving better performance. To improve the instance-level and interclass discriminability of the diffusion model, we introduce Contrastive Learning to build discriminative representations, which has a positive effect on the classification task. Furthermore, instead of selecting manually time step \(t\) to build multi-timestep representations, we propose a simple but effective way to adaptively choose multi-timestep representations based on pixel-level spectral angle mapping (SAM), simultaneously, the proposed AWAM and CTSSFM module can be effectively combined with DiffCRN to get refined features. Extensive evaluations show that the proposed model, DiffCRN, performs favorably against state-of-the-art supervised and unsupervised methods methods on hyperspectral image classification on the four widely used datasets.

However, the proposed methods utilized a two-stage learning strategy, which leads to heavy training burden and inconvenience when used for practice. In future work, we will improve the proposed methods to be end-to-end framework, and explore conditional diffusion model to generate realistic hyperspectral image via language description.

\bibliographystyle{IEEEtran}
\bibliography{IEEEabrv,ref}


\end{document}